\documentclass[lettersize,journal]{IEEEtran}
\usepackage{amsmath,amsfonts}
\usepackage{algorithmic}
\usepackage{algorithm}
\usepackage{array}
\usepackage[caption=false,font=footnotesize]{subfig}
\usepackage{textcomp}
\usepackage{stfloats}
\usepackage{url}
\usepackage{verbatim}
\usepackage{graphicx}
\usepackage{cite}
\usepackage{amssymb}
\newtheorem{definition}{Definition}

\newtheorem{proposition}{Proposition}
\newtheorem{property}{Property}
\def\ie{\emph{i.e.}}
\def\eg{\emph{e.g.}}

\usepackage[]{hyperref}
\usepackage{multirow, booktabs, colortbl, threeparttable, makecell}  % 表格相关
\usepackage{enumitem}  % 自定义列表环境，会影响IEEE默认值

\makeatletter

\newcommand{\Rmnum}[1]{\expandafter\@slowromancap\romannumeral #1@}
\makeatother

\begin{document}
% \begin{CJK*}{UTF8}{gkai}

\title{Weak Pareto Boundary: The Achilles' Heel of Evolutionary Multi-Objective Optimization}


\author{Ruihao Zheng, Jingda Deng, and Zhenkun Wang,~\IEEEmembership{Senior Member,~IEEE}
        % <-this % stops a space
% \thanks{This work was partially supported by the National Natural Science Foundation of China (Grant No. 62476118), the Natural Science Foundation of Guangdong Province (Grant No. 2024A1515011759), the Natural Science Foundation of Shenzhen (Grant No. JCYJ20220530113013031) and the Guangdong Science and Technology Program (Grant No. 2024B1212010002). (Corresponding author: Zhenkun Wang)}
\thanks{R. Zheng is with the School of Automation and Intelligent Manufacturing, Southern University of Science and Technology, Shenzhen 518055, P.R. China. (e-mail: 12132686@mail.sustech.edu.cn)}
\thanks{J. Deng is with the School of Computer Science and Engineering, Xi'an University of Technology, Xi’an 710048, P.R. China. (e-mail: jddeng@xaut.edu.cn)}
\thanks{Z. Wang is with the School of Automation and Intelligent Manufacturing and also with the Department of Computer Science and Engineering, Southern University of Science and Technology, Shenzhen 518055, P.R. China. (e-mail: wangzhenkun90@gmail.com)}
\thanks{Corresponding author: Zhenkun Wang.}
}

% \author{Anonymous submission}


% \markboth{IEEE Transactions on Evolutionary Computation}%
% {Zheng \MakeLowercase{\textit{et al.}}: (Title)}


\maketitle

\begin{abstract}

The weak Pareto boundary ($WPB$) refers to a boundary in the objective space of a multi-objective optimization problem, characterized by weak Pareto optimality rather than Pareto optimality. The $WPB$ brings severe challenges to multi-objective evolutionary algorithms (MOEAs), as it may mislead the algorithms into finding dominance-resistant solutions (DRSs), \ie, solutions that excel on some objectives but severely underperform on the others, thereby missing Pareto-optimal solutions. Although the severe impact of the $WPB$ on MOEAs has been recognized, a systematic and detailed analysis remains lacking. To fill this gap, this paper studies the attributes of the $WPB$. In particular, the category of a $WPB$, as an attribute derived from its weakly Pareto-optimal property, is theoretically analyzed. The analysis reveals that the dominance resistance degrees of DRSs induced by different categories of $WPB$s exhibit distinct asymptotic growth rates as the DRSs in the objective space approach the $WPB$s, where a steeper asymptotic growth rate indicates a greater hindrance to MOEAs. Beyond that, experimental studies are conducted on various new test problems to investigate the impact of $WPB$'s attributes. The experimental results demonstrate consistency with our theoretical findings. Experiments on other attributes show that the performance of an MOEA is highly sensitive to some attributes. Overall, no existing MOEAs can comprehensively address challenges brought by these attributes.

\end{abstract}


\section{Introduction}
\IEEEPARstart{M}{any} real-world optimization problems, from engineering design~\cite{osyczka1978approach} to scientific discovery~\cite{macleod2022self}, involve optimizing more than one objective simultaneously. They are known as multi-objective optimization problems (MOPs)~\cite{hillermeier2001nonlinear,jahn2011vector}. Since the objectives are usually conflicting, a set of Pareto-optimal solutions is demanded rather than a single one.
The multi-objective evolutionary algorithm (MOEA), by virtue of its population-based stochastic search strategy, has become a preferred method for efficiently identifying a diverse set of approximate solutions in a single run~\cite{deb2002fast,zhang2007moea,coello2007evolutionary}. The obtained set of solutions enables the decision-maker to analyze trade-offs among candidate solutions and identify the most preferred one in a posteriori manner~\cite{deb2011multi}.

The weak Pareto boundary ($WPB$), as illustrated in \figurename~\ref{fig:illu_WPB_2D}, is a boundary in the objective space. It contains only objective vectors that are weakly Pareto-optimal but not Pareto-optimal. The $WPB$ is observed in many real-world MOPs, including the multi-objective procedural map generation~\cite{togelius2010towards}, the multi-line distance minimization problem~\cite{li2018multiline}, the multi-objective traveling salesman problem~\cite{corne2007techniques}, and the MOP introduced by the interactive method~\cite{branke2008multiobjective}.
% with the region of interest
The $WPB$ poses significant challenges to MOEAs~\cite{wang2024multi}, and thus, it is also called the hardly-dominated boundary in some studies~\cite{wang2019scalable,ishibuchi2020effects}. Specifically, the $WPB$ can induce the MOEA to maintain numerous dominance-resistant solutions (DRSs)~\cite{ikeda2001failure} in its population. On the one hand, the DRS describes an undesirable solution that exhibits exceptionally poor function values for certain objectives. On the other hand, unfortunately, it is very difficult for the MOEA to eliminate DRSs from the population. This is because the DRS achieves favorable function values for the other objectives. The objective vector on the $WPB$, a special DRS in the objective space, is especially intractable due to its weak Pareto optimality. As a result, the presence of $WPB$s in the MOP can substantially impair the performance of MOEAs.

% \IEEEpubidadjcol

Although the negative effects of the $WPB$ on MOEAs have been widely recognized, no studies have systematically analyzed the $WPB$. The current studies on the $WPB$ only have a few empirical findings. For example, the $WPB$ is considered similar to the boundary of the extremely convex Pareto front~\cite{yang2021hard}. In the experiments of~\cite{wang2024multi}, the $WPB$s of an MOP are studied as a whole, without a detailed examination of the differences among $WPB$s. The limited understanding of the $WPB$ and its effects on MOEAs hinders the development of efficient coping strategies. Furthermore, existing test problems fail to adequately capture the diversity of $WPB$s and are often coupled with other problem characteristics (\eg, complicated Pareto set~\cite{lin2024multi}). Using them to evaluate the impact of the $WPB$ on MOEAs does not yield comprehensive insights and may lead to misleading conclusions.

This work takes the first step in investigating the impact of the $WPB$ theoretically and experimentally. Four attributes of the $WPB$ are involved: category, shape, size, and spatial relation to the Pareto front. The contributions of this paper are summarized as follows:
\begin{itemize}
    \item The $WPB$ is defined, and its categories are delineated based on the weak Pareto optimality (Section~\ref{ssec:wpb}).
    \item The degree of dominance resistance (DDR) of a DRS is defined and utilized to theoretically analyze the hindrance of the $WPB$ to MOEAs (Section~\ref{sec:analysis}). Different categories of $WPB$s impose substantially different levels of hindrance on MOEAs, as the DDRs of their corresponding DRSs exhibit distinct asymptotic growth rates when the objective vectors of DRSs approach the $WPB$s. The shape of a $WPB$ also significantly affects DDRs of its corresponding DRSs.
    \item Two test problem generators are developed (Section~\ref{sec:problem}). Each of them can be used to produce various test problems, encompassing different categories, shapes, and sizes of $WPB$s, while the difference between them lies in the spatial relation between the $WPB$ and the Pareto front. To highlight the $WPB$'s impact on the MOEAs' performance, the obtained test problems do not contain other challenging problem characteristics.
    \item The performance of MOEAs is comprehensively evaluated using new test problems obtained by the two generators (Section~\ref{sec:experiment}). The experimental results validate the theoretical findings in Section~\ref{sec:analysis}, and show that the size and spatial relation of the $WPB$ can also significantly degrade the performance of MOEAs. Overall, no single algorithm can perform well on all scenarios.
\end{itemize}

The remainder of this paper is organized as follows. Section~\ref{sec:background} introduces basic concepts in multi-objective optimization and defines the $WPB$ and its categories. Section~\ref{sec:analysis} establishes the DDR and analyzes this metric for various $WPB$s. Section~\ref{sec:problem} presents two novel test problem generators. Sequentially, Section~\ref{sec:experiment} conducts experimental studies utilizing the test problem generators. Finally, Section~\ref{sec:conclusion} summarizes this paper and discusses future directions. A summary of symbols in this paper is listed in Table~\ref{tab:notation} for ease of reference. 

\begin{table}[ht]
  \centering
  \caption{Notation used in the paper.}
    \setlength{\tabcolsep}{2mm}{
    % \begin{tabular}{cl}
    % \begin{threeparttable}
    \begin{tabular}{cp{52ex}}
    \toprule
    Symbol & \multicolumn{1}{c}{Description} \\
    \midrule
    $m, n$ & The number of objectives and variables, respectively. \\
    $\mathbf{x}, \mathbf{z}$ & Solution and objective vector. \\
    $\Omega, Z$ & Feasible region and feasible objective region. \\
    $PF, PS$ & Pareto front and Pareto set. See Definition~\ref{def:pf}. \\
    $WPF$ & Weak Pareto front. See Definition~\ref{def:wpf}. \\
    $\mathbf{z}^{nad}, \mathbf{z}^{ide}$ & The nadir and ideal objective vectors, respectively. See Definition~\ref{def:ideal_nadir}. \\
    $WPB$ & Weak Pareto boundary. See Definition~\ref{def:wpb}. \\
    $\nu$ & The maximum number of objectives that cannot be improved without degrading other objectives. $\nu\in[m-1]$, where $[m-1]$ is short for $\{1, \ldots, m-1\}$. \\
    $WPB_{\nu,i}$ & A $WPB$ in the category of $WPB$s parameterized by $\nu$. See Definition~\ref{def:wpb_nu_i}. \\
    $I_{\nu,i}$ & The $i$-th combination in $\dbinom{[m]}{\nu}$. $\overline{I}_{\nu,i} = [m] \setminus I_{\nu,i}$. \\
    $\mathbf{r}$ & DRS in the objective space. \\
    $\Delta$ & The normal distance between the $WPB_{\nu,i}$ and $\mathbf{r}$. \\
    \bottomrule
    \end{tabular}%
    % \begin{tablenotes}
    %     \item[] The meanings of $i,j,k,l$ depend on context.
    % \end{tablenotes}
    % \end{threeparttable}
    }
  \label{tab:notation}%
\end{table}%

% \section{Background}
\section{Preliminaries}\label{sec:background}

\subsection{Basic Concepts in Multi-Objective Optimization}\label{ssec:mo_concepts}
An MOP can be written as
\begin{equation}\label{eqn:MOP}
    \begin{aligned}
        \mbox{min.}\quad & \mathbf{f}(\mathbf{x}) = (f_1(\mathbf{x}),\ldots,f_m(\mathbf{x}))^{\intercal}, \\
        \mbox{s.t.}\quad & \mathbf{x} \in \Omega,
    \end{aligned}
\end{equation}
where $\mathbf{x} = (x_1,\ldots, x_n)^{\intercal}$ is the solution, and $\Omega \subset \mathbb{R}^n$ denotes the feasible region. $\mathbf{f}: \mathbb{R}^n \rightarrow \mathbb{R}^m$ is composed of $m$ objective functions, and $\mathbf{z}=\mathbf{f}(\mathbf{x})$ is the objective vector corresponding to $\mathbf{x}$. The image of the feasible region in the objective space is referred to as the feasible objective region denoted as $Z$.

Some basic concepts are introduced as follows~\cite{miettinen1998nonlinear,tu2024random}:
\begin{definition}\label{def:dominate}
Given two vectors $\mathbf{u},\mathbf{v}\in\mathbb{R}^m$, $\mathbf{u}$ is said to \textbf{\em dominate} $\mathbf{v}$ (denoted as $\mathbf{u}\prec\mathbf{v}$), if and only if $u_i \leq v_i$ for every $i\in [m]$ and $u_{j}<v_{j}$ for at least one $j\in [m]$.
\end{definition}

\begin{definition}\label{def:PO}
A decision vector $\mathbf{x}^{*}$ and the corresponding objective vector $\mathbf{f}(\mathbf{x}^{*})$ are \textbf{\em Pareto-optimal}, if there is no $\mathbf{x}\in\Omega$ such that $\mathbf{f}(\mathbf{x})$ dominates $\mathbf{f}(\mathbf{x}^{*})$ according to Definition~\ref{def:dominate}.
\end{definition}

\begin{definition}\label{def:weaklyPO}
A decision vector $\mathbf{x}^{\prime}\in\Omega$ and the corresponding objective vector $\mathbf{f}(\mathbf{x}^{\prime})$ are \textbf{\em weakly Pareto-optimal}, if there does not exist another decision vector $\mathbf{x}\in\Omega$ such that $f_i(\mathbf{x})< f_i(\mathbf{x}^{\prime})$ for all $i \in [m]$.
\end{definition}

\begin{definition}\label{def:pf}
The set of all Pareto-optimal solutions is called the \textbf{\em Pareto set} (denoted as $PS$), and its image in the objective space is called the \textbf{\em Pareto front} (denoted as $PF$).
% and the set of all Pareto-optimal objective vectors is called the \textbf{\em Pareto front} (denoted as $PF$).
\end{definition}

\begin{definition}\label{def:wpf}
% [From~\cite{tu2024random}]
The set of all weakly Pareto-optimal objective vectors is called the \textbf{\em weak Pareto front} (denoted as $WPF$).
\end{definition}

\begin{definition}\label{def:ideal_nadir}
The \textbf{\em ideal objective vector} $\mathbf{z}^{ide}$ is composed of the lower bounds of the $PF$, i.e., $z^{ide}_i = \mathop{\min}_{\mathbf{x}\in PS}{f_{i}}(\mathbf{x})$ for $i \in [m]$. The \textbf{\em nadir objective vector} $\mathbf{z}^{nad}$ consists of the upper bounds of the $PF$, \ie, $z^{nad}_i = \mathop{\max}_{\mathbf{x}\in PS}{f_{i}}(\mathbf{x})$ for $i \in [m]$. 
\end{definition}

\begin{figure}[t]
\centering
\subfloat[Continuous $PF$]{\label{fig:illu_WPB_2D_1}\includegraphics[width=0.5\linewidth]{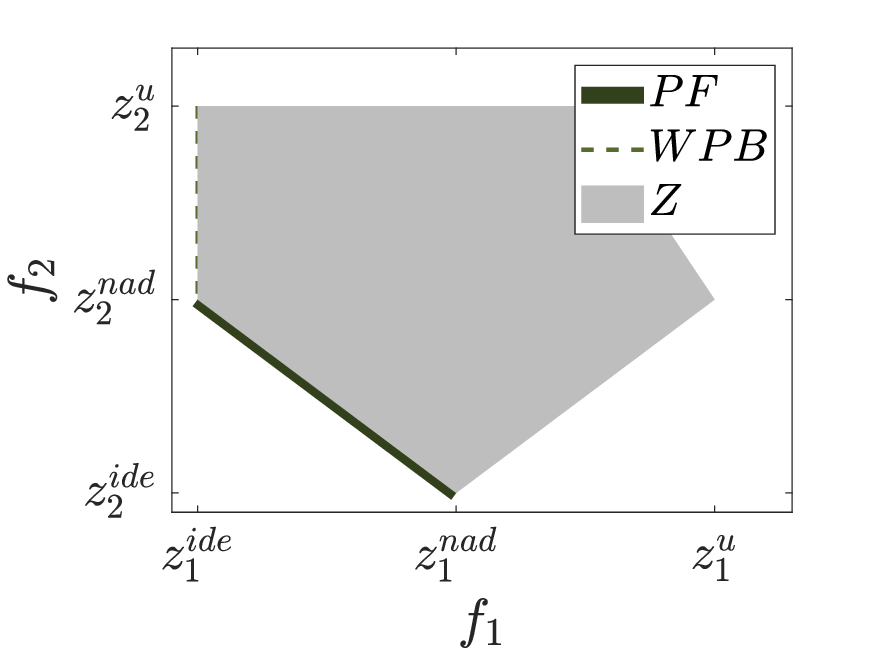}}
\hfil
\subfloat[Discontinuous $PF$]{\label{fig:illu_WPB_2D_2}\includegraphics[width=0.5\linewidth]{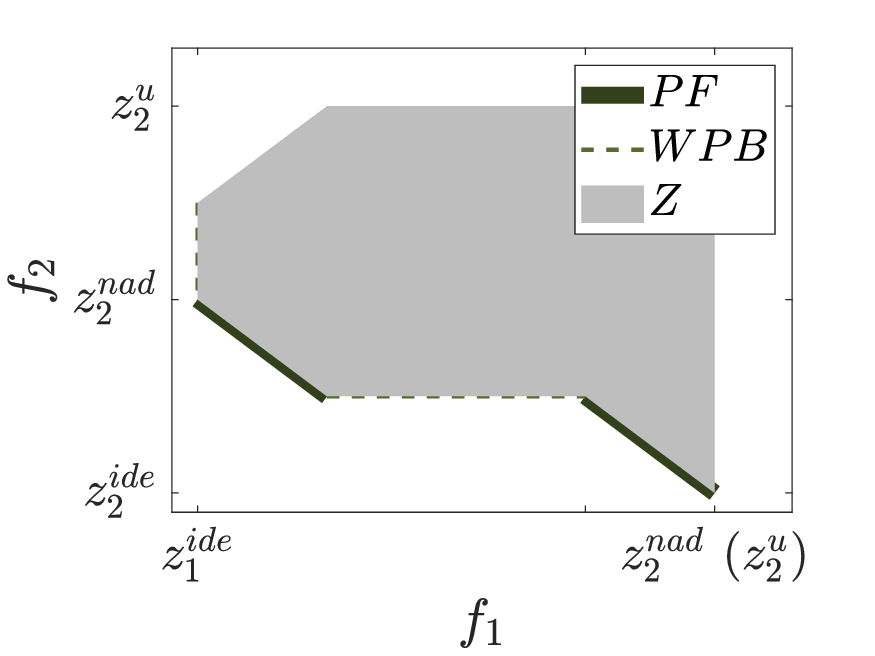}}
\caption{Examples of the $WPB$ in the 2-objective case.}
\label{fig:illu_WPB_2D}
\end{figure}

\subsection{Weak Pareto Boundary}\label{ssec:wpb}
Building upon the Definitions~\ref{def:PO} and~\ref{def:weaklyPO}, the $WPB$ is formally defined in Definition~\ref{def:wpb}.

\begin{definition}\label{def:wpb}
A set is called the \textbf{\em weak Pareto boundary} (denoted as $WPB$) if each objective vector in the set is weakly Pareto-optimal but not Pareto-optimal.
\end{definition}

Let $\mathbf{z}^\prime$ be an objective vector of the $WPB$. According to Definition~\ref{def:weaklyPO}, there exists an objective vector $\mathbf{z}$ such that $z_i<z_i^\prime$ for some objective(s) and $z_j=z_j^\prime$ for the remaining objective(s). That is, $\mathbf{z}^\prime$ has some objectives that can be improved without causing a deterioration in others.
Two 2-objective examples are illustrated in \figurename~\ref{fig:illu_WPB_2D}. The solid and dashed lines represent the $PF$ and the $WPB$, respectively. In \figurename~\ref{fig:illu_WPB_2D_1}, the objective vector of the $WPB$ attains the optimal value for $f_1$, while the value of $f_2$ can be further improved. Moreover, the $PF$ can be either a connected set or a disconnected set. In \figurename~\ref{fig:illu_WPB_2D_2}, a $WPB$ is located within the discontinuous region of the $PF$. The objective vector on the $WPB$ does not achieve the optimal value for any objective. Nevertheless, improving the value of $f_1$ does not necessitate a corresponding degradation in the value of $f_2$.

\begin{figure}[ht]
\centering
\subfloat[Continuous $PF$]{\label{fig:illu_WPB_subdiv_3D_1}\includegraphics[width=0.5\linewidth]{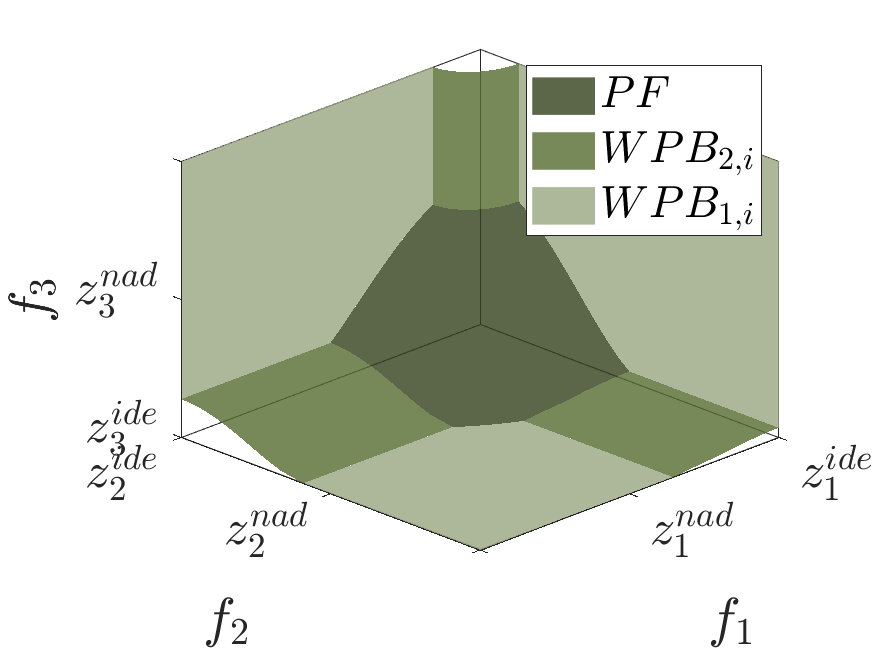}}
\hfil
\subfloat[Discontinuous $PF$]{\label{fig:illu_WPB_subdiv_3D_2}\includegraphics[width=0.5\linewidth]{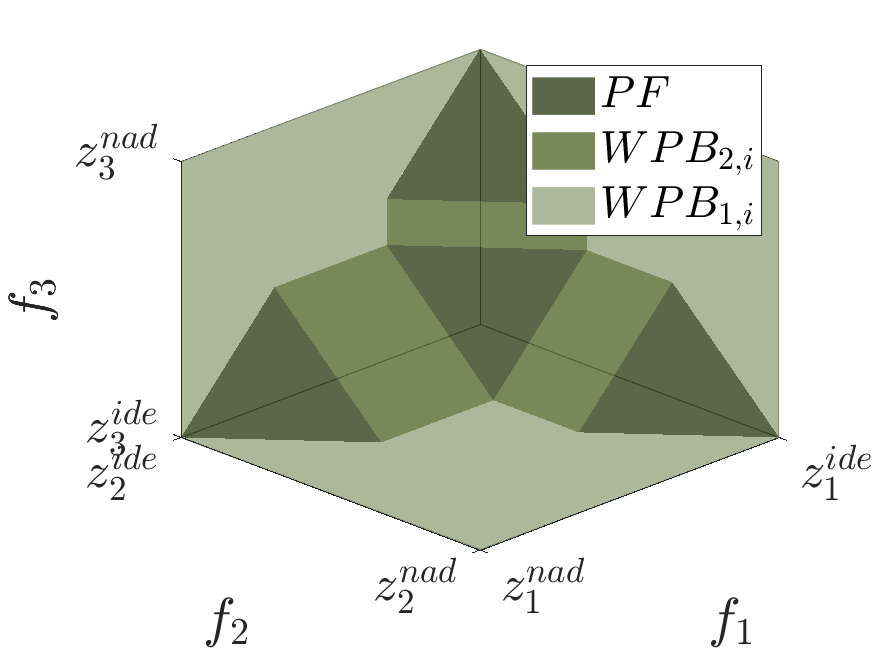}}
\caption{Examples of the $WPB_{\nu,i}$ in the 3-objective case.}
\label{fig:illu_WPB_subdiv_3D}
\end{figure}

The $WPB$ can be divided into different categories. For an objective vector of the $WPB$, $\nu \in [m-1]$ is defined as the maximum number of objectives that cannot be improved without degrading other objectives. 
% \rh{If $\nu \geq 2$, $\nu$ objective values in such an objective vector achieve Pareto optimality.仍然是错误的} 
According to $\nu$, the $WPB$ can be divided. Furthermore, for a given value of $\nu$, there exist $C_m^\nu$ distinct combinations of $\nu$ objectives chosen from $m$ objectives, where $C_m^\nu$ represents the binomial coefficient. 
% We denote ``MOP$_{\nu,i}$'' as an MOP with $\nu$ objectives selected from the $m$ objectives of the original MOP.
Let $I_{\nu,i}$ be the $i$-th combination in $\dbinom{[m]}{\nu}$ (\eg, $I_{\nu,i}$ can be $\{1,2\}$, $\{1,3\}$, or $\{2,3\}$, if $m=3$ and $\nu=2$) and $\overline{I}_{\nu,i}$ represent $[m] \setminus I_{\nu,i}$.
% $\left(f_j(\mathbf{x})\right)_{j \in I_{\nu,i}}$ be the objective vector of the MOP$_{\nu,i}$.
A $WPB$ denoted as $WPB_{\nu,i}$ is defined in Definition~\ref{def:wpb_nu_i}. In other words, the objective vector of the $WPB_{\nu,i}$ satisfies: the objective with respect to $I_{\nu,i}$ cannot be further improved without compromising the other objectives, whereas the objective with respect to $\overline{I}_{\nu,i}$ can be improved without compromising the other objectives. $WPB_{\nu,i}$ for $i=1,\ldots,C_m^\nu$ represent all members belonging to the category parameterized by $\nu$. Moreover, Properties~\ref{prop:union}--\ref{prop:nu_1_plane} can be intuitively derived from Definition~\ref{def:wpb_nu_i}.

\begin{definition}\label{def:wpb_nu_i}
Each objective vector of the $WPB_{\nu,i}$, denoted as $\mathbf{z}^\prime$, satisfies the following conditions:
\begin{itemize}
    \item $\mathbf{z}^\prime \in Z$;
    \item $\forall j \in I_{\nu,i},~ \forall k \in [m]\setminus\{j\},~ \nexists \mathbf{z} \in Z: ~ z_j^\prime < z_j,~ z_k^\prime \leq z_k$;
    \item $\forall j \in \overline{I}_{\nu,i},~ \forall k \in [m]\setminus\{j\},~ \exists \mathbf{z} \in Z: ~ z_j^\prime < z_j,~ z_k^\prime \leq z_k$.
\end{itemize}
\end{definition}

\begin{property}\label{prop:union}
% \begin{equation}
%     % WPB = \bigcup_{\nu=1}^{m-1}\bigcup_{i=1}^{\frac{m!}{\nu!(m-\nu)!}} WPB_{\nu,i}
%     WPB = \bigcup_{\nu=1}^{m-1}\bigcup_{i=1}^{C_m^\nu} WPB_{\nu,i}.
% \end{equation}
$\bigcup_{\nu=1}^{m-1}\bigcup_{i=1}^{C_m^\nu} WPB_{\nu,i} = WPF \setminus PF$.
\end{property}

\begin{property}\label{prop:intersection}
% 面的相交处，一定是属于nu大的
$WPB_{\nu_1,i} \bigcap WPB_{\nu_2,j} = \emptyset$ if and only if $\nu_1 \neq \nu_2$ or $i \neq j$.
\end{property}

% \rh{补充形状、尺寸可以灵活变化}

% 关于大小或尺寸的性质
\begin{property}\label{prop:empty}
% 不一定是真子集
% $WPB_{\nu,i} \subseteq WPB$.
For some MOPs, $\exists \nu,i$ such that $ WPB_{\nu,i}=\emptyset$.
\end{property}

% \begin{IEEEproof}[Proof Sketch]
% \end{IEEEproof} 

% 关于形状的性质
\begin{property}\label{prop:nu_1_plane}
If $WPB_{1,i} \neq \emptyset$ for $i \in [m]$, then the objective vectors of the $WPB_{1,i}$ must lie in the hyperplane perpendicular to a specific axis.
% defined by $z_i = \min_{\mathbf{x}\in\Omega} f_i(\mathbf{x})$.
\end{property}

In addition to the category, the $WPB_{\nu,i}$ possesses other attributes, including shape, size, and spatial relationship to the $PF$. \figurename~\ref{fig:illu_WPB_subdiv_3D} shows two 3-objective examples with distinct $PF$s. $WPB_{2,i}$ for $i=1,2,3$ have various shapes and sizes in \figurename~\ref{fig:illu_WPB_subdiv_3D_1}. $WPB_{1,i}$ for $i=1,2,3$ are linear in the examples, as stated in Property~\ref{prop:nu_1_plane}. It is important to note that the objective vector on the $WPB_{\nu,i}$ do not necessarily achieve Pareto optimality with respect to the objectives indexed by $I_{\nu,i}$, as indicated by the $WPB_{2,i}$ in \figurename~\ref{fig:illu_WPB_subdiv_3D_2}. Thus, Definition~\ref{def:wpb_nu_i} considers a more general scenario than the definition in~\cite{wang2024multi}. Compared to the $WPB_{2,i}$ in \figurename~\ref{fig:illu_WPB_subdiv_3D_1}, the one in \figurename~\ref{fig:illu_WPB_subdiv_3D_2} exhibits a distinct spatial relation to the $PF$. 

\section{Quantization of Dominance Resistance}\label{sec:analysis}
\begin{figure}[ht]
\centering
\includegraphics[width=0.32\textwidth]{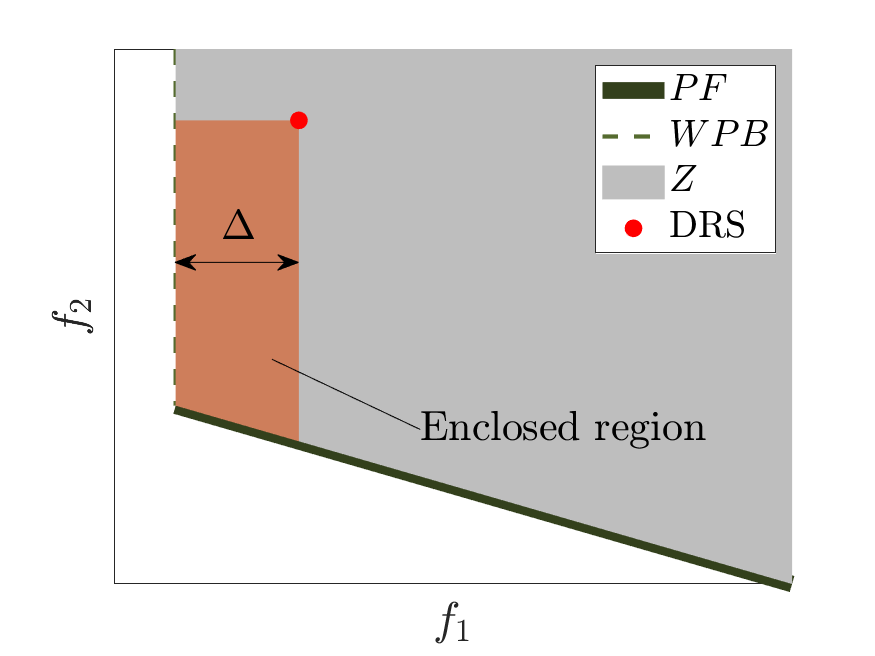}
\caption{Illustration of a DRS and its enclosed region in the 2-objective case.}
\label{fig:illustration_DRS_2D}
\end{figure}

The $WPB$ often misleads MOEAs into finding many DRSs, since the objective vectors close to the $WPB$ are usually regarded as the DRSs\footnote{Without causing confusion, we refer to ``the corresponding objective vector of a DRS'' or ``a DRS in the objective space'' simply as ``a DRS'' in this paper.}. As illustrated in \figurename~\ref{fig:illustration_DRS_2D}, an objective vector close to the $WPB$ can demonstrate small deviations from Pareto-optimal objective vectors in some dimensions while exhibiting significant differences in others. The superiority of this objective vector on some objectives makes it difficult to eliminate, and therefore, become a DRS. Once the DRSs induced by the $WPB$ are generated, the MOEA may maintain most of these DRSs and conduct reproduction based on them, which in turn more likely generates new DRSs nearby and misses the Pareto-optimal solutions. As a result, the performance of the MOEA deteriorates.

To enable a quantitative analysis of dominance resistance, we define the degree of dominance resistance (DDR) for a feasible solution as the reciprocal probability of sampling an objective vector that dominates its associated objective vector. A solution characterized by a high DDR is difficult to eliminate once it is included in the population. 
This section investigates the DDR of a DRS in two scenarios: DRS on the $WPB$; DRS not on the $WPB$. For the latter case, let $\mathbf{r}$ be an objective vector corresponding to a DRS. We have $\mathbf{r}\in Z$. Moving $\mathbf{r}$ toward the $WPB_{\nu,i}$ along its normal vector reduces the $\nu$ elements of $\mathbf{r}$, while leaving the remaining $m-\nu$ elements unchanged. The distance between the $WPB_{\nu,i}$ and $\mathbf{r}$ is represented by $\Delta$. Different categories (\ie, values of $\nu$) and shapes are considered for the $WPB_{\nu,i}$.
The analysis leaves out the intricate process of generating a feasible solution. For this, we suppose that:
\begin{itemize}
    \item The resulting distribution for sampling a feasible objective vector is a uniform distribution over the feasible objective region.
    % \item The region where the feasible objective vector dominates a given objective vector $\mathbf{r}$ is entirely bounded by a part of the $WPF$ and $m$ planes defined by $f_i(\mathbf{x}) = r_i$ for $i=1,\ldots,m$.
    % the region内部不能包含不可行区域
    \item The feasible objective subregion, within which the objective vector dominates $\mathbf{r}$, is fully enclosed by a part of the $WPF$ and $m$ planes characterized by $f_i(\mathbf{x}) = r_i$ for $i = 1, \ldots, m$. \figurename~\ref{fig:illustration_DRS_2D} depicts an example of the enclosed region in the 2-objective case.
\end{itemize}
Under these assumptions, the hypervolume~\cite{zitzler1999multiobjective} enclosed by the $PF$ and $\mathbf{r}$ (denoted as $\mathcal{H}(PF,\mathbf{r})$) can indicate the DDR of $\mathbf{r}$ (denoted as $DDR(\mathbf{r})$). That is, $\mathcal{H}(PF,\mathbf{r}) = \mathcal{H}(WPF,\mathbf{r}) \propto \frac{1}{DDR(\mathbf{r})}$, indicating a larger volume means a smaller DDR.

\subsection{DRS on the \texorpdfstring{$WPB$}{WPB}}
Any low-dimensional geometric object (such as surfaces, curves, or point sets in $\mathbb{R}^m$) has a high-dimensional Lebesgue measure of zero. Therefore, the probability of a generated objective vector lying on the $WPB$ (\ie, low-dimensional manifold) is zero, regardless of the $m$-dimensional random sampling method used. Formally, 
\begin{equation}
    \mathrm{Pr}(\mathbf{z} \in WPB) = \frac{\mu(WPB)}{\mu(Z)} = 0,
\end{equation}
where $\mathbf{z}$ is a randomly generated feasible objective vector and $\mu(\cdot)$ represents the Lebesgue measure. As a result, the DDR of the DRS on the $WPB$ is infinite, indicating that eliminating the DRS on the $WPB$ is almost impossible.

\begin{figure*}[ht]
\centering
\subfloat[]{\label{fig:cs_p1_legend}\includegraphics[width=0.32\linewidth]{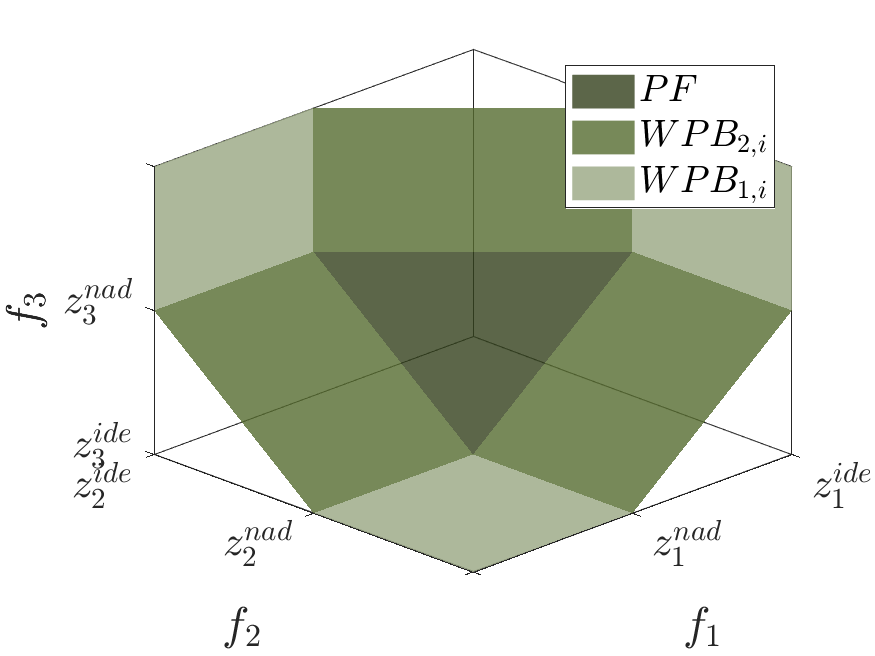}}
\hfil
\subfloat[]{
\label{fig:cs_p1_move}
\includegraphics[width=0.32\linewidth]{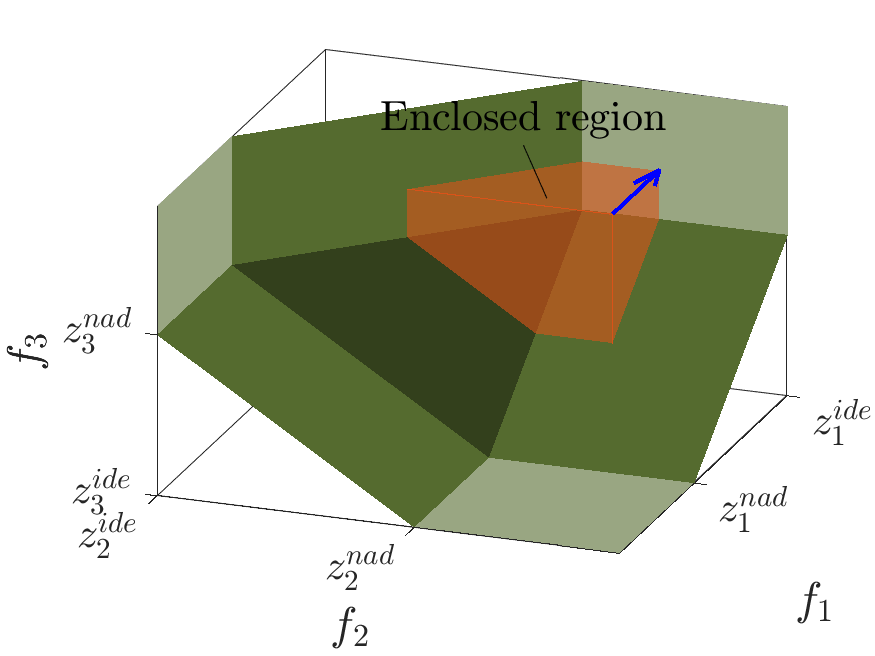}
\includegraphics[width=0.32\linewidth]{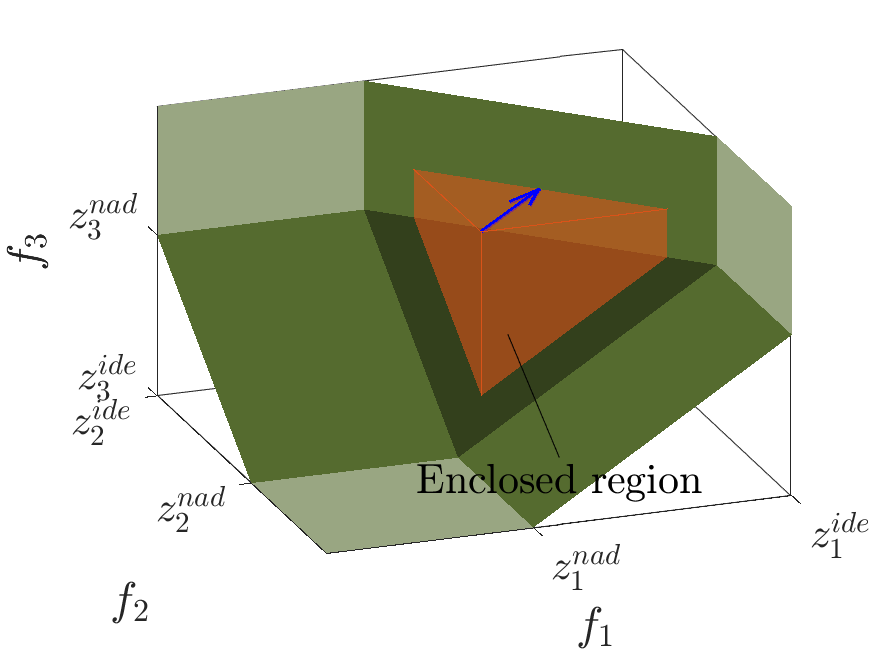}
}
% \hfil
% \subfloat[]{\includegraphics[width=2.3in]{figs/case_studies/case_studies_p1_nu2.eps}}
\caption{(a) Illustration of the $PF$ and all $WPB$s in the case with 3 objectives and $p=1$. (b) Visualization of moving $\mathbf{r}$ toward the $WPB_{1,i}$ and the $WPB_{2,i}$. For better clarity, $\Delta$ is set to a relatively large value of 0.5 in (b) and (c).}
\label{fig:cs_p1}
\end{figure*}

\begin{figure*}[ht]
\centering
\subfloat[$m=3$]{\label{fig:vol_p1_m3_13}\includegraphics[width=0.32\linewidth]{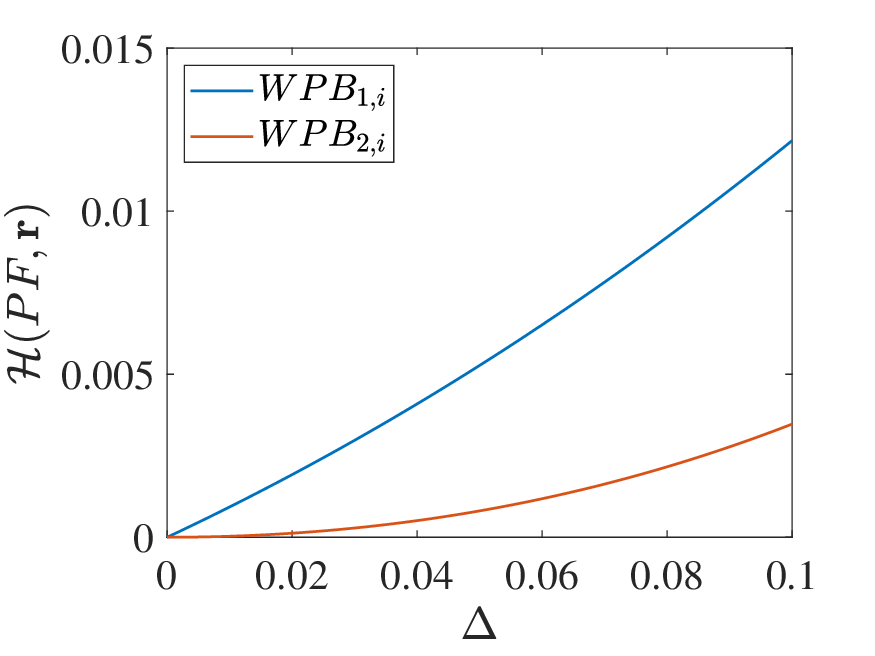}}
\hfil
\subfloat[$m=4$]{\includegraphics[width=0.32\linewidth]{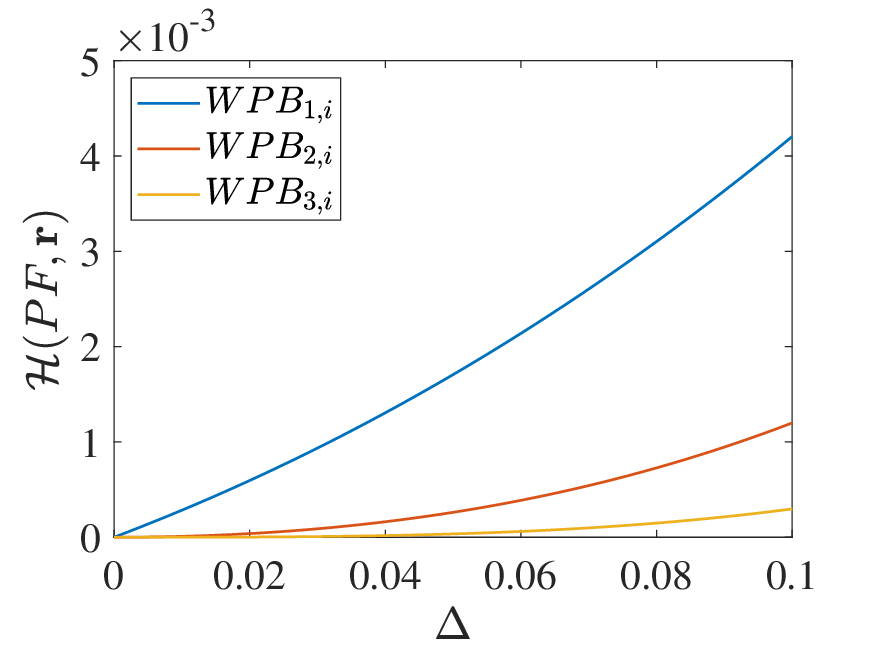}}
\hfil
\subfloat[$m=5$]{\includegraphics[width=0.32\linewidth]{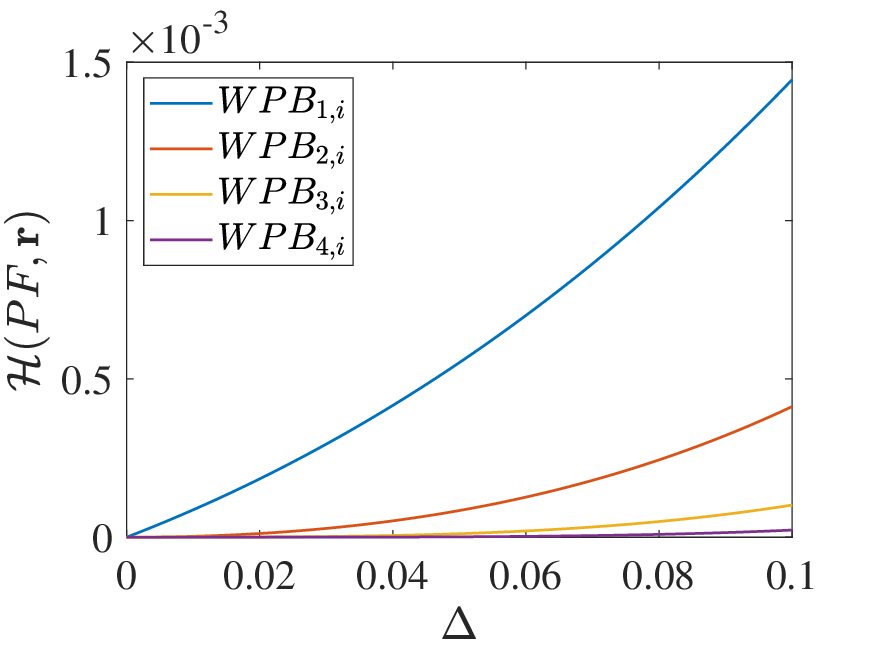}}
\caption{Curves of $\mathcal{H}(PF,\mathbf{r})$ with respect to $\Delta$ for the case where $p=1$ and $r_j=1.3$ for every $j \in \overline{I}_{2,i}$.
% (\ie, the enclosed volume) (\ie, the distance from the $WPB_{\nu,i}$ to $\mathbf{r}$)
}
\label{fig:vol_p1}
\end{figure*}

\subsection{DRS Not on the \texorpdfstring{$WPB$}{WPB}: Case Studies}\label{ssec:wpb_cs}
\begin{figure*}[ht]
\centering
\subfloat[]{\label{fig:cs_p2_legend}\includegraphics[width=0.32\linewidth]{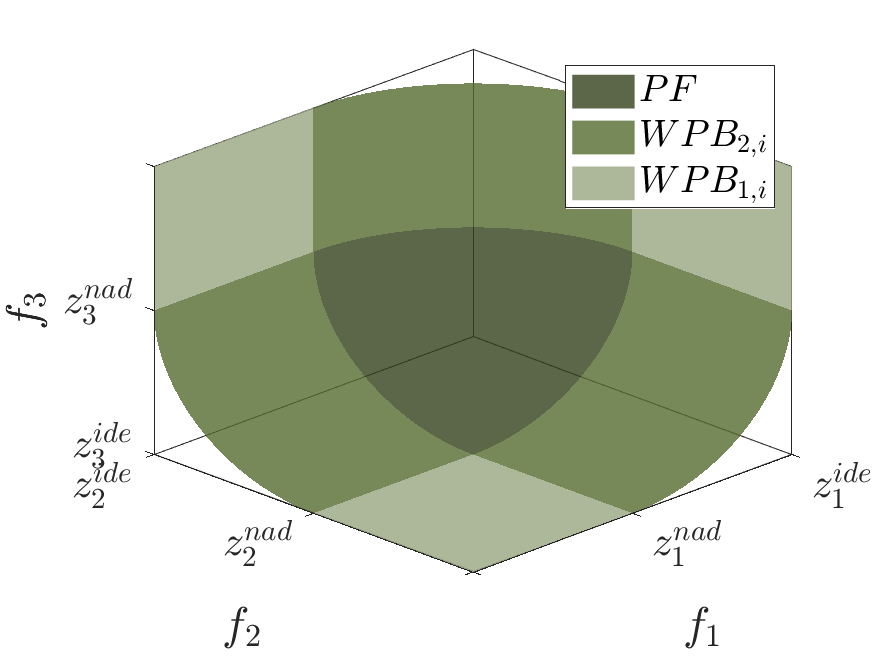}}
\hfil
\subfloat[]{\includegraphics[width=0.32\linewidth]{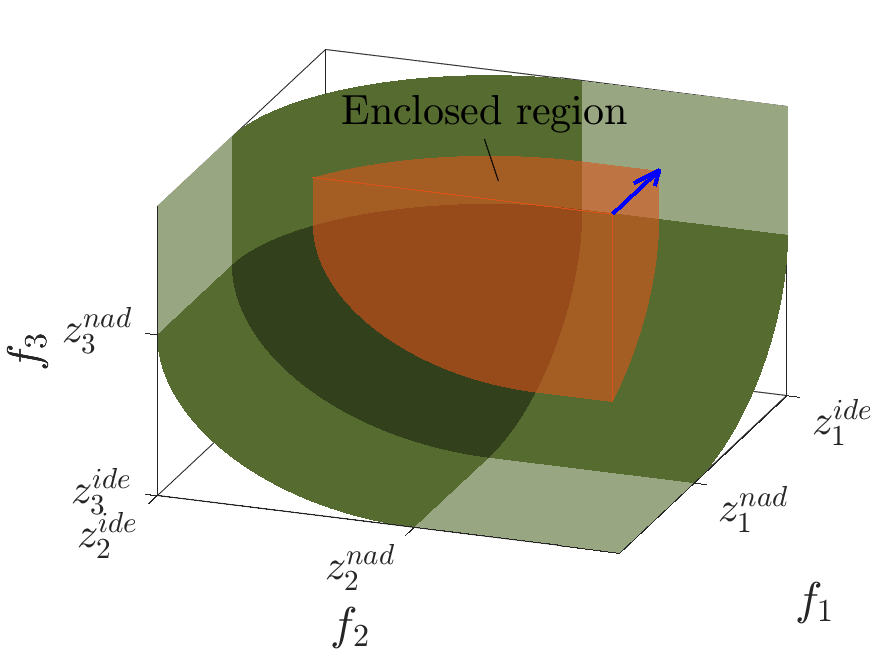}}
\hfil
\subfloat[]{\label{fig:cs_p2_nu2}\includegraphics[width=0.32\linewidth]{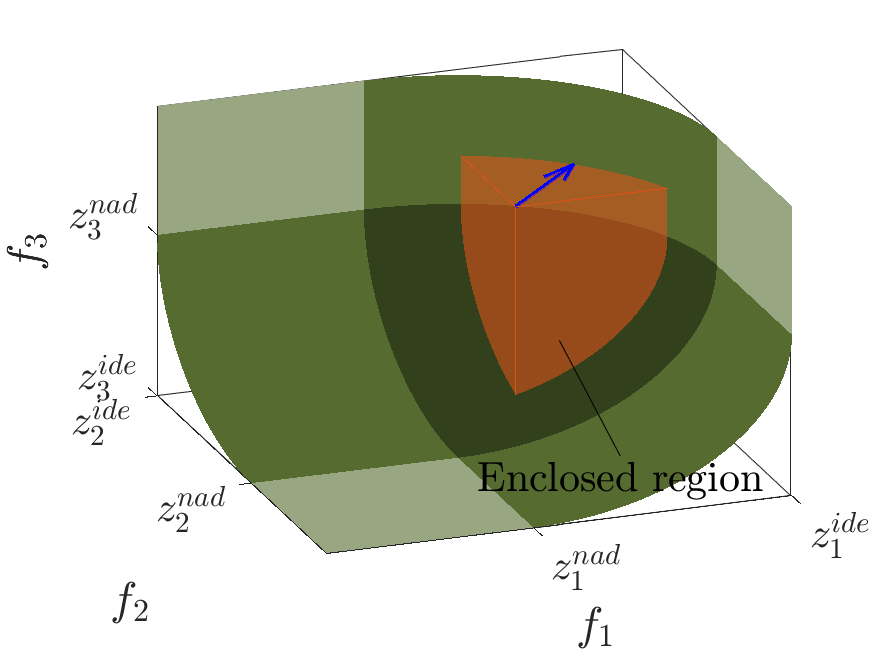}}
\caption{(a) Illustration of the $PF$ and all $WPB$s in the case with 3 objectives and $p=2$. (b) Visualization of moving $\mathbf{r}$ toward the $WPB_{1,i}$. (c) Visualization of moving $\mathbf{r}$ toward the $WPB_{2,i}$. For better clarity, $\Delta$ is set to a relatively large value of 0.5 in (b) and (c).}
\label{fig:cs_p2}
\end{figure*}

To provide intuitive insights, the DDR of the DRS near the $WPB$ is investigated through several case studies. Three specific cases are present, whose $PF$s are derived from the following surface equation
\begin{equation}\label{eqn:cs_pf}
    \sum_{i=1}^m \left(1-\frac{z_i-z_i^{ide}}{z_i^{nad}-z_i^{ide}}\right)^p = 1,
\end{equation}
where $0 \leq z_i \leq 1$ and $p>0$. $\mathbf{z}=(z_1,\ldots,z_m)^\intercal$ denotes the objective vector and $p$ is a control parameter. The objective vector associated with the optimal value of each objective is unique. These $m$ extreme objective vectors are denoted as
\begin{equation}\label{eqn:ext_obj}
    z_j^{ext(i)}=
    \begin{cases}
        z_j^{nad}, & j \neq i, \\
        z_j^{ide}, & j = i,
    \end{cases}
    \quad i=1,\ldots,m.
\end{equation}
The boundaries of the $PF$ are extended.
% in directions parallel to the axis vectors. \rh{高维时不是这样}
Then, the $WPB_{\nu,i}$ is constructed, which is described as
\begin{equation}
    \sum_{j\in I_{\nu,i}} \left(1-\frac{z_j-z_j^{ide}}{z_j^{nad}-z_j^{ide}}\right)^p = 1,
\end{equation}
where $I_{\nu,i}\subset[m]$.
$p$ can adjust the shapes of the $PF$ and the $WPB_{\nu,i}$. Specifically, the $PF$ and the $WPB_{\nu,i}$ with $\nu \geq 2$ are linear (\eg, \figurename~\ref{fig:cs_p1_legend}), convex (\eg, \figurename~\ref{fig:cs_p2_legend}), or concave (\eg, \figurename~\ref{fig:cs_p05_legend}) when $p = 1$, $p > 1$, or $0 < p < 1$.

\subsubsection{\texorpdfstring{$p=1$}{p=1}}\label{sssec:peq1}
The $PF$ and any $WPB_{\nu,i}$ are hyperplanes in this case. A half-space representation (\ie, a system of inequalities) of the enclosed region is formulated as
\begin{equation}\label{eqn:sys_er}
\begin{cases}
    \sum\limits_{j=1}^m \left(1-\dfrac{z_j-z_j^{ide}}{z_j^{nad}-z_j^{ide}}\right) \geq 1, & \\
    \sum\limits_{j\in I_{\nu,i}} \left(1-\dfrac{z_j-z_j^{ide}}{z_j^{nad}-z_j^{ide}}\right) \geq 1, & I_{\nu,i} \in \bigcup\limits_{k=1}^{m-1} \dbinom{[m]}{k}, \\
    z_j \leq r_j, & j = 1,\ldots,m.
\end{cases}
\end{equation}
The enclosed region is a convex polyhedron. The computation of the volume of a convex polyhedron requires several sequential steps, as no universal formula is available for this purpose. One commonly employed pipeline involves dividing the polyhedron into several simplices using Delaunay triangulation, and then the volume is determined by summing the volumes of these individual simplices.

\begin{table}[t]
  \centering
  \caption{
  Estimated polynomial coefficients from Lasso regression for the curves in \figurename~\ref{fig:vol_p1}.
  % the enclosed volume versus the distance in \figurename~\ref{fig:vol_p1}.
  }
    \begin{threeparttable}
    \begin{tabular}{ccccccc}
    \toprule
    $m$     & $\nu$    & 1     & 2     & 3     & 4     & 5 \\
    \midrule
    \multirow{2}[2]{*}{3} & 1     & 8.72e-02 & 3.08e-01 & 0      & 0      & 0 \\
          & 2     & 0      & 2.95e-01 & 4.19e-01 & 0      & 0 \\
    \midrule
    \multirow{3}[2]{*}{4} & 1     & 2.55e-02 & 1.52e-01 & 0      & 0      & 0 \\
          & 2     & 0      & 8.72e-02 & 2.92e-01 & 0      & 0 \\
          & 3     & 0      & 0      & 2.56e-01 & 3.27e-01 & 0 \\
    \midrule
    \multirow{4}[2]{*}{5} & 1     & 7.30e-03 & 6.63e-02 & 0      & 0      & 0 \\
          & 2     & 0      & 2.54e-02 & 1.46e-01 & 0      & 0 \\
          & 3     & 0      & 0      & 7.58e-02 & 2.29e-01 & 0 \\
          & 4     & 0      & 0      & 0      & 2.00e-01 & 1.95e-01 \\
    \bottomrule
    \end{tabular}%
    \begin{tablenotes}
        % \item[] Blank cells indicate that coefficients are exactly zero.
        \item[] The coefficient of the 6-th degree term is zero.
    \end{tablenotes}
    \end{threeparttable}
  \label{tab:vol_poly_p1}%
\end{table}%

\begin{figure}[t]
\centering
\includegraphics[width=0.32\textwidth]{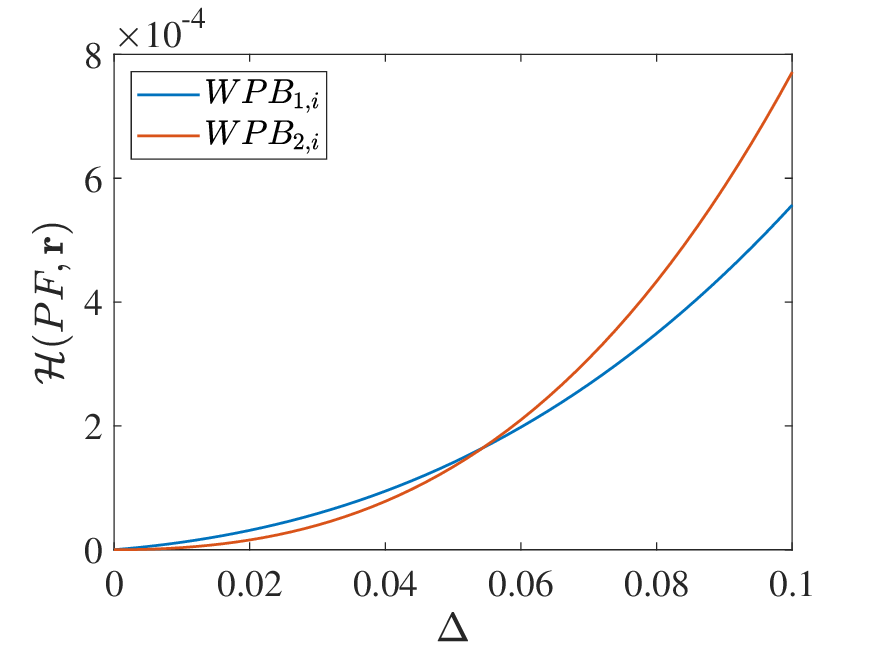}
\caption{Curves of $\mathcal{H}(PF,\mathbf{r})$ with respect to $\Delta$ for the case where $m=3$, $p=1$ and $r_j=1.03$ for every $j \in \overline{I}_{2,i}$. The estimated polynomials are $\left(1.77\times10^{-4}\right) \Delta + \left(4.56\times10^{-2}\right) \Delta^2 + \left(5.78\times10^{-2}\right) \Delta^3$ for the $WPB_{1,i}$ and $\left(2.89\times10^{-2}\right) \Delta^2 + \left(4.60\times10^{-1}\right)\Delta^3$ for the $WPB_{2,i}$.}
\label{fig:vol_p1_m3_103}
\end{figure}

Numerical experiments are performed in the following. \figurename~\ref{fig:cs_p1_move} displays $\mathbf{r}$ in the 3-objective case. We set $\mathbf{z}^{ide}=\mathbf{0}$, $\mathbf{z}^{nad}=\mathbf{1}$, $r_i = \frac{1}{m-1}+\frac{\Delta}{\sqrt{\nu}}$ for $i=1,\ldots,\nu$, and $r_i = 1.3$ for $i=\nu+1,\ldots,m$. The setting of $\mathbf{r}$ ensures that $\mathbf{r}$ lies on the $WPB_{\nu,i}$ when $\Delta=0$.
After configuration, the vertices of the polyhedron are determined. Specifically, we select all $m$-combinations from System~\eqref{eqn:sys_er}, convert the $m$ inequalities of each $m$-combination into $m$ equations, solve the resulting linear systems, and verify whether the intersection points (\ie, the solutions of systems) lie within the polyhedron. All valid intersection points represent all the vertices of the polyhedron. Finally, we can perform the Delaunay triangulation for the valid intersection points and compute the total volume of the resulting simplices.
The volume change curves are presented in \figurename~\ref{fig:vol_p1}. It is evident that the volume associated with a higher $\nu$ is smaller. Furthermore, polynomial fitting is employed for these curves, and Lasso regression is applied to select individual terms of the polynomial. Since the curve passes through the origin, the constant term can be omitted. The results are displayed in Table~\ref{tab:vol_poly_p1}, signifying that a smaller $\nu$ introduces lower-order terms in the volume-distance relationship.

\begin{figure}[t]
\centering
\subfloat[]{\label{fig:cs_p1_nu2_m2b_1}\includegraphics[width=0.5\linewidth]{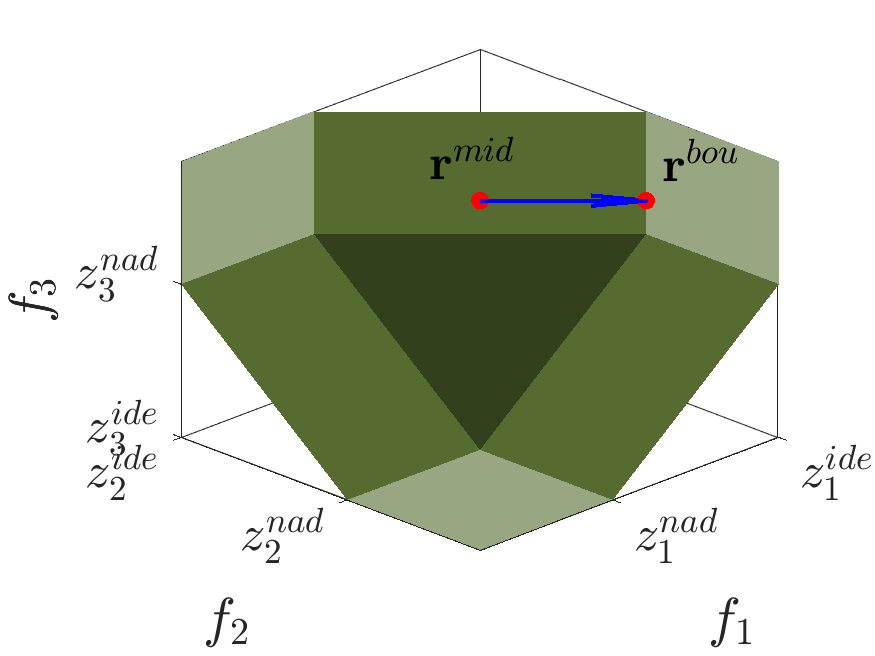}}
\hfil
\subfloat[]{\label{fig:cs_p1_nu2_m2b_2}\includegraphics[width=0.5\linewidth]{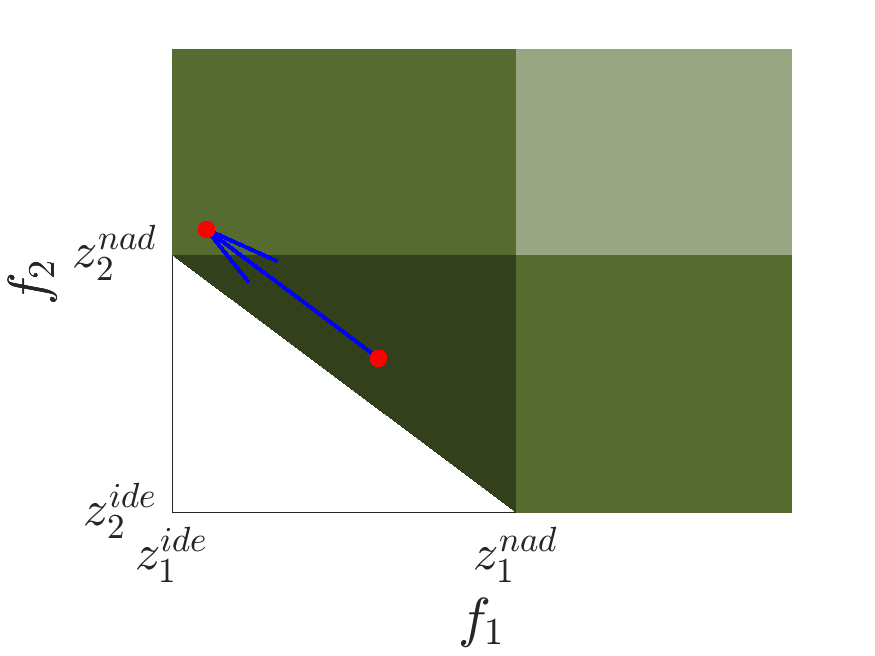}}
\caption{Illustration of moving the DRS from the middle (denoted as $\mathbf{r}^{mid}$) to the boundary (denoted as $\mathbf{r}^{bou}$) of the $WPB_{2,i}$ when $p=1$.}
\label{fig:cs_p1_nu2_m2b}
\end{figure}

\begin{figure}[t]
\centering
\includegraphics[width=0.32\textwidth]{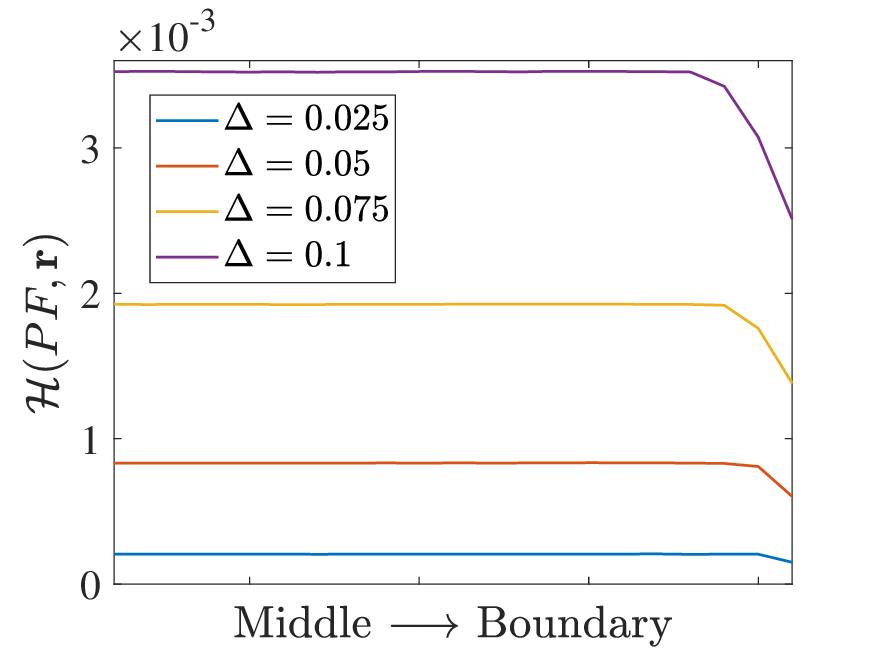}
\caption{Curves of $\mathcal{H}(PF,\mathbf{r})$ with respect to the relative position of $\mathbf{r}$ to the $WPB_{2,i}$ for the case where $m=3$, $p=1$ and $r_j=1.3$ for every $j \in \overline{I}_{2,i}$.}
\label{fig:vol_p1_m3_13_WPB2_m2b}
\end{figure}

Further investigations examine the effects of the DRS's relative position to the $WPB$. We modify $r_i$ to $1.03$ for $i=\nu+1,\ldots,m$, representing $\mathbf{r}$ is close to the $PF$. Comparing \figurename~\ref{fig:vol_p1_m3_103} with \figurename~\ref{fig:vol_p1_m3_13}, the enclosed volume decreases as the DRS approaches the $PF$. This result is straightforward and holds irrespective of the shapes of the $PF$ and the $WPB$. \figurename~\ref{fig:vol_p1_m3_103} also shows that the relative volume difference between the two curves becomes less pronounced, and the enclosed volume with respect to the $WPB_{2,i}$ is not consistently smaller. The intersection point of the two curves is approximately 0.54.
We also study the volume changes as the DRS transitions from the middle to the boundary of the $WPB_{2,i}$ while maintaining a fixed distance from the $PF$, as illustrated in \figurename~\ref{fig:cs_p1_nu2_m2b}. The results in \figurename~\ref{fig:vol_p1_m3_13_WPB2_m2b} indicate that no difference in enclosed volume is observed between $\mathbf{r}^{mid}$ and most other locations. The volume with respect to $\mathbf{r}^{bou}$ is evidently lower since the DRS is also close to the $WPB_{1,i}$. Nevertheless, the volume difference between $\mathbf{r}^{mid}$ and $\mathbf{r}^{bou}$ diminishes as $\Delta$ decreases.

To sum up, a higher $\nu$ results in a lower-order term in the volume-distance relationship. The DDR of a DRS is much higher as long as the DRS is sufficiently close to the $WPB_{\nu,i}$ with a larger $\nu$. As the disadvantaged objectives of DRSs improve (\ie, DRSs are closer to the $PF$), their relative differences in the DDR become less noticeable. 

\begin{figure*}[ht]
\centering
\subfloat[]{\label{fig:vol_p2_m3_13}\includegraphics[width=0.32\linewidth]{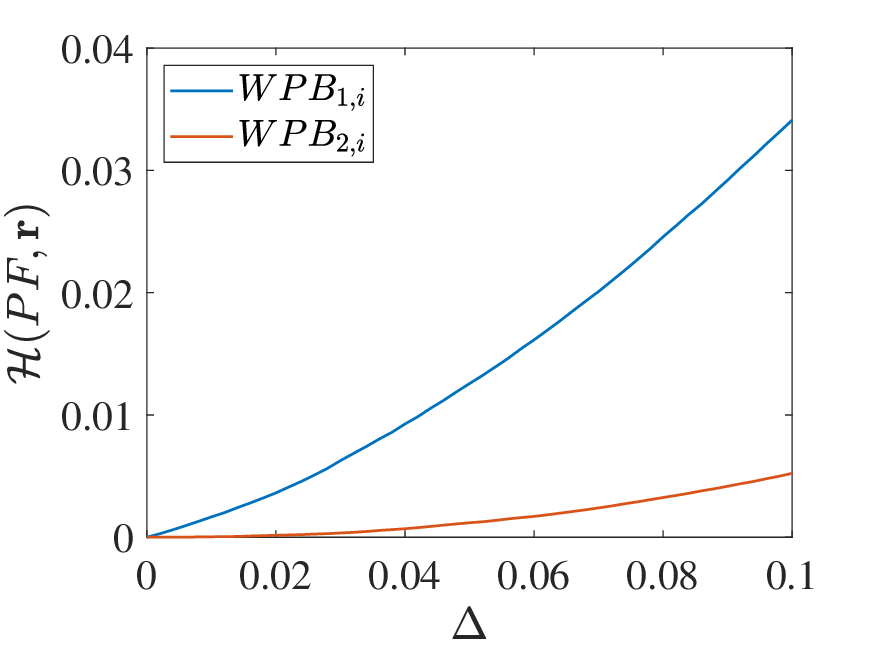}}
\hfil
\subfloat[]{\label{fig:vol_p2_m3_103}\includegraphics[width=0.32\linewidth]{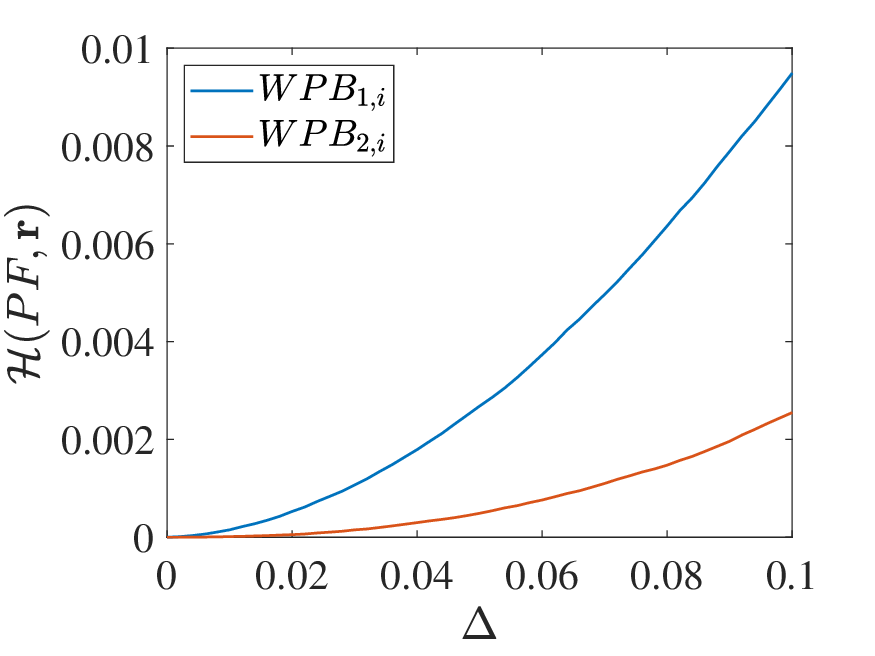}}
\hfil
\subfloat[]{\label{fig:vol_p2_m3_13_WPB2_m2b}\includegraphics[width=0.32\linewidth]{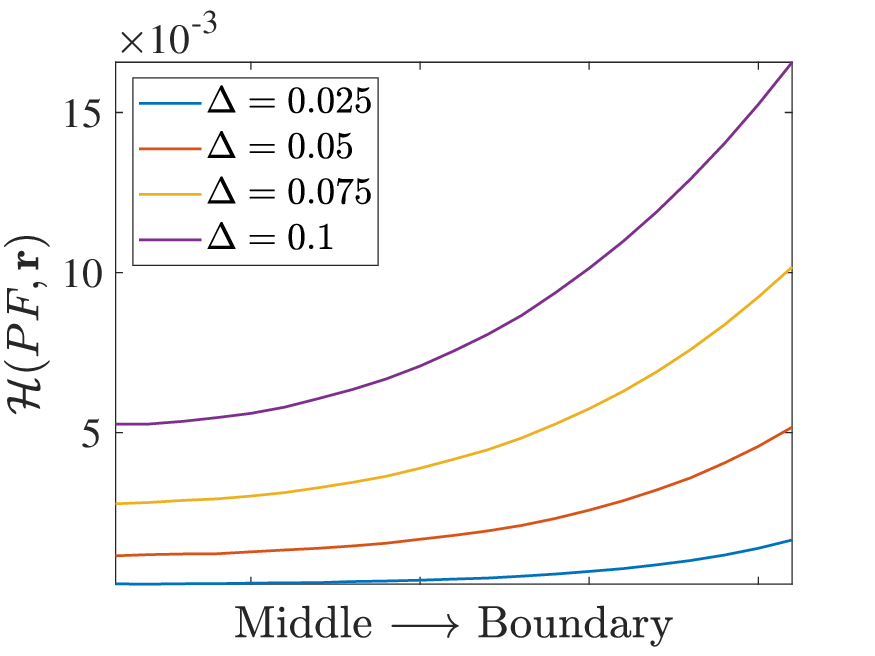}}
\caption{Curves of $\mathcal{H}(PF,\mathbf{r})$ with respect to $\Delta$ for the following cases: (a) $r_j=1.3$ for every $j \in \overline{I}_{2,i}$; (b) $r_j=1.03$ for every $j \in \overline{I}_{2,i}$. (c) Curves of $\mathcal{H}(PF,\mathbf{r})$ with respect to the relative position of $\mathbf{r}$ to the $WPB_{2,i}$ for the case where $r_j=1.3$ for every $j \in \overline{I}_{2,i}$. In these cases, $m=3$ and $p=2$.}
\label{fig:vol_p2}
\end{figure*}

\subsubsection{\texorpdfstring{$p \neq 1$}{p neq 1}}\label{sssec:pneq1}
Two extreme cases, $p\rightarrow\infty$ and $p\rightarrow0$, are examined first. In the two extreme cases, any $WPB_{\nu,i}$ with $\nu \geq 2$ is an ($m-\nu$)-dimensional object (\eg, a line when $m=3$ and $\nu=2$). 
% 满足等号的方程左侧：一个无穷接近1（但不是1）的数之和
When $p \rightarrow \infty$, there is only one Pareto-optimal objective vector, which is also the ideal objective vector. Therefore, the enclosed region is a hyperrectangle. The enclosed volume is straightforward, which is given by
\begin{equation}\label{eqn:vol_inf_p}
    \mathcal{H}(PF,\mathbf{r}) = \prod_{j\in [m]}\left(r_j-z_j^{ide}\right) \propto \prod_{j\in I_{\nu,i}}\left(r_j-z_j^{ide}\right).
\end{equation}
$r_{j} - z_j^{ide} \geq 0$ for $j=1,\ldots,m$ since $\mathbf{r} \in Z$. Consequently, $\mathcal{H}(PF,\mathbf{r})$ is $\Theta(\Delta^\nu)$ since $r_{j} - z_j^{ide} = \Theta(\Delta)$ for every $j \in I_{\nu,i}$.
% $O(\epsilon^\nu)$ when $r_{j} - z_j^{ide} = O(\epsilon)$ for every $j \in I_{\nu,i}$
$\mathcal{H}(PF,\mathbf{r})$ must reduce as $\nu$ increases, provided that $r_{j} - z_j^{ide}$ for every $j \in I_{\nu,i}$ is sufficiently small.

When $p \rightarrow 0$, the surface described by Eq.~\eqref{eqn:cs_pf} consists of $m$ hyperplanes characterized by $f_i(\mathbf{x}) = z_i^{nad}$ for $i = 1, \ldots, m$. Consequently, the $PF$ consists of the $m$ extreme objective vectors, which are defined in Eq.~\eqref{eqn:ext_obj}. The enclosed region has two possible configurations: if $r_i > z_i^{nad}$ for all $i=1,\ldots,m$, it is the union of $(m+1)$ disjoint hyperrectangles; otherwise, it is a single hyperrectangle. For the former case, let $D_i = \{\mathbf{z} | \mathbf{z}^{ext(i)} \prec \mathbf{z} \}$ for $i=1,\ldots,m$ and $D_0 = \{\mathbf{z} | \mathbf{z} \prec \mathbf{r}\}$. According to Eq.~\eqref{eqn:ext_obj}, $\left( D_k \cap D_l \right) = \{\mathbf{z} | \mathbf{z}^{nad}\prec\mathbf{z}\}$ for any $k, l \in [m]$ with $k \neq l$. Then, the enclosed region is $\left( \widetilde{D}_0 \cup \bigcup_{i=1}^m \widetilde{D}_i \right) \cap D_0$ where $\widetilde{D}_0 = \bigcap_{j=1}^m D_j = \{\mathbf{z} | \mathbf{z}^{nad}\prec\mathbf{z}\}$ and $\widetilde{D}_i = \left( D_i \setminus \bigcap_{j=1}^m D_j \right) = \left\{ \mathbf{z} \big| \left( z_i \leq z_i^{nad} \right) \wedge \left( \mathbf{z}^{ext(i)} \prec \mathbf{z} \right) \right\}$. For the latter case, the enclosed region is $\widetilde{D}_i \cap D_0$. Generally, the enclosed volume is
\begin{equation}\label{eqn:vol_small_p}
\begin{aligned}
    \mathcal{H}(PF,\mathbf{r})
    = & \sum_{k=1}^m \Bigg( \left(\min\left\{z_k^{nad},r_k\right\}-z_k^{ide}\right) \\
    & \prod_{j\in [m]\setminus\{k\}} \max\left\{r_j - z_j^{nad}, 0\right\}\Bigg) + \\
    & \prod_{j\in [m]} \max\left\{r_j - z_j^{nad}, 0\right\}.
\end{aligned}
\end{equation}
If $\mathbf{r}$ is close enough to the $WPB_{\nu,i}$, the calculation of the enclosed volume can be simplified as
\begin{equation}
    \mathcal{H}(PF,\mathbf{r}) \propto \left(r_k-z_k^{ide}\right) \prod_{j\in I_{\nu,i}\setminus\{k\}} \left(r_j - z_j^{nad}\right),
\end{equation}
where the proportionality constant is $\prod_{j\in \overline{I}_{\nu,i}} \left(r_j - z_j^{nad}\right)$, and $k$ denotes that $\mathbf{r}$ resides within the region dominated by $\mathbf{z}^{ext(k)}$ while remaining not dominated by all the other ones. Similar to the first extreme case, $\mathcal{H}(PF,\mathbf{r})$ is $\Theta(\Delta^\nu)$ since $r_{k} - z_k^{ide} = \Theta(\Delta)$ and $r_{j} - z_j^{nad} = \Theta(\Delta)$ for every $j \in I_{\nu,i}\setminus\{k\}$.

\begin{figure*}[ht]
\centering
\subfloat[]{\label{fig:cs_p05_legend}\includegraphics[width=0.32\linewidth]{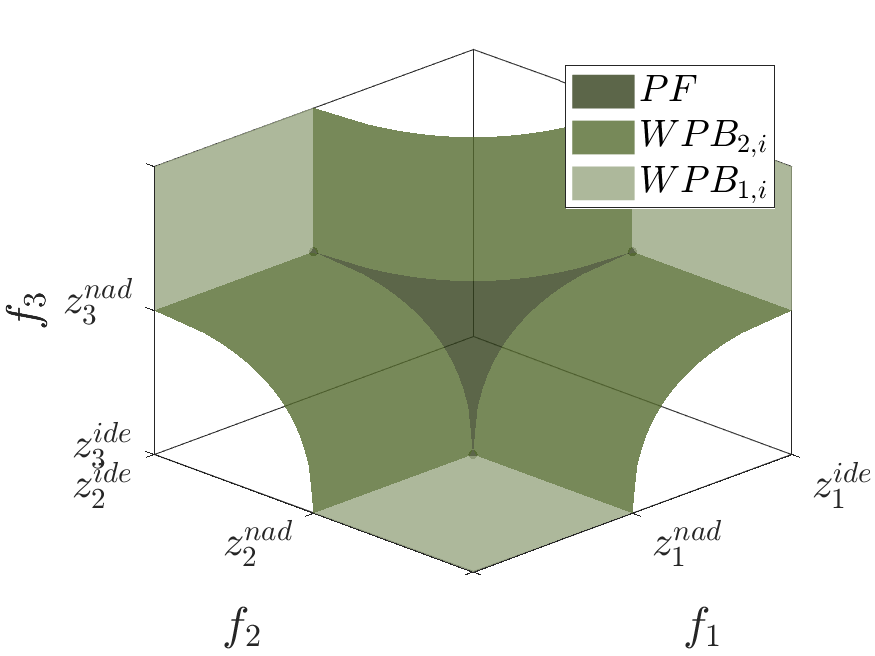}}
\hfil
\subfloat[]{\includegraphics[width=0.32\linewidth]{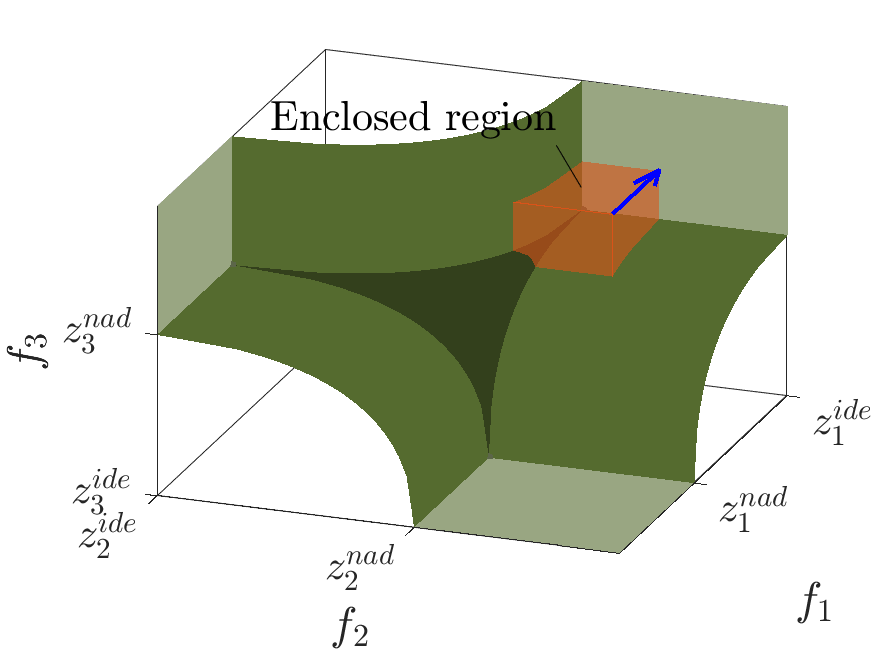}}
\hfil
\subfloat[]{\label{fig:cs_p05_nu2}\includegraphics[width=0.32\linewidth]{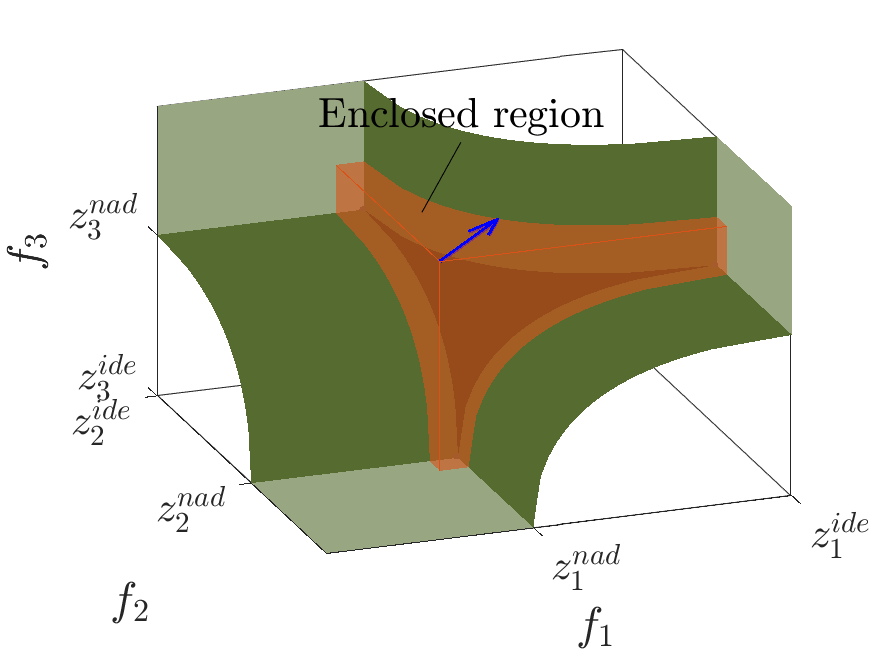}}
\caption{(a) Illustration of the $PF$ and all $WPB$s in the case with 3 objectives and $p=0.5$. (b) Visualization of moving $\mathbf{r}$ toward the $WPB_{1,i}$. (c) Visualization of moving $\mathbf{r}$ toward the $WPB_{2,i}$. For better clarity, $\Delta$ is set to a relatively large value of 0.5 in (b) and (c).}
\label{fig:cs_p05}
\end{figure*}

\begin{figure*}[ht]
\centering
\subfloat[]{\label{fig:vol_p05_m3_13}\includegraphics[width=0.32\linewidth]{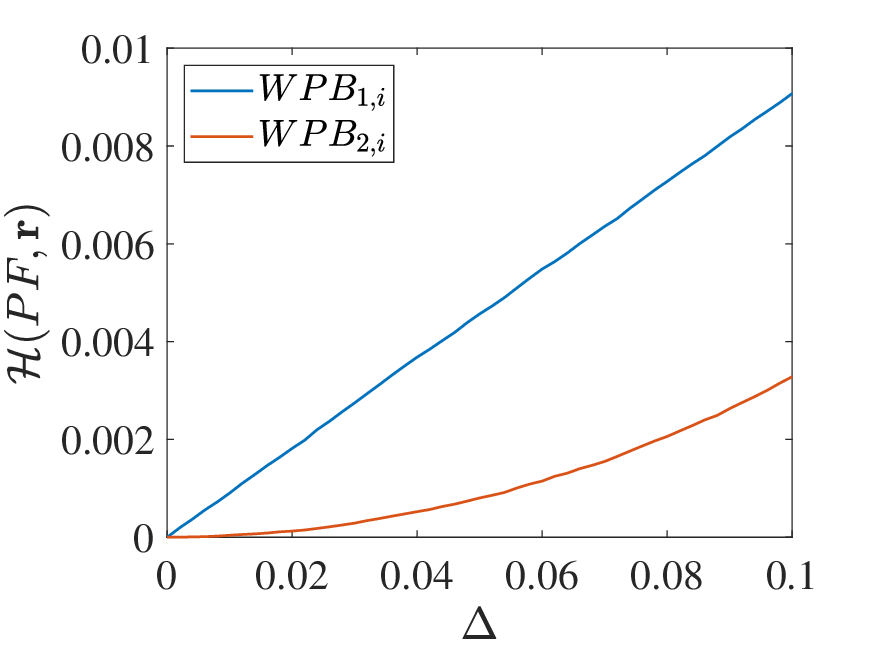}}
\hfil
\subfloat[]{\label{fig:vol_p05_m3_103}\includegraphics[width=0.32\linewidth]{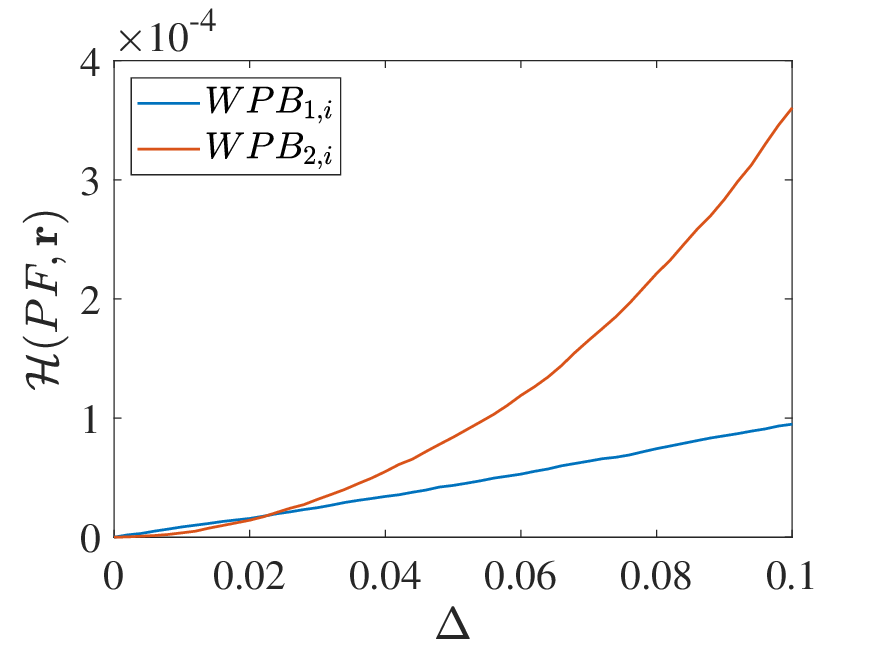}}
\hfil
\subfloat[]{\label{fig:vol_p05_m3_13_WPB2_m2b}\includegraphics[width=0.32\linewidth]{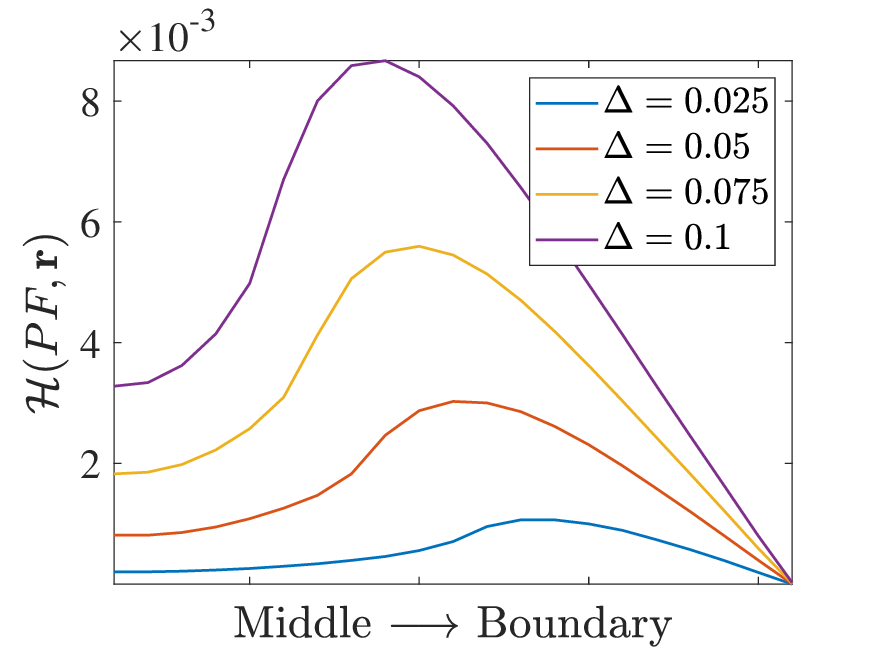}}
\caption{Curves of $\mathcal{H}(PF,\mathbf{r})$ with respect to $\Delta$ for the following cases: (a) $r_j=1.3$ for every $j \in \overline{I}_{2,i}$; (b) $r_j=1.03$ for every $j \in \overline{I}_{2,i}$. (c) Curves of $\mathcal{H}(PF,\mathbf{r})$ with respect to the relative position of $\mathbf{r}$ to the $WPB_{2,i}$ for the case where $r_j=1.3$ for every $j \in \overline{I}_{2,i}$. In these cases, $m=3$ and $p=0.5$.}
\label{fig:vol_p05}
\end{figure*}

\begin{table}[t]
  \centering
  \caption{Estimated polynomial coefficients from Lasso regression for the curves in \figurename~\ref{fig:vol_p2_m3_13},~\ref{fig:vol_p2_m3_103},~\ref{fig:vol_p05_m3_13}, and~\ref{fig:vol_p05_m3_103}.}
    \begin{threeparttable}
    \begin{tabular}{cccccc}
    \toprule
    $p$     & $\{r_i\}_{i=\nu+1}^m$    & $\nu$    & 1     & 2     & 3 \\
    \midrule
    \multirow{4}[4]{*}{2} & \multirow{2}[2]{*}{1.3} & 1     & 1.72e-01 & 1.68e+00 & 0 \\
          &       & 2     & 0      & 4.14e-01 & 9.95e-01 \\
\cmidrule{2-6}          & \multirow{2}[2]{*}{1.03} & 1     & 1.00e-02 & 8.38e-01 & 0 \\
          &       & 2     & 0      & 1.42e-01 & 1.15e+00 \\
    \midrule
    \multirow{4}[4]{*}{0.5} & \multirow{2}[2]{*}{1.3} & 1     & 8.82e-02 & 0      & 0 \\
          &       & 2     & 0      & 2.93e-01 & 2.50e-01 \\
\cmidrule{2-6}          & \multirow{2}[2]{*}{1.03} & 1     & 8.26e-04 & 8.53e-04 & 0 \\
          &       & 2     & 0      & 2.76e-02 & 7.09e-02 \\
    \bottomrule
    \end{tabular}%
    \begin{tablenotes}
        % \item[] Blank cells indicate that coefficients are exactly zero.
        \item[] The coefficient of the 4-th degree term is zero.
    \end{tablenotes}
    \end{threeparttable}
  \label{tab:vol_poly_pneq1}%
\end{table}%

\begin{figure}[ht]
\centering
\subfloat[$p=2$]{\label{fig:cs_p2_nu2_m2b}\includegraphics[width=0.5\linewidth]{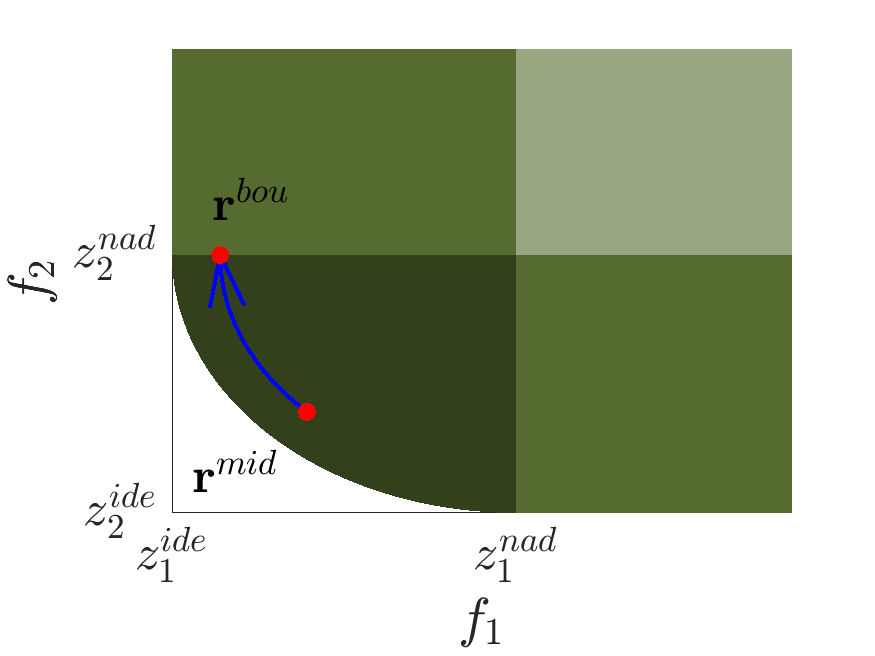}}
\hfil
\subfloat[$p=0.5$]{\label{fig:cs_p05_nu2_m2b}\includegraphics[width=0.5\linewidth]{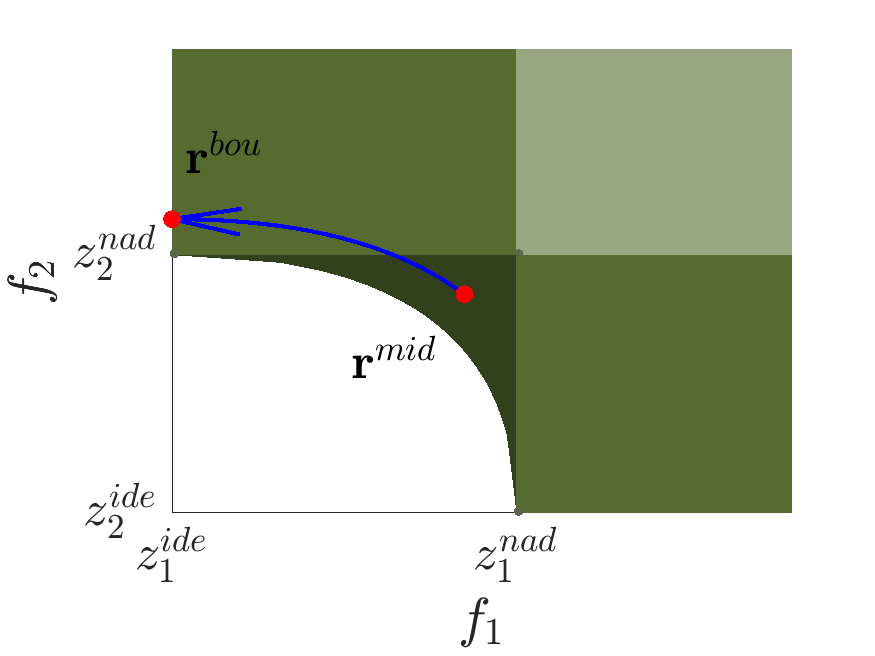}}
\caption{Illustration of moving the DRS from the middle (denoted as $\mathbf{r}^{mid}$) to the boundary (denoted as $\mathbf{r}^{bou}$) of the $WPB_{2,i}$.}
\label{fig:cs_p2n05_m2b}
\end{figure}

We now consider that $p$ is constant and $p \neq 1$. Accurately calculating the enclosed volume with respect to any $\mathbf{r}$ is nontrivial in this case~\cite{singh2025extended}. Hence, the enclosed volume is estimated via Monte Carlo sampling. Specifically, a set of points, uniformly distributed within a hyperrectangle defined by $\mathbf{z}^{ide}$ and $\mathbf{r}$, is sampled, and subsequently, infeasible ones (\ie, lies outside the enclosed region) are rejected. The volume is determined by multiplying the hyperrectangle's volume by the ratio of remaining points to sampled points.
In the following, $p=2$ and $p=0.5$ are investigated when $m=3$. $r_i$ is set to be $0.29+\frac{\Delta}{\sqrt{\nu}}$ for $i=1,\ldots,\nu$ and  $0.75+\frac{\Delta}{\sqrt{\nu}}$ for $i=1,\ldots,\nu$ for $p=2$ and $p=0.5$, respectively. Other settings remain consistent with the specified parameters in Section~\ref{sssec:peq1}.
% $\mathbf{z}^{ide}=\mathbf{0}$ and $\mathbf{z}^{nad}=\mathbf{1}$; $r_i = 1.3$ for $i=\nu+1,\ldots,3$ or $r_i = 1.03$ for $i=\nu+1,\ldots,3$.
%
When $r_i = 1.3$ for $i=\nu+1,\ldots,3$, the volume change results are presented in \figurename~\ref{fig:vol_p2_m3_13} and~\ref{fig:vol_p05_m3_13}. The enclosed volume with respect to the $WPB_{2,i}$ is always smaller in both cases, which closely aligns with the result of Case $p=1$.

When $r_i = 1.03$ for $i=\nu+1,\ldots,3$, the volume change results are presented in \figurename~\ref{fig:vol_p2_m3_103} and~\ref{fig:vol_p05_m3_103}. Both cases consistently show a reduction in the relative volume difference between the two curves. The result of Case $p=0.5$ is more similar to that of Case $p=1$. That is, the enclosed volume with respect to the $WPB_{2,i}$ is larger if $\Delta$ exceeds a threshold. The threshold in \figurename~\ref{fig:vol_p05_m3_103} is approximately 0.27. However, the enclosed volume with respect to the $WPB_{2,i}$ is consistently larger in Case $p=2$. Polynomial fitting, as previously described, is applied to the curves in \figurename~\ref{fig:vol_p2_m3_13},~\ref{fig:vol_p05_m3_13},~\ref{fig:vol_p2_m3_103}, and~\ref{fig:vol_p05_m3_103}. The results are presented in Table~\ref{tab:vol_poly_pneq1}, which also demonstrates that increasing the value of $\nu$ elevates the order of the lowest-order term in the volume-distance relationship.

Lastly, we examine the transition of $\mathbf{r}$ from $\mathbf{r}^{mid}$ to $\mathbf{r}^{bou}$. The trajectories are illustrated in \figurename~\ref{fig:cs_p2n05_m2b}. For Case $p=2$, as depicted in \figurename~\ref{fig:vol_p2_m3_13_WPB2_m2b}, the volume monotonically expands as $\mathbf{r}$ approaches the boundary of the $WPB_{2,i}$. This observation suggests that the enclosed volume of $\mathbf{r}^{mid}$ is the smallest. For Case $p=0.5$, as depicted in \figurename~\ref{fig:vol_p05_m3_13_WPB2_m2b}, the volume initially exhibits an increase analogous to the trend in \figurename~\ref{fig:vol_p2_m3_13_WPB2_m2b}. But the volume subsequently decreases to zero, mirroring a final trend similar to that in \figurename~\ref{fig:vol_p1_m3_13_WPB2_m2b}. This is because $\mathbf{r}^{bou}$ is on the $WPB_{1,i}$ in Case $p=0.5$. Nevertheless, the enclosed volume of $\mathbf{r}^{mid}$ is still typically lower than those of many locations. 

In conclusion, the DDR of a DRS remains higher provided that the DRS is sufficiently proximal to the $WPB_{\nu,i}$ with a larger $\nu$, which aligns with the finding in Case $p=1$. As the disadvantaged objectives of DRSs improve (\ie, DRSs are closer to the $PF$), their differences in the DDR also diminish. In contrast to Case $p=1$, $\mathbf{r}^{mid}$ demonstrates a higher DDR than most positions in Case $p=2$ and Case $p=0.5$.

\subsection{DRS Not on the \texorpdfstring{$WPB$}{WPB}: General Analysis}
This section analyzes the effect of $\nu$ on the enclosed volume in general. We conclude that the enclosed volume is $\Theta(\Delta^\nu)$ when any DRS sufficiently approaches the $WPB_{\nu,i}$, which is consistent with the findings in the case studies.

\subsubsection{General Formulation for Hypervolume Calculation}
The analysis utilizes the expression from~\cite{zhang2023hypervolume} for calculating the hypervolume between a $PF$ and a reference point $\mathbf{r}$, which is
\begin{equation}\label{eqn:general_hv}
\begin{aligned}
    \mathcal{H}(PF,\mathbf{r}) & = \mathcal{H}(WPF,\mathbf{r}) \\
    & = c_m \int_{\mathbb{S}_+^m} ~ \rho_\Omega(\mathbf{r},\boldsymbol{\lambda})^m ~ \mathrm{d}\lambda_1\cdots\mathrm{d}\lambda_{m},
\end{aligned}
\end{equation}
where
$\rho_\Omega(\mathbf{r}, \boldsymbol{\lambda}) = \max_{\mathbf{x} \in \Omega} \min_{i \in [m]} \left\{ \frac{r_i - f_i(\mathbf{x})}{\lambda_i} \right\}$, $c_m$ is a constant coefficient, and $\boldsymbol{\lambda}\in{\mathbb{S}_+^m}=\left\{\boldsymbol{\lambda}\in[0, 1]^m\left|\sum_{i=1}^m\lambda_i^2=1\right.\right\}$ is a unit vector on the unit sphere within the non-negative orthant of $\mathbb{R}^m$. We let $\rho(\mathbf{x}, \mathbf{r}, \boldsymbol{\lambda}) = \min_{i \in [m]} \left\{ \frac{r_i - f_i(\mathbf{x})}{\lambda_i} \right\}$, which defines a weighted Tchebycheff scalarization function~\cite{miettinen1998nonlinear}. We also let $\rho_\Omega(\mathbf{r}, \boldsymbol{\lambda}) = \min_{i\in[m]}\left\{ \frac{r_i - f_i^*(\mathbf{r},\boldsymbol{\lambda})}{\lambda_i} \right\}$, where $\mathbf{f}^*(\mathbf{r},\boldsymbol{\lambda}) = \left(f_1^*(\mathbf{r},\boldsymbol{\lambda}),\ldots,f_m^*(\mathbf{r},\boldsymbol{\lambda})\right)^\intercal$ represents the optimal objective vector associated with the single-objective optimization subproblem $\max_{\mathbf{x}\in\Omega} \rho(\mathbf{x}, \mathbf{r}, \boldsymbol{\lambda})$.

\subsubsection{Setup}
In~\cite{zhang2023hypervolume}, $r_i \geq z_i^{nad}, i=1,\ldots,m$. According to Proposition~\ref{pro:opt_inside_r}, $\mathbf{r}$ can be generalized to any objective vector dominated by at least one Pareto-optimal objective vector. Thus, $\mathbf{r}$ also can be a DRS in the objective space.

\begin{proposition}\label{pro:opt_inside_r}
if $\mathbf{r}$ is dominated by some feasible objective vector, then $r_i > f_i^*(\mathbf{r},\boldsymbol{\lambda})$ for $i=1,\ldots,m$.
\end{proposition}

\begin{IEEEproof}
If $r_i > f_i(\mathbf{x})$ holds for all $i=1,\ldots,m$, the function $\rho(\mathbf{x},\mathbf{r},\boldsymbol{\lambda})$ returns a positive value. Otherwise, $\rho(\mathbf{x},\mathbf{r},\boldsymbol{\lambda})$ returns a negative value. $\max_{\mathbf{x}\in\Omega}\rho(\mathbf{x},\mathbf{r},\boldsymbol{\lambda})$ must take the positive value.
\end{IEEEproof}

The contour surface of $\rho(\mathbf{x}, \mathbf{r}, \boldsymbol{\lambda})$ consists of the following $m$ hyperplanes
\begin{equation}
    \frac{r_i-z_i}{\lambda_i} = \rho(\mathbf{x},\mathbf{r},\boldsymbol{\lambda}) \text{~~for~~} i=1,\ldots,m,
\end{equation}
where $\mathbf{z}$ is an objective vector.
% at the optimum
Then, we can let
\begin{equation}\label{eqn:cusp}
    \rho_\Omega(\mathbf{r}, \boldsymbol{\lambda}) = \frac{r_1 - f_1^*(\mathbf{r},\boldsymbol{\lambda})}{\lambda_1} = \cdots = \frac{r_m - f_m^*(\mathbf{r},\boldsymbol{\lambda})}{\lambda_m},
\end{equation}
and $\mathbf{f}^*(\mathbf{r},\boldsymbol{\lambda}) = \mathbf{r} - \rho_\Omega(\mathbf{r},\boldsymbol{\lambda}) \boldsymbol{\lambda}$.

\begin{figure}[t]
\centering
\includegraphics[width=0.32\textwidth]{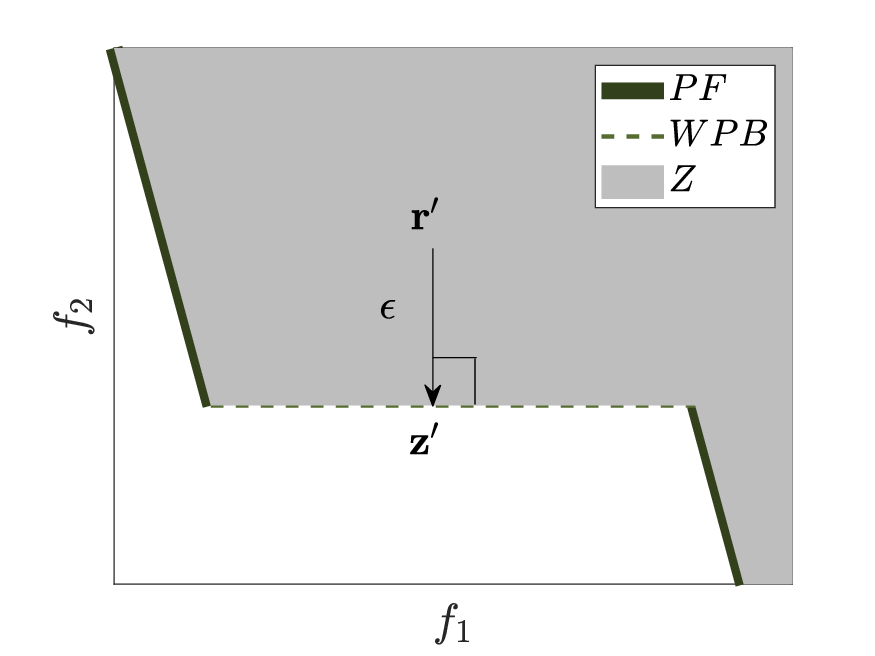}
\caption{Illustration of $\mathbf{r}^\prime$ and $\mathbf{z}^\prime$.}
\label{fig:asymptotically_approach}
\end{figure}

Moreover, we consider that a DRS denoted by $\mathbf{r}^\prime$ asymptotically approaches the $WPB_{\nu,i}$. That is, there exists an objective vector $\mathbf{z}^\prime$ on the $WPB_{\nu,i}$ such that $\epsilon^2 < \left(r_j^\prime - z_j^\prime\right) < \epsilon \text{~~for each~~} j\in I_{\nu,i}$ and $r_j^\prime = z_j^\prime \text{~~for each~~} j \in \overline{I}_{\nu,i}$, where $\epsilon$ is a first-order infinitesimal. \figurename~\ref{fig:asymptotically_approach} illustrates an example of $\mathbf{r}^\prime$ and $\mathbf{z}^\prime$ in the 2-objective case. For any $\boldsymbol{\lambda}$, we have $f^*_j(\mathbf{r}^\prime, \boldsymbol{\lambda}) < r_j^\prime$ for each $j\in I_{\nu,i}$ according to Proposition~\ref{pro:opt_inside_r}. We assume that the $PF$ achieves proper Pareto optimality~\cite{geoffrion1968proper,miettinen1998nonlinear}. Thus, we have $r_j^\prime - f_j^*(\mathbf{r}^\prime,\boldsymbol{\lambda}) \rightarrow 0^+$ as $\epsilon \rightarrow 0$ for any $\boldsymbol{\lambda}$ and $j \in I_{\nu,i}$.

\subsubsection{Analysis}
$\rho(\mathbf{x}, \mathbf{r}^\prime, \boldsymbol{\lambda})$ is maximized. Consequently, $\frac{r_j^\prime - f_j^*(\mathbf{r}^\prime,\boldsymbol{\lambda})}{\lambda_j}$ for $j=1,\ldots,m$ are constants if and only if $\lambda_j \rightarrow 0$ for all $j \in I_{\nu,i}$. Otherwise, they are first-order infinitesimals. To be clear, let 
\begin{equation}
\!\!\!\sigma(\mathbf{r}^\prime,\boldsymbol{\lambda})\!=\!
\begin{cases}
\dfrac{r_j^\prime \!-\! f_j^*(\mathbf{r}^\prime,\boldsymbol{\lambda})}{\lambda_j}, & \text{if}~ \lambda_j \!\rightarrow\! 0 ~\text{for all}~ j \!\in\! I_{\nu,i}, \\
\dfrac{1}{\epsilon}\dfrac{r_j^\prime \!-\! f_j^*(\mathbf{r}^\prime,\boldsymbol{\lambda})}{\lambda_j}, & \text{otherwise}.
\end{cases}
\end{equation}
That is, $\sigma(\mathbf{r}^\prime,\boldsymbol{\lambda}) = \Theta(1)$ with respect to $\epsilon$.
Furthermore, let $\Lambda_{\nu,i}$ be the set of $\boldsymbol{\lambda}$ such that $\lambda_j \rightarrow 0$ for every $j \in I_{\nu,i}$, and $\overline{\Lambda}_{\nu,i}$ be the complement of $\Lambda_{\nu,i}$. Eq.~\eqref{eqn:general_hv} is rewritten as
\begin{equation}\label{eqn:general_hv_re}
\begin{aligned}
    \mathcal{H}(PF,\mathbf{r}^\prime) =
    & c_m \int_{\Lambda_{\nu,i}} \sigma(\mathbf{r}^\prime,\boldsymbol{\lambda})^{m} \mathrm{d}\lambda_1\cdots\mathrm{d}\lambda_{m} + \\
    & c_m \int_{\overline{\Lambda}_{\nu,i}} \left(\epsilon\cdot\sigma(\mathbf{r}^\prime,\boldsymbol{\lambda})\right)^{m} \mathrm{d}\lambda_1\cdots\mathrm{d}\lambda_{m}. \\
\end{aligned}
\end{equation}
Without loss of generality, we consider $I_{\nu,i} = [\nu]$. According to the definition of $\sigma(\mathbf{r}^\prime,\boldsymbol{\lambda})$, the first term can be estimated as
\begin{equation}
\begin{aligned}
& c_m\int_{\Lambda_{\nu,i}} \sigma(\mathbf{r}^\prime,\boldsymbol{\lambda})^{m} \mathrm{d}\lambda_1\cdots\mathrm{d}\lambda_{m} \\
\le &~ c_m \alpha_1^m\int_{\Lambda_{\nu,i}}\mathrm{d}\lambda_1\cdots\mathrm{d}\lambda_{m} \\
% \int_{[0,O(\epsilon)]^\nu}
= &~ c_m \alpha_1^m \int_0^{O(\epsilon)}\mathrm{d}\lambda_1 \cdots \int_0^{O(\epsilon)}\mathrm{d}\lambda_{\nu} \\
& \int_{0}^{\sqrt{1-\sum_{i=1}^{\nu}\lambda_i^2}}\mathrm{d}\lambda_{\nu+1}\cdots\int_{0}^{\sqrt{1-\sum_{i=1}^{m-1}\lambda_i^2}}\mathrm{d}\lambda_m \\
= &~ \alpha_2\epsilon^\nu,
\end{aligned}
\end{equation}
where $\alpha_1$ and $\alpha_2$ are constants. Analogously, we also have 
\begin{equation}
\begin{aligned}
& c_m\int_{\Lambda_{\nu,i}} \sigma(\mathbf{r}^\prime,\boldsymbol{\lambda})^{m} \mathrm{d}\lambda_1\cdots\mathrm{d}\lambda_{m} \geq \alpha_3\epsilon^\nu,
\end{aligned}
\end{equation}
where $\alpha_3<\alpha_2$. The second term can be estimated as
\begin{equation}
\begin{aligned}
& c_m\int_{\overline{\Lambda}_{\nu,i}} \left(\epsilon\cdot\sigma(\mathbf{r}^\prime,\boldsymbol{\lambda})\right)^{m}
\mathrm{d}\lambda_1\cdots\mathrm{d}\lambda_{m} \\
\le &~ c_m \alpha_4^m \epsilon^m \int_{\overline{\Lambda}_{\nu,i}}\mathrm{d}\lambda_1\cdots\mathrm{d}\lambda_{m} \\
= &~ \alpha_5 \epsilon^m,
\end{aligned}
\end{equation}
where $\alpha_4$ and $\alpha_5$ are constants. Analogously, we also have 
\begin{equation}
\begin{aligned}
& c_m\int_{\overline{\Lambda}_{\nu,i}} \left(\epsilon\cdot\sigma(\mathbf{r}^\prime,\boldsymbol{\lambda})\right)^{m}
\mathrm{d}\lambda_1\cdots\mathrm{d}\lambda_{m} \geq \alpha_6\epsilon^m,
\end{aligned}
\end{equation}
where $\alpha_6<\alpha_5$. Finally, we obtain $\mathcal{H}(PF,\mathbf{r}^\prime) = \Theta(\epsilon^\nu) + \Theta(\epsilon^m)= \Theta(\epsilon^\nu)$. In the case that a DRS sufficiently approaches the $WPB_{\nu,i}$ of any shape, its DDR consistently remains smaller as the value of $\nu$ increases. In other words, a larger $\nu$ typically results in a lower DDR.

\section{Test Problem with Weak Pareto Boundary}\label{sec:problem}

\subsection{Existing Test Problems}
% We seek to 
Various test problems are required to comprehensively evaluate the impact of the $WPB$ on MOEAs while minimizing the influence of confounding factors. Specifically, the test problems should encompass diverse shapes of $PF$s (\eg, linear, convex, concave, and discontinuous), and accordingly, the associated $WPB$s exhibit distinct shapes. The sizes of $WPB$s should also vary in these test problems. Moreover, the test problems should not have other complicated problem characteristics as summarized in~\cite{zhou2019set,brockhoff2022using}, enabling the application of a wide range of reproduction operators (\eg, SBX and PM~\cite{purshouse2007evolutionary}) to efficiently produce high-quality solutions. For example, the test problem has mild bias~\cite{huband2006review}, and thus uniformly sampled solutions of the test problem can exhibit a nicely even distribution in the objective space.

However, existing test problems with $WPB$s are insufficiently equipped to fulfill the intended purposes. Certain test problems exhibit distance functions that are multimodal~\cite{wang2019scalable,ishibuchi2020effects,wang2023dilemma}, or their $PS$s are highly complicated~\cite{lin2024multi}. These characteristics can hinder the population from effectively covering the $PF$, thereby misleading the analysis of the $WPB$'s impact. Besides, they do not have various $WPB$s. The test problem generator proposed in \cite{wang2024multi} is currently the most capable of adjusting the $WPB$ and alleviating optimization challenges associated with other problem characteristics. However, the variation of the generated problems is still limited. For example, the test problem generator only allows modification of the overall size of the $WPB$ and does not support adjustment of the relative sizes among $WPB$s; the $PF$ can only change convexity based on an inverted triangular shape. Besides, its position functions are written as
\begin{equation}
\begin{cases}
    h_1(x_1,\ldots,x_{m-1}) = 1 \!-\! \left(\prod_{j=1}^{m-1}\cos\left(0.5 \pi x_j\right)\right)^\beta, \\
    h_i(x_1,\ldots,x_{m-i}) = \\
    1 - \left(\sin\left(0.5 \pi x_{m-i+1}\right) \prod_{j=1}^{m-i}\cos\left(0.5 \pi x_j\right)\right)^\beta, \\
    \qquad\qquad\qquad\qquad\qquad\qquad i=2,\ldots,m-1, \\
    h_m(x_1) = 1 - \sin^\beta\left(0.5 \pi x_1\right),
\end{cases}
\end{equation}
which signify different levels of optimization difficulty and introduce bias.

Based on these considerations, two test problem generators are developed. Both generators are capable of creating test problems involving $WPB$s of diverse categories, shapes, and sizes. The primary distinction between the two generators lies in the spatial relation between the $WPB$ and the $PF$. The $PF$ is continuous for the first generator, and the whole $PF$ lies on one side of any given $WPB_{\nu,i}$. In contrast, the second generator produces a discontinuous $PF$, where any $WPB_{\nu,i}$ with $\nu \geq 2$ is situated at some discontinuous part of the $PF$.

\subsection{Proposed Test Problem Generators}
Each objective function of the proposed test problem generators takes the following form
\begin{equation}\label{eqn:f_form}
f_i(\mathbf{x}) = s_i h_i(\mathbf{x}_{\Rmnum1})(1+g_i(\mathbf{x}_{\Rmnum2})) + z_i^{ide},
\end{equation}
where $s_i$ is a parameter such that different objective functions can have different ranges of values; $\mathbf{x}\in\Omega=[0,1]^n\subset\mathbb{R}^n$, and $\mathbf{x}_{\Rmnum1}=(x_1,\ldots,x_{m})^{\intercal}$ and $\mathbf{x}_{\Rmnum2}=(x_{m+1},\ldots,x_{2m})^{\intercal}$ are subvectors of $\mathbf{x}$. $g_i(\mathbf{x}_{\Rmnum2})\geq 0$ is called the distance function, which specifies as
\begin{equation}
    g_i(\mathbf{x}_{\Rmnum2}) = \ell_i \left(2x_{m+i}-1\right),
\end{equation}
where $\boldsymbol{\ell} = \left( \ell_1,\ldots,\ell_m \right)^\intercal$ is a parameter that controls the overall size of the $WPB$. $h_i(\mathbf{x}_{\Rmnum1})$ is termed as the position function, which determines the shape of the $PF$. Some remarks are highlighted as follows:
\begin{itemize}
    \item Eq.~\eqref{eqn:f_form} is called the multiplication-based form. According to~\cite{wang2019generator}, this form facilitates a more uniform distribution of the objective vector across the feasible objective region compared to the addition-based form.
    \item Eq.~\eqref{eqn:f_form} also enables setting the ideal objective vector, in case the algorithm implicitly defaults the ideal objective vector to the zero vector.
    \item The number of objectives is scalable, whereas the number of variables is a minimal value corresponding to the given number of objectives (\ie, $2m$). This is because an excessive number of variables reduces optimization efficiency, potentially causing misleading conclusions when analyzing the impact of the $WPB$.
\end{itemize}

\begin{figure*}[ht]
\centering
\subfloat[]{\label{fig:generator1_example_convex}\includegraphics[width=0.32\linewidth]{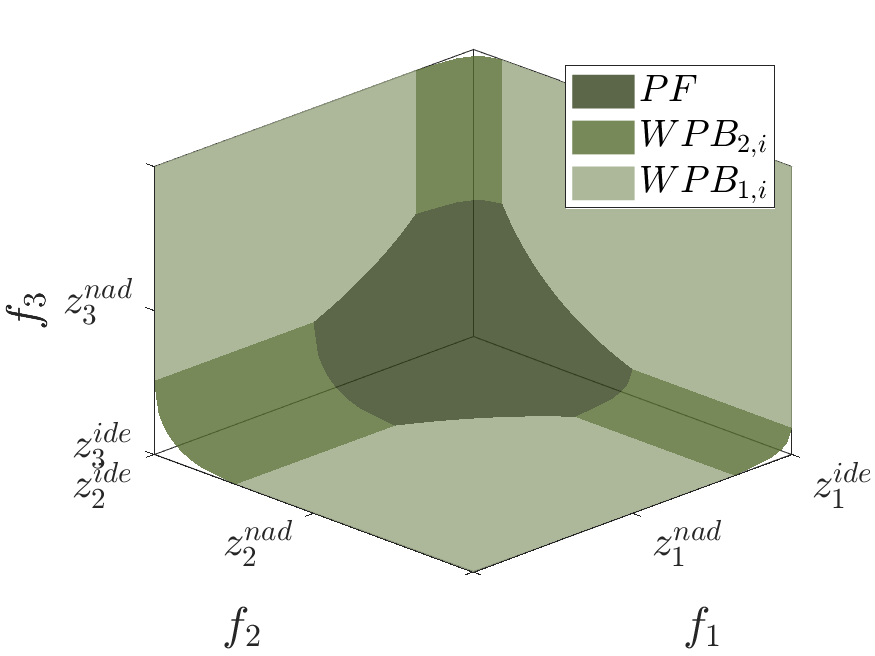}}
\hfil
\subfloat[]{\label{fig:generator1_example_mixed}\includegraphics[width=0.32\linewidth]{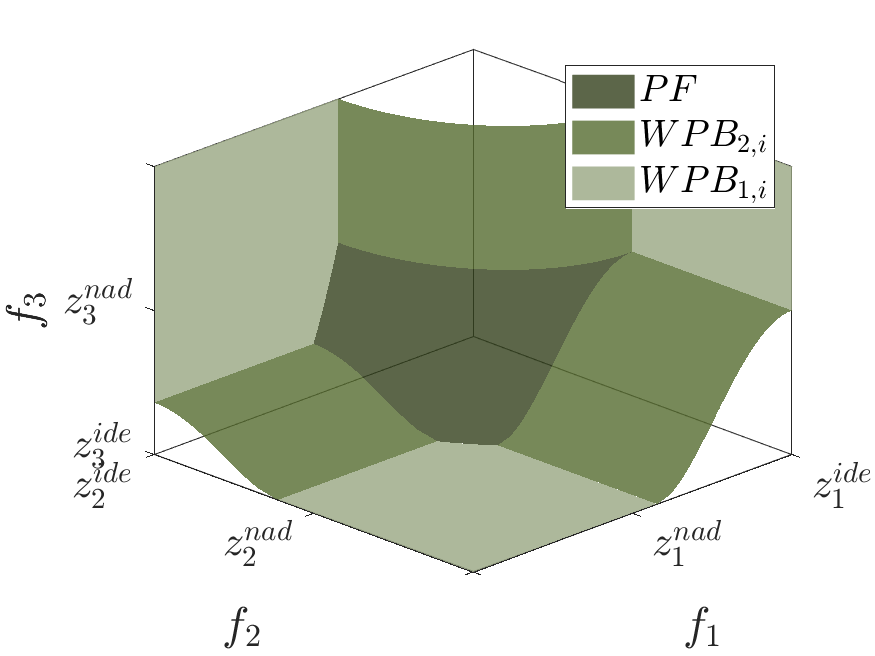}}
\hfil
\subfloat[]{\label{fig:generator2_example_concave}\includegraphics[width=0.32\linewidth]{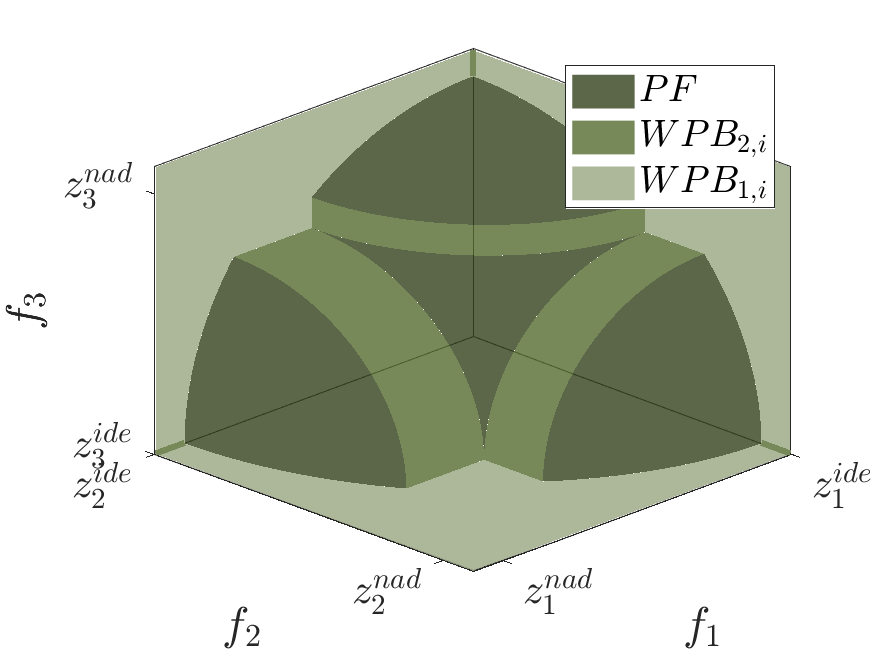}}
\hfil
\caption{Impact of generator parameter settings on the feasible objective region. The settings of $\mathbf{s}$ and $\mathbf{z}^{ide}$ are arbitrary, and each element of $\boldsymbol{\ell}$ is sufficiently large.
(a) $\mathbf{p} \!=\! (2,2,2)^\intercal,\mathbf{d} \!=\! (0.5,0.7,0.7)^\intercal$. (b) $\mathbf{p} \!=\! (0.5,0.5,2)^\intercal,\mathbf{d} \!=\! (0.7,0.5,0.5)^\intercal$. (c) $\mathbf{p} \!=\! (0.5,0.5,0.5)^\intercal,\mathbf{d} \!=\! (0.5,0.7,0.7)^\intercal,\mathbf{r} \!=\! (0.6,0.4,0.2)^\intercal$.}
% (convex) $\mathbf{p}=(2,2,2)^\intercal,\mathbf{d}=(0.5,0.7,0.5)^\intercal$.
\label{fig:generator_example}
\end{figure*}

\subsubsection{Generator 1}
The generator is used to obtain test problems with continuous $PF$s.
Firstly, $\hat{\mathbf{y}}=(\hat{y}_1,\ldots,\hat{y}_m)^\intercal$ is used to represent the following expressions
\begin{equation}\label{eqn:y_hat}
    \hat{y}_i = \frac{x_i}{\sum_{j=1}^m x_j} \mbox{ for } i = 1,\ldots,m.
\end{equation}
$\left\{\hat{\mathbf{y}}|\mathbf{x}_{\Rmnum1}\in[0,1]^m\right\}$ is an ($m-1$)-dimensional unit simplex. If $\sum_{j=1}^m x_j=0$, we define $\hat{y}_i=\frac{1}{m}$ for $i=1,\ldots,m$.
Then, let
\begin{equation}\label{eqn:yhat2y}
\begin{aligned}
    y_i(\mathbf{x}_{\Rmnum1}|\mathbf{d}) = & \min\left\{\hat{y}_i,d_i\right\} + \\
    & \frac{\max\left\{0,d_i-\hat{y}_i\right\}\sum_{j=1}^m \max\left\{0,\hat{y}_j-d_j\right\}}{\sum_{j=1}^m \max\left\{0,d_j-\hat{y}_j\right\}},
\end{aligned}
\end{equation}
where $\mathbf{d}$ is a parameter satisfying $d_i\geq\frac{1}{m-1}$ for $i=1,\ldots,m$. We have $\sum_{i=1}^m \hat{y}_i-d_i = 1-\sum_{i=1}^m d_i \leq 0$, and thus,
\begin{equation}
    \frac{\sum_{j=1}^m \max\left\{0,\hat{y}_j-d_j\right\}}{\sum_{j=1}^m \max\left\{0,d_j-\hat{y}_j\right\}} \leq 1.
\end{equation}
This indicates that $y_i(\mathbf{x}_{\Rmnum1}|\mathbf{d}) \leq d_i$ for $i=1,\ldots,m$ and $\sum_{i=1}^m y_i(\mathbf{x}_{\Rmnum1}|\mathbf{d})=1$ always hold, even if $\exists i$ such that $\hat{y}_i > d_i$.
Finally, the position function value vector is given by
\begin{equation}\label{eqn:f_pos_1}
    \mathbf{h}(\mathbf{x}_{\Rmnum1}|\mathbf{d},\mathbf{p}) = \left( \mathbf{y}(\mathbf{x}_{\Rmnum1}|\mathbf{d}) \oslash \mathbf{d} \right)^{\circ \mathbf{p}},
\end{equation}
where ``$\oslash$'' and ``$\circ$'' denote the Hadamard division and Hadamard power, respectively, and $\mathbf{p}$ is a positive vector.

In Eq.~\eqref{eqn:yhat2y}, $\mathbf{d}$ modifies boundaries of the $PF$, thereby altering the relative sizes among $WPB_{\nu,i}$ for all $\nu$ and $i$. 
In particular, as $d_i$ reduces, the relative size of the $WPB_{m-1,i}$ increases, whereas the relative size of the $WPB_{1,i}$ decreases.
% The remaining parameters are explained as follows.
The division by $\mathbf{d}$ in Eq.~\eqref{eqn:f_pos_1} is to align the nadir objective vector of any setting of $\mathbf{p}$ and $\mathbf{d}$. The nadir objective vector is always $\mathbf{s}+\mathbf{z}^{ide}$. $\mathbf{p}$ determines the $PF$ shape. If $p_i=1$ for $i=1,\ldots,m$, the $PF$ is linear. If $p_i>1$ for $i=1,\ldots,m$, the $PF$ is convex. If $p_i<1$ for $i=1,\ldots,m$, the $PF$ is concave. If $p_i>1$ for some $i$ and $p_j<1$ for some $j$, then the $PF$ contains both convex and concave segments.

The relative size is discussed in detail below. The relative size of the $WPB_{\nu,i}$ is determined by the size of its intersection with the $PF$. Note that the intersection is also a boundary of the $PF$. If an objective vector lies on the intersection, the corresponding $\mathbf{y}(\mathbf{x}_{\Rmnum1}|\mathbf{d})$ satisfies
\begin{equation}
    0 \leq y_j(\mathbf{x}_{\Rmnum1}|\mathbf{d}) \leq d_j \text{ for } j=1,\ldots,m,
\end{equation}
\begin{equation}\label{eqn:y_nu}
    \sum_{j \in I_{\nu,i}} y_j(\mathbf{x}_{\Rmnum1}|\mathbf{d}) = \max\left\{0, 1 - \sum_{j\in\overline{I}_{\nu,i}}d_j\right\},
\end{equation}
and
\begin{equation}\label{eqn:y_m_minus_nu}
    \sum_{j \in \overline{I}_{\nu,i}} y_j(\mathbf{x}_{\Rmnum1}|\mathbf{d}) = \min\left\{1, \sum_{j\in\overline{I}_{\nu,i}} d_j\right\}.
\end{equation}
Particularly, the relative size of the $WPB_{m-1,i}$  is positively correlated with the size of the ($m-2$)-dimensional simplex characterized by the following $m-1$ vertices
\begin{equation}
v_{j,k} = 
\begin{cases}
    1 - d_i, & k = j, \\
    d_i, & k = m - i + 1, \\
    0, & \text{otherwise},
\end{cases}
\end{equation}
for all $j \in [m] \setminus \{m-i+1\}$. Each element in $\mathbf{d}$ can independently control one of $WPB_{m-1,i}$ for $i=1,\ldots,m$. As $d_i$ reduces, any vertex becomes more distant from the others, and accordingly, the relative size of the $WPB_{m-1,i}$ grows.
Conversely, the relative size of the $WPB_{1,i}$ is negatively correlated with the size of the ($m-2$)-dimensional simplex. The $m-1$ vertices are
\begin{equation}
v_{j,k} = 
\begin{cases}
    d_i, & k = j, \\
    0, & \text{otherwise},
\end{cases}
\end{equation}
for all $j \in [m] \setminus \{i\}$. As $d_i$ reduces, any vertex becomes closer to the others, which diminishes the relative size.

\begin{table}[t]
  \centering
  \caption{Parameters in the test problem generators.}
    \renewcommand{\arraystretch}{1.5}\setlength{\tabcolsep}{1mm}{
    \begin{tabular}{ccp{15em}}
    \toprule
    Symbol & Range & \multicolumn{1}{c}{Description} \\
    \midrule
    $\mathbf{s}$     & $(0,\infty)^m$ & Ranges of objective functions \\
    $\mathbf{p}$     & $(0,\infty)^m$ & Shape of the $PF$ or the $WPB$ \\
    $\boldsymbol{\ell}$   & $[0,\infty)^m$ & Overall size of the $WPB$ \\
    $\mathbf{d}$     & $\left[\frac{1}{m-1},1\right)^m$ & Relative sizes among $WPB_{\nu,i}$ for all $\nu$ and $i$ \\
    $\mathbf{r}$ (Generator 2 only) & $[0,\infty)^m$ & Overall size of the $WPB_{\nu,i}$ with $\nu \geq 2$ \\
    \bottomrule
    \end{tabular}%
    }
  \label{tab:generator_paras}%
\end{table}%

\subsubsection{Generator 2}
The generator is used to obtain test problems with discontinuous $PF$s, each of which includes one middle part and at most $m$ boundary parts. This generator also has $\mathbf{d}$ and $\mathbf{p}$ as parameters. Besides, a new parameter $\mathbf{r}$ satisfying $r_i \geq 0$ for $i=1,\ldots,m$ is introduced. $r_i$ determines the distance along the $f_i$-axis between the $i$-th boundary part and the central part. Analogously, we define
\begin{equation}
    y_i(\mathbf{x}_{\Rmnum1}|\mathbf{d},\mathbf{r}) =
    \begin{cases}
        \hat{y}_i + r_i, & \text{if } \hat{y}_i>d_i, \\
        \hat{y}_i, & \text{otherwise},
    \end{cases}
\end{equation}
where $\hat{\mathbf{y}}$ is identically given by Eq.~\eqref{eqn:y_hat}. The position functions are formulated as
\begin{equation}
    \mathbf{h}(\mathbf{x}_{\Rmnum1}|\mathbf{d},\mathbf{r},\mathbf{p}) = \left(\mathbf{y}(\mathbf{x}_{\Rmnum1}|\mathbf{d},\mathbf{r}) \oslash (1+\mathbf{r})\right)^{\circ\mathbf{p}}.
\end{equation}
The nadir objective vector of any test problem obtained by this generator is also $\mathbf{s}+\mathbf{z}^{ide}$.

\subsubsection{Examples}
The description of control parameters is summarized in Table~\ref{tab:generator_paras}. If $m=3$ and $d_i=\frac{1}{2}$ for $i=1,2,3$, then Case $p_i = 1$ is identical to the case in \figurename~\ref{fig:cs_p1_legend}. Case $p_i = 2$ and Case $p_i = 0.5$ illustrate shapes of the $WPF$ similar to those in \figurename~\ref{fig:cs_p2_legend} and~\ref{fig:cs_p05_legend}, respectively.
We further specify $s_i = 1$ and $\ell_i = \ell_0$ for $i=1,2,3$. In \figurename~\ref{fig:cs_p1_legend}, the size of the $WPB_{2,i}$ is its relative size multiplied by $\ell_0$, \ie, $\sqrt{2}\ell_0$, and the size of the $WPB_{1,i}$ is $\ell_0^2$. If $\ell_0 > \sqrt{2}$, then the size of the $WPB_{1,i}$ exceeds that of the $WPB_{m-1,i}$. Generally, given that $s_i = 1$, $p_i = 1$, $\ell_i = \ell_0$, and $d_i=\frac{1}{m-1}$ for $i=1,\ldots,m$, the sizes of the $WPB_{m-1,i}$ and the $WPB_{1,i}$ are $\frac{\sqrt{m} \cdot \ell_0}{(m-1)!}$ and $\ell_0^{m-1}$, respectively. Evidently, the size of the $WPB_{1,i}$ exceeds that of the $WPB_{m-1,i}$ when $\ell_0$ and $m$ are sufficiently large.

Three other examples are shown in \figurename~\ref{fig:generator_example}. In the 3-objective case, the relative size can be viewed as the width of the $WPB_{2,i}$. \figurename~\ref{fig:generator1_example_convex} shows that $WPB_{2,i}$ for $i=1,2,3$ have convex shapes and different relative sizes. \figurename~\ref{fig:generator1_example_mixed} shows that two of $WPB_{2,i}$ for $i=1,2,3$ have mixed shapes but different relative sizes, and the remaining one has a concave shape. In \figurename~\ref{fig:generator2_example_concave}, $WPB_{2,i}$ for $i=1,2,3$ with the same width but varying lengths are located at the discontinuous part of the $PF$. Moreover, the $WPB_{1,i}$ is always linear in these examples, as stated in Property~\ref{prop:nu_1_plane}.

\section{Experimental Studies}\label{sec:experiment}
This section validates our theoretical analysis through experiments, including the impact of $WPB$s of different categories and shapes on the MOEAs' performance. Additional experiments are provided in the appendices:
\begin{itemize}
    \item The impact of $WPB$s of different sizes is analyzed in Appendix~\ref{sec:impact_wpb_size}. Some MOEAs are more sensitive to changes in the overall size of the $WPB$, whereas the others are more sensitive to variations in the relative sizes between the $WPB_{2,i}$ and the $WPB_{1,i}$.
    \item The impact of the spatial relation between the $WPB$ and the $PF$ is discussed in Appendix~\ref{sec:impact_disc_pf}. All examined MOEAs exhibit severe performance degradation on the MOP with a convex and discontinuous $PF$.
    \item Existing representative MOEAs are evaluated on various benchmark instances in Appendix~\ref{sec:benchmarking}. No algorithm consistently performs best across the majority of instances.
\end{itemize}

\subsection{Experimental Setup}
\subsubsection{Test Instances for Analysis}
Table~\ref{tab:ins_analysis_paras} shows 16 example instances, which are used to investigate the impact of the $WPB$. EMOP1--EMOP3 feature $WPB$s of different shapes and enable extension to have more than 3 objectives. A larger number of objectives indicates more categories of $WPB$s. They are employed in Section~\ref{ssec:impact_nu}. EMOP4--8 have different overall sizes of the $WPB$, while EMOP9--EMOP13 have different relative sizes of the $WPB_{\nu,i}$.
They are employed in Appendix~\ref{sec:impact_wpb_size}. EMOP14--EMOP16 with discontinuous $PF$s are employed in Appendix~\ref{sec:impact_disc_pf}.

\begin{table}[t]
  \centering
  \caption{Example instances. $z_i^{ide}=0$ and $s_i = 1$ for $i=1,\ldots,m$.}
  % \vspace{-2ex}
    \setlength{\tabcolsep}{5.5mm}{
    \begin{threeparttable}
    \begin{tabular}{ccccc}
    \toprule
    Instance & $\mathbf{p}$     & $\boldsymbol{\ell}$     & $\mathbf{d}$     & $\mathbf{r}$ \\
    \midrule
    EMOP1 & 2     & 4     & $\frac{1}{m-1}$ & NA \\
    EMOP2 & 1     & 4     & $\frac{1}{m-1}$ & NA \\
    EMOP3 & 0.5   & 4     & $\frac{1}{m-1}$ & NA \\
    \midrule
    EMOP4 & 1     & 4     & 0.7   & NA \\
    EMOP5 & 1     & 40    & 0.7   & NA \\
    EMOP6 & 1     & 400   & 0.7   & NA \\
    EMOP7 & 1     & 4000  & 0.7   & NA \\
    EMOP8 & 1     & 40000 & 0.7   & NA \\
    \midrule
    EMOP9 & 1     & 400   & 0.9   & NA \\
    EMOP10 & 1     & 400   & 0.8   & NA \\
    EMOP11 & 1     & 400   & 0.7   & NA \\
    EMOP12 & 1     & 400   & 0.6   & NA \\
    EMOP13 & 1     & 400   & 0.5   & NA \\
    \midrule
    EMOP14 & 2     & 4     & 0.5   & 1 \\
    EMOP15 & 1     & 4     & 0.5   & 1 \\
    EMOP16 & 0.5   & 4     & 0.5   & 1 \\
    \bottomrule
    \end{tabular}%
    \begin{tablenotes}
        \item[] ``NA'' in the cell with respect to $\mathbf{r}$ indicates that their corresponding instances are obtained by Generator 1 in Section~\ref{sec:problem}. The remaining instances are obtained by Generator 2.
        \item[] A scalar in the cell indicates that all elements of the corresponding parameter equal this scalar. For example, ``2'' for $\mathbf{p}$ means $\mathbf{p} = (2,2,2)^\intercal$.
    \end{tablenotes}
    \end{threeparttable}
    }
  \label{tab:ins_analysis_paras}%
\end{table}%

\subsubsection{Candidate Algorithms}
Existing MOEAs can be categorized into dominance-based MOEAs, decomposition-based MOEAs, and indicator-based MOEAs. From each category, one representative algorithm is selected:
\begin{itemize}
    \item MultiGPO~\cite{zhu2022new} (dominance-based). It employs the relaxed dominance criterion in~\cite{zhu2016generalization}, which belongs to the family of cone dominance~\cite{noghin1997relative}. An objective vector $\mathbf{u}$ is said to dominate another objective vector $\mathbf{v}$ under this criterion, if and only if 
    \begin{equation}
    \begin{cases}
    \forall i \!\in\! [m] : u_i \!+\! \sum_{k \neq i} \delta_{i} u_k \!\leq\! v_i \!+\! \sum_{k \neq i} \delta_{i} v_k, \\
    \exists j \!\in\! [m] : u_j \!+\! \sum_{k \neq j} \delta_{j} u_k \!<\! v_j \!+\! \sum_{k \neq j} \delta_{j} v_k,
    \end{cases}
    \end{equation}
    where $\delta_i$ for $i=1,\ldots,m$ are parameters. Cone dominance has been demonstrated to be effective in eliminating DRSs~\cite{ishibuchi2020effects,pang2020nsga,zheng2024boundary}.
    \item MOEA/D-Gen~\cite{giagkiozis2013generalized,wang2019scalable} (decomposition-based). It employs the framework of MOEA/D~\cite{zhang2007moea} and defines the scalarized subproblem as
    \begin{equation}
    \begin{aligned}
        & g^{gen}(\mathbf{x}|\mathbf{w},\mathbf{z}^*) = \\
        & \max_{1 \leq i \leq m} \left\{ w_i \left( f_i(\mathbf{x}) \!-\! z_i^* \!+\! \rho\sum_{j=1}^m \left(f_j(\mathbf{x}) \!-\! z_j^*\right) \right) \right\},
    \end{aligned}
    \end{equation}
    where $\mathbf{w}$ is a weight vector, $\mathbf{z}^*$ is the estimated ideal objective vector, and $\rho$ is a small positive parameter to avoid weak Pareto optimality. $\mathbf{z}^*$ usually takes the current lower bounds of objective function values as its entries.
    \item IMOEA-ARP~\cite{wang2024multi} (indicator-based). It leverages the hypervolume indicator~\cite{zitzler1999multiobjective} and prioritizes selecting solutions that contribute most significantly to the overall hypervolume. IMOEA-ARP is an effective method since the DRS often has a very small hypervolume contribution. The reference point in IMOEA-ARP is adaptive to enhance the diversity of its population.
\end{itemize}

The parameters of the candidate algorithms are set based on the corresponding references. The population size in our experiments is set to 91 for 3-objective instances, 165 for 4-objective instances, and 330 for 5-objective instances. The maximum number of function evaluations is 50,000 for 3-objective instances, 90,000 for 4-objective instances, and 180,000 for 5-objective instances. SBX and PM~\cite{purshouse2007evolutionary} are used for reproduction. Each algorithm is executed independently on each instance 30 times. All experiments are implemented on the PlatEMO platform~\cite{tian2017platemo}.

\begin{figure*}[ht]
\centering
\subfloat[MultiGPO on EMOP1]{\includegraphics[width=0.32\linewidth]{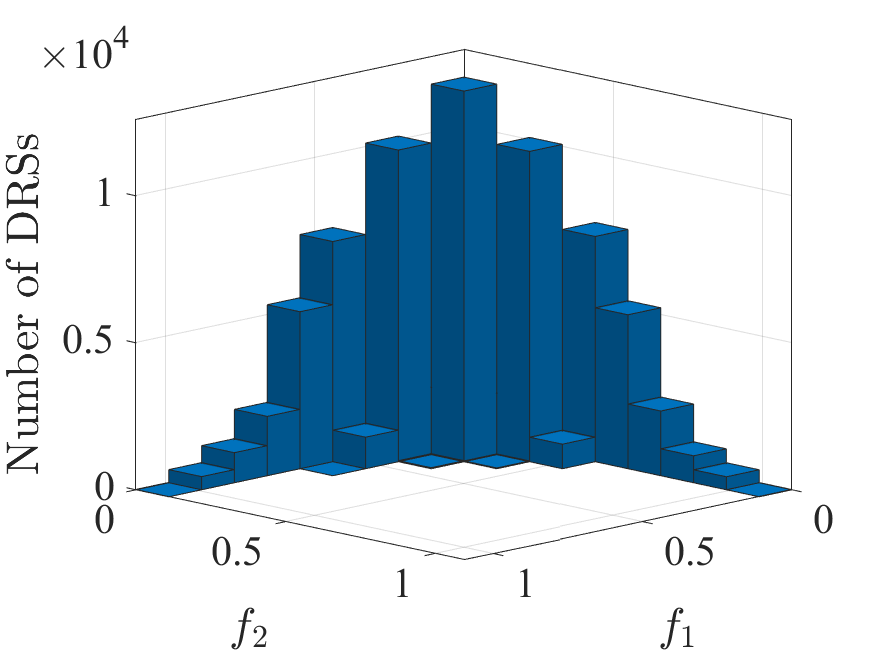}}
\hfil
\subfloat[MOEA/D-Gen on EMOP1]{\includegraphics[width=0.32\linewidth]{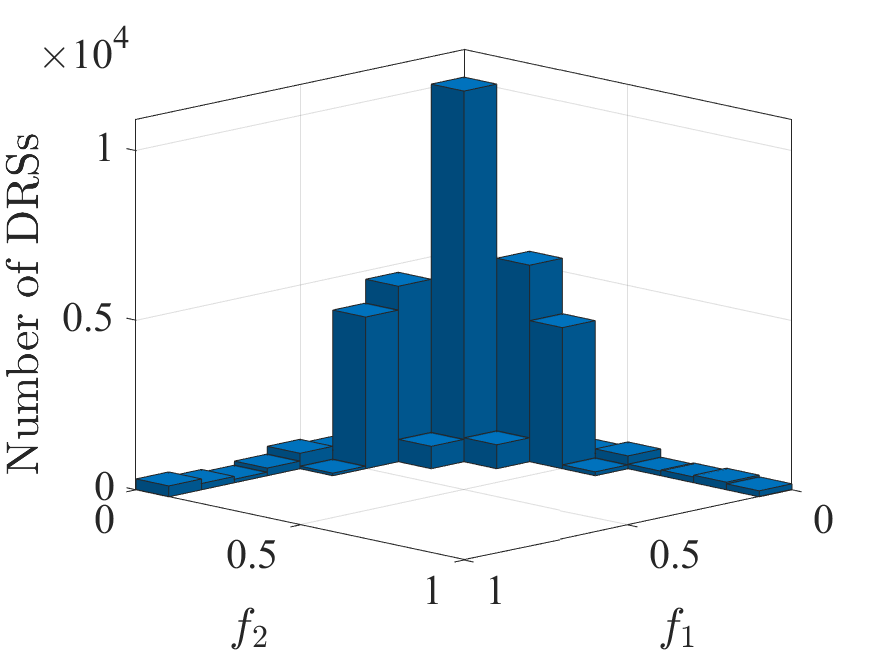}}
\hfil
\subfloat[IMOEA-ARP on EMOP1]{\includegraphics[width=0.32\linewidth]{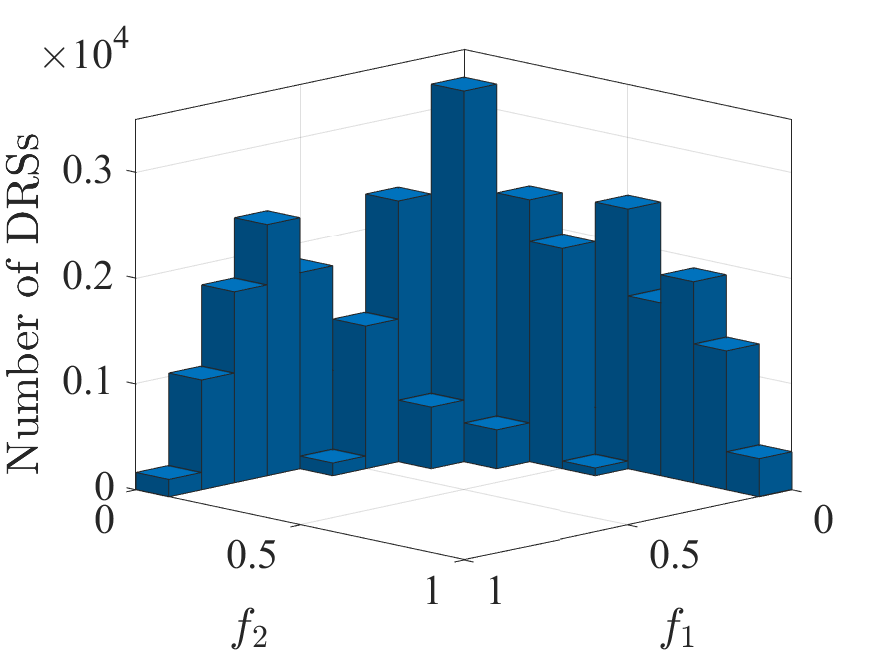}}
\hfil
\vspace{-1ex}
\subfloat[MultiGPO on EMOP2]{\includegraphics[width=0.32\linewidth]{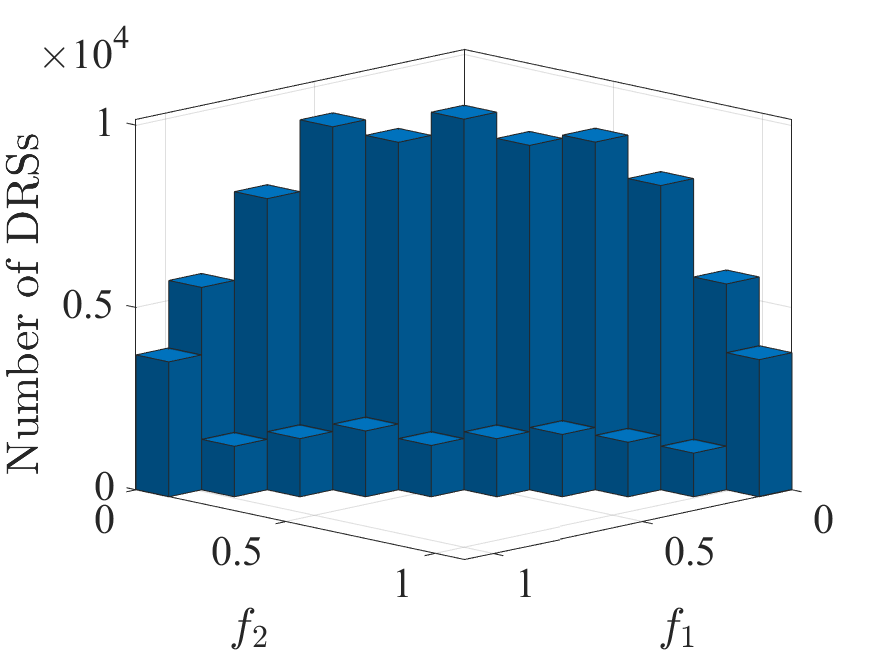}}
\hfil
\subfloat[MOEA/D-Gen on EMOP2]{\includegraphics[width=0.32\linewidth]{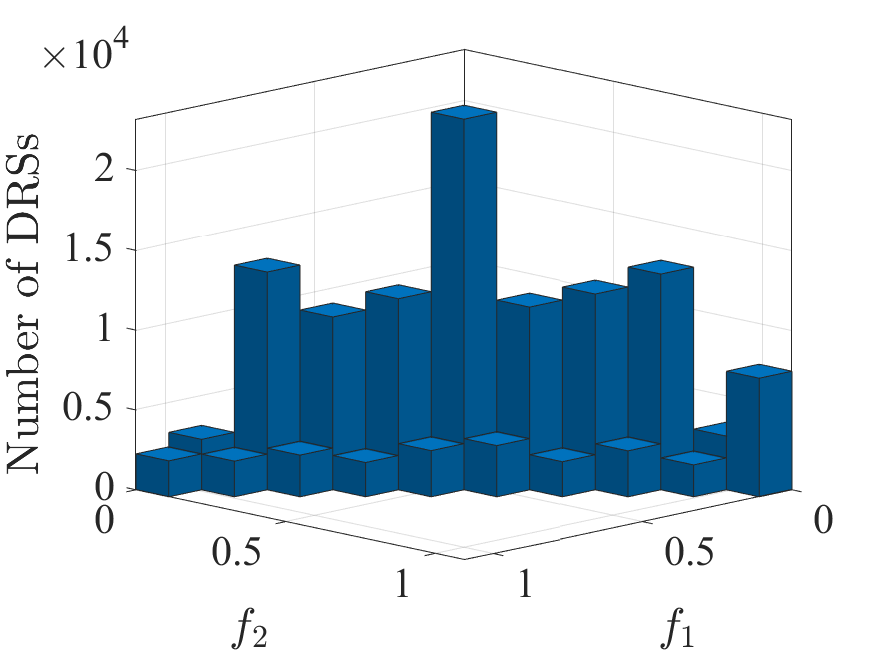}}
\hfil
\subfloat[IMOEA-ARP on EMOP2]{\includegraphics[width=0.32\linewidth]{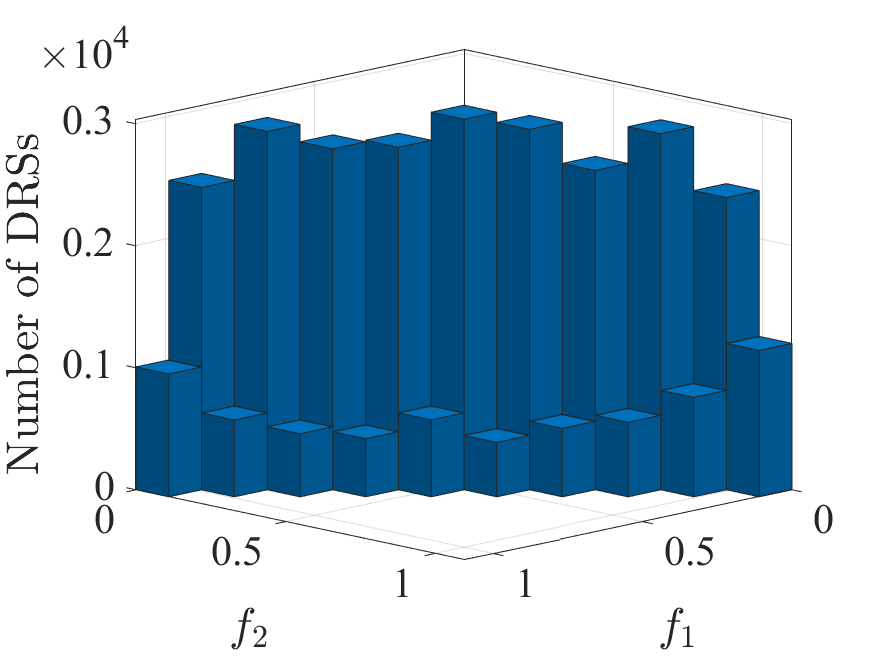}}
\hfil
\vspace{-1ex}
\subfloat[MultiGPO on EMOP3]{\includegraphics[width=0.32\linewidth]{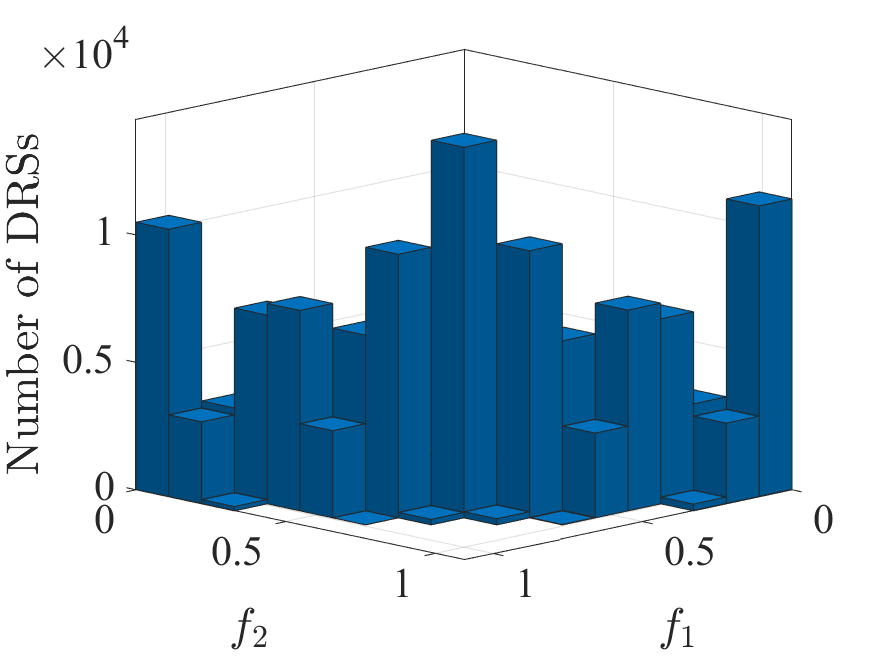}}
\hfil
\subfloat[MOEA/D-Gen on EMOP3]{\includegraphics[width=0.32\linewidth]{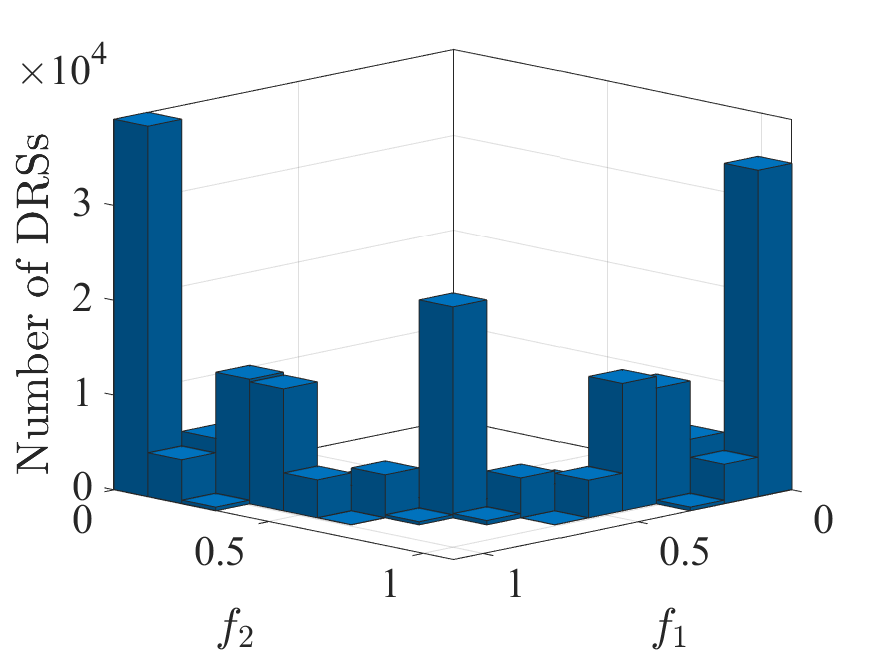}}
\hfil
\subfloat[IMOEA-ARP on EMOP3]{\includegraphics[width=0.32\linewidth]{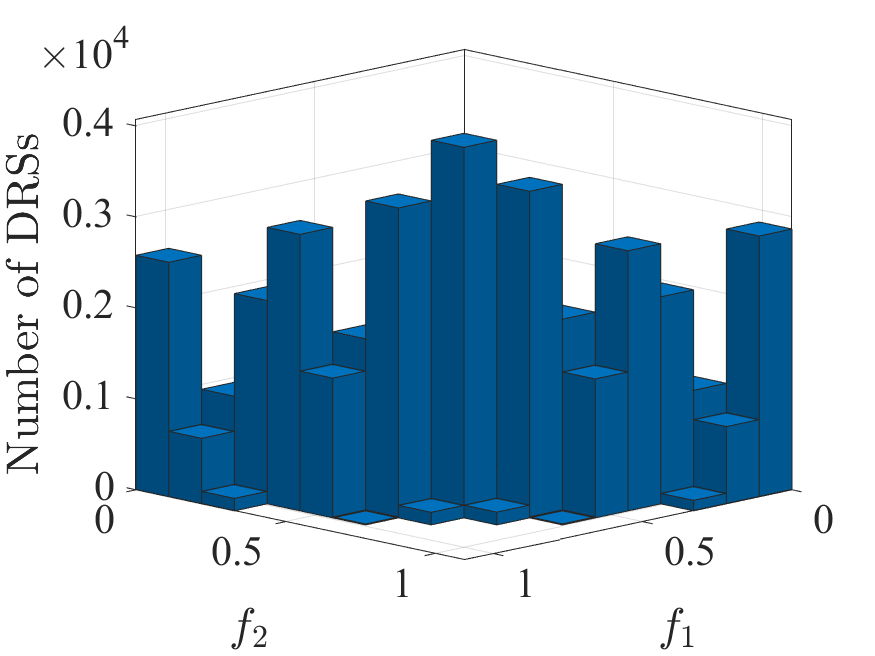}}
\hfil
\caption{Histograms illustrating the number of DRSs near the $WPB_{2,1}$ across all iterations in the 3-objective case.}
\label{fig:hist_drs}
\end{figure*}

\subsubsection{Performance Evaluation}
The number of DRSs close to the $WPB_{\nu,i}$ (denoted as $\gamma_\nu$) is reported. Specifically, the population is recorded at each iteration. 
In the set containing populations across all iterations and all runs, the objective vectors satisfying $\Delta<0.05$ for some $WPB$ are considered close. $\gamma_\nu$ is the number of DRSs close to one of $WPB_{\nu,i}$ for $i = 1,\ldots, C_m^\nu$.

Besides, the inverted generational distance (IGD) metric~\cite{coello2005solving} is employed. This metric quantifies the distance between the approximate set obtained by the MOEA and a given baseline set. The two sets are normalized based on the ideal objective vector and the nadir objective vector before calculating the IGD metric value. The given baseline set is usually set to the $PF$. To assess the degradation of the algorithm's performance, the baseline set is specified as the ideal set obtained by the MOEA, containing 30 times the population size of objective vectors for each instance. Specifically, we construct a corresponding instance containing only the position function for each instance in Table~\ref{tab:ins_analysis_paras}. Any sampled solution is Pareto-optimal in these modified instances. The ideal set is constructed as the union of the final populations obtained from 30 independent runs of the MOEA. It represents the ideal distribution of the objective vectors obtained by the MOEA on the $PF$. We denote this modified IGD metric as the IGD$\downarrow$ metric. Under this setting, a lower IGD$\downarrow$ metric value indicates less performance degradation.

\begin{table}[ht]
  \centering
  \caption{
  $\gamma_\nu$ of MultiGPO, MOEA/D-Gen, and IMOEA-ARP on EMOP1--EMOP3.
  % Total number of DRSs near the $WPB_{\nu,i}$ across all iterations.
  }
  % \vspace{-2ex}
    \setlength{\tabcolsep}{1mm}{
    \begin{threeparttable}
    \begin{tabular}{ccccccc}
    \toprule
    Algorithm    & Problem     & $m$     & $\gamma_1$  & $\gamma_2$  & $\gamma_3$  & $\gamma_4$ \\
    \midrule
    \multirow{9}[6]{*}{MultiGPO} & \multirow{3}[2]{*}{EMOP1} & 3     & 5.91e+03 & \cellcolor[rgb]{ .749,  .749,  .749}2.04e+05 & NA & NA \\
          &       & 4     & 5.93e+03 & 2.56e+04 & \cellcolor[rgb]{ .749,  .749,  .749}1.29e+06 & NA \\
          &       & 5     & 5.16e+03 & 1.76e+04 & 2.99e+05 & \cellcolor[rgb]{ .749,  .749,  .749}3.37e+06 \\
\cmidrule{2-7}          & \multirow{3}[2]{*}{EMOP2} & 3     & 1.63e+04 & \cellcolor[rgb]{ .749,  .749,  .749}3.02e+05 & NA & NA \\
          &       & 4     & 2.39e+03 & 3.86e+04 & \cellcolor[rgb]{ .749,  .749,  .749}1.72e+06 & NA \\
          &       & 5     & 4.35e+02 & 2.14e+04 & 3.91e+05 & \cellcolor[rgb]{ .749,  .749,  .749}3.90e+06 \\
\cmidrule{2-7}          & \multirow{3}[2]{*}{EMOP3} & 3     & 1.87e+03 & \cellcolor[rgb]{ .749,  .749,  .749}3.74e+05 & NA & NA \\
          &       & 4     & 5.59e+02 & 4.79e+04 & \cellcolor[rgb]{ .749,  .749,  .749}1.61e+06 & NA \\
          &       & 5     & 1.64e+02 & 2.46e+04 & 3.57e+05 & \cellcolor[rgb]{ .749,  .749,  .749}4.18e+06 \\
    \midrule
    \multirow{9}[6]{*}{MOEA/D-Gen} & \multirow{3}[2]{*}{EMOP1} & 3     & 1.76e+04 & \cellcolor[rgb]{ .749,  .749,  .749}1.01e+05 & NA & NA \\
          &       & 4     & 4.24e+04 & 2.16e+04 & \cellcolor[rgb]{ .749,  .749,  .749}4.04e+05 & NA \\
          &       & 5     & 6.12e+04 & 3.26e+04 & 1.99e+05 & \cellcolor[rgb]{ .749,  .749,  .749}2.03e+06 \\
\cmidrule{2-7}          & \multirow{3}[2]{*}{EMOP2} & 3     & 9.60e+04 & \cellcolor[rgb]{ .749,  .749,  .749}3.81e+05 & NA & NA \\
          &       & 4     & 3.45e+04 & 4.21e+04 & \cellcolor[rgb]{ .749,  .749,  .749}8.71e+05 & NA \\
          &       & 5     & 1.31e+04 & 4.27e+04 & 2.76e+05 & \cellcolor[rgb]{ .749,  .749,  .749}2.74e+06 \\
\cmidrule{2-7}          & \multirow{3}[2]{*}{EMOP3} & 3     & 3.67e+03 & \cellcolor[rgb]{ .749,  .749,  .749}5.07e+05 & NA & NA \\
          &       & 4     & 3.78e+03 & 6.94e+04 & \cellcolor[rgb]{ .749,  .749,  .749}1.04e+06 & NA \\
          &       & 5     & 6.10e+02 & 6.08e+04 & 3.26e+05 & \cellcolor[rgb]{ .749,  .749,  .749}3.26e+06 \\
    \midrule
    \multirow{3}[2]{*}{IMOEA-ARP} & EMOP1 & 3     & 4.39e+03 & \cellcolor[rgb]{ .749,  .749,  .749}7.63e+04 & NA & NA \\
          & EMOP2 & 3     & 4.38e+03 & \cellcolor[rgb]{ .749,  .749,  .749}1.09e+05 & NA & NA \\
          & EMOP3 & 3     & 6.00e+02 & \cellcolor[rgb]{ .749,  .749,  .749}1.16e+05 & NA & NA \\
    \bottomrule
    \end{tabular}%
    \begin{tablenotes}
        % \item[] ``\textbackslash'' in the cell indicates that the metric is not applicable to the corresponding instance.
         \item[] The largest value on an instance is highlighted using light gray shading.
    \end{tablenotes}
    \end{threeparttable}
    }
  \label{tab:n_drs_nu}%
\end{table}%

\begin{table}[ht]
  \centering
  \caption{IGD$\downarrow$ metric results on EMOP1--EMOP3 obtained by MultiGPO, MOEA/D-Gen, and IMOEA-ARP.}
    \setlength{\tabcolsep}{1.9mm}{
    \begin{threeparttable}
    \begin{tabular}{cccccc}
    \toprule
    Instance & $m$     & IGD$\downarrow$   & MultiGPO & MOEA/D-Gen & IMOEA-ARP \\
    \midrule
    \multirow{6}[2]{*}{EMOP1} & \multirow{2}[1]{*}{3} & mean  & 0.05112 & 0.0319 & 0.05201 \\
          &       & std.  & 0.00447 & 0.03334 & 0.0008254 \\
          & \multirow{2}[0]{*}{4} & mean  & \cellcolor[rgb]{ .749,  .749,  .749}0.1119 & \cellcolor[rgb]{ .502,  .502,  .502}0.1366 & NA \\
          &       & std.  & \cellcolor[rgb]{ .749,  .749,  .749}0.002738 & \cellcolor[rgb]{ .749,  .749,  .749}0.01023 & NA \\
          & \multirow{2}[1]{*}{5} & mean  & \cellcolor[rgb]{ .502,  .502,  .502}0.1623 & \cellcolor[rgb]{ .749,  .749,  .749}0.07244 & NA \\
          &       & std.  & \cellcolor[rgb]{ .502,  .502,  .502}0.004613 & \cellcolor[rgb]{ .502,  .502,  .502}0.01877 & NA \\
    \midrule
    \multirow{6}[2]{*}{EMOP2} & \multirow{2}[1]{*}{3} & mean  & 0.03982 & 0.0739 & 0.03457 \\
          &       & std.  & 0.003306 & 0.0134 & 0.0006206 \\
          & \multirow{2}[0]{*}{4} & mean  & \cellcolor[rgb]{ .749,  .749,  .749}0.08649 & \cellcolor[rgb]{ .502,  .502,  .502}0.1459 & NA \\
          &       & std.  & \cellcolor[rgb]{ .502,  .502,  .502}0.01405 & \cellcolor[rgb]{ .749,  .749,  .749}0.0242 & NA \\
          & \multirow{2}[1]{*}{5} & mean  & \cellcolor[rgb]{ .502,  .502,  .502}0.1195 & \cellcolor[rgb]{ .749,  .749,  .749}0.1165 & NA \\
          &       & std.  & \cellcolor[rgb]{ .749,  .749,  .749}0.008665 & \cellcolor[rgb]{ .502,  .502,  .502}0.05279 & NA \\
    \midrule
    \multirow{6}[2]{*}{EMOP3} & \multirow{2}[1]{*}{3} & mean  & 0.03475 & 0.06387 & 0.02709 \\
          &       & std.  & 0.006019 & 0.01059 & 0.007032 \\
          & \multirow{2}[0]{*}{4} & mean  & \cellcolor[rgb]{ .749,  .749,  .749}0.05504 & \cellcolor[rgb]{ .749,  .749,  .749}0.08862 & NA \\
          &       & std.  & \cellcolor[rgb]{ .749,  .749,  .749}0.003595 & \cellcolor[rgb]{ .749,  .749,  .749}0.02447 & NA \\
          & \multirow{2}[1]{*}{5} & mean  & \cellcolor[rgb]{ .502,  .502,  .502}0.07196 & \cellcolor[rgb]{ .502,  .502,  .502}0.1211 & NA \\
          &       & std.  & \cellcolor[rgb]{ .502,  .502,  .502}0.004771 & \cellcolor[rgb]{ .502,  .502,  .502}0.03916 & NA \\
    \bottomrule
    \end{tabular}%
    \begin{tablenotes}
        % \item[] ``\textbackslash'' in the cell indicates that the metric is not applicable to the corresponding instance.
        \item[] The worst and second-worst values across different settings of $m$ are highlighted using gray and light gray shading, respectively.
    \end{tablenotes}
    \end{threeparttable}
    }
  \label{tab:igd_nu}%
\end{table}%

\subsection{Experimental Results}\label{ssec:impact_nu}
This section investigates the impact of $WPB$s of different categories and shapes from an experimental perspective. EMOP1--EMOP3 in Table~\ref{tab:ins_analysis_paras} are used, which represent three distinct types of continuous $PF$s (convex, linear, and concave $PF$s, respectively). The cases involving 3, 4, and 5 objectives are considered. The computational time of calculating the hypervolume in IMOEA-ARP increases significantly for the case with more than 3 objectives~\cite{while2012fast}. Although several methods exist for estimating the hypervolume~\cite{shang2020new,boelrijk2023multi}, imprecise hypervolume metric values can degrade the performance in handling DRSs~\cite{wang2019scalable}. Therefore, the results of IMOEA-ARP are reported only for the 3-objective case.

The results in terms of $\gamma_\nu$ are shown in Table~\ref{tab:n_drs_nu}. In most cases, $\gamma_\nu$ with a larger $\nu$ tends to be greater, indicating that more objective vectors gather near the $WPB_{\nu,i}$ with a larger $\nu$. There are only two exceptions: $\gamma_1>\gamma_2$ for MOEA/D-Gen on 4- and 5-objective EMOP1. These exceptions may be attributed to the predefined weight vector distribution in MOEA/D-Gen when coping with the convex $PF$ of EMOP1. The convexity of the $PF$ significantly impacts the performance of the decomposition-based method with predefined weight vectors, as reported in~\cite{ishibuchi2017performance,wang2024multi}. 
It is also observed that $\gamma_{m-1}$ for EMOP3 exhibits the largest value among EMOP1--EMOP3. This suggests that when the $PF$ is concave, the corresponding $WPB_{m-1,i}$ exerts a stronger attraction on the population of the MOEA. 
Table~\ref{tab:igd_nu} shows the results of the IGD$\downarrow$ metric. As the number of objectives increases, the performance and stability of both MultiGPO and MOEA/D-Gen deteriorate more noticeably. It highlights that an increase in $\nu$ exacerbates the negative impact of the $WPB_{\nu,i}$.

In addition, \figurename~\ref{fig:hist_drs} provides histograms describing the distributions of the number of DRSs near the $WPB_{2,1}$ with respect to $f_1$ and $f_2$ in 3-objective cases. These experimental results exhibit a strong correlation with the theoretical findings presented in \figurename~\ref{fig:vol_p1_m3_13_WPB2_m2b},~\ref{fig:vol_p2_m3_13_WPB2_m2b}, and~\ref{fig:vol_p05_m3_13_WPB2_m2b}. On EMOP1 and EMOP3, three MOEAs consistently retain many DRSs near the middle of the $WPB_{2,1}$, which suggests that the DRS located near the middle of the convex or concave $WPB_{2,i}$ is harder to eliminate than those in most positions. The number of DRSs near the middle remains high on EMOP2; however, the disparity between the middle and other positions is less significant than the results observed on EMOP1 and EMOP3. We also find that three MOEAs tend to retain considerable DRSs near the boundaries of the concave $WPB_{2,1}$.

Overall, the DRS close to the $WPB_{\nu,i}$ with a larger $\nu$, exhibiting a generally higher DDR, presents greater challenges for elimination and degrades the algorithm's performance. Besides, the DRS located near the middle of the $WPB_{2,i}$ is difficult to eliminate for the 3-objective case with a convex or concave $PF$. The experimental results align with all theoretical findings in Section~\ref{sec:analysis}. 

\section{Conclusion}\label{sec:conclusion}
This paper provides a comprehensive study of the $WPB$ from both theoretical and experimental perspectives. The $WPB$ is defined and categorized into different $WPB_{\nu,i}$. The theoretical analysis reveals that the $WPB_{\nu,i}$ with a larger $\nu$ makes the nearby DRS have a generally higher DDR, which brings a greater challenge to the MOEA. Moreover, it is found that the shape of the $WPB$ significantly affects DDRs of its nearby DRSs.
Subsequently, two test problem generators are proposed to facilitate a holistic evaluation of the $WPB$'s impact on MOEAs. 16 example instances are generated, and three representative MOEAs are tested. Experimental results on several example instances confirm the greater challenge brought by the $WPB_{\nu,i}$ with a larger $\nu$ and the effects of different shapes' $WPB$s. Further experiments on other example instances highlight the challenges posed by the overall size of the $WPB$, the relative sizes among $WPB$s, and the $WPB$ located at the discontinuous part of the $PF$. In addition, a set of benchmark instances is generated to examine more MOEAs. All these experiments indicate that existing algorithms fail to achieve satisfactory performance across all scenarios.

Future work will build upon the findings of this paper by incorporating broader settings in the theoretical analysis and additional problem characteristics in the experiments, and developing strategies to comprehensively cope with the challenges posed by the $WPB$.

% \section*{Acknowledgments}
% This should be a simple paragraph before the References to thank those individuals and institutions who have supported your work on this article.

% {
% \appendix[Proof of the Zonklar Equations]
% Use $\backslash${\tt{appendix}} if you have a single appendix:
% Do not use $\backslash${\tt{section}} anymore after $\backslash${\tt{appendix}}, only $\backslash${\tt{section*}}.
% If you have multiple appendixes use $\backslash${\tt{appendices}} then use $\backslash${\tt{section}} to start each appendix.
% You must declare a $\backslash${\tt{section}} before using any $\backslash${\tt{subsection}} or using $\backslash${\tt{label}} ($\backslash${\tt{appendices}} by itself
%  starts a section numbered zero.)
% }


\bibliographystyle{IEEEtran}
\bibliography{references}

\clearpage
\newpage
\appendices

This is the supplementary document for ``Weak Pareto Boundary: The Achilles' Heel of Evolutionary Multi-Objective Optimization''.

\section{Impact of \texorpdfstring{$WPB$s}{WPBs} of Different Sizes}\label{sec:impact_wpb_size}
This section tests MOEAs on MOPs with $WPB$s of different sizes. The experimental settings are the same as those in Section~\ref{sec:experiment}. The 3-objective case is considered. In such a case, the overall size of both the $WPB_{1,i}$ and the $WPB_{2,i}$ can be adjusted by $\ell_i$, while their relative sizes are controlled by $d_i$. 10 example instances, EMOP4--EMOP13, are generated and shown in Table~\ref{tab:ins_analysis_paras}. From EMOP4 to EMOP8, the overall size of the $WPB$ is getting larger. From EMOP9 to EMOP13, the size of the $WPB_{2,i}$ relative to $WPB_{1,i}$ increases.

\begin{figure*}[ht]
\centering
\subfloat[MultiGPO]{\includegraphics[width=0.32\linewidth]{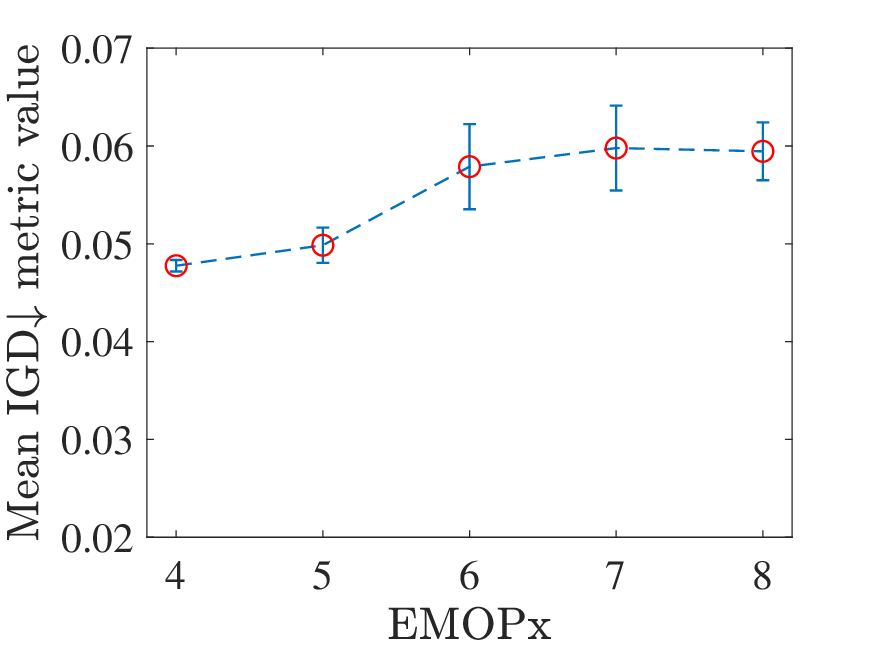}}
\hfil
\subfloat[MOEA/D-Gen]{\includegraphics[width=0.32\linewidth]{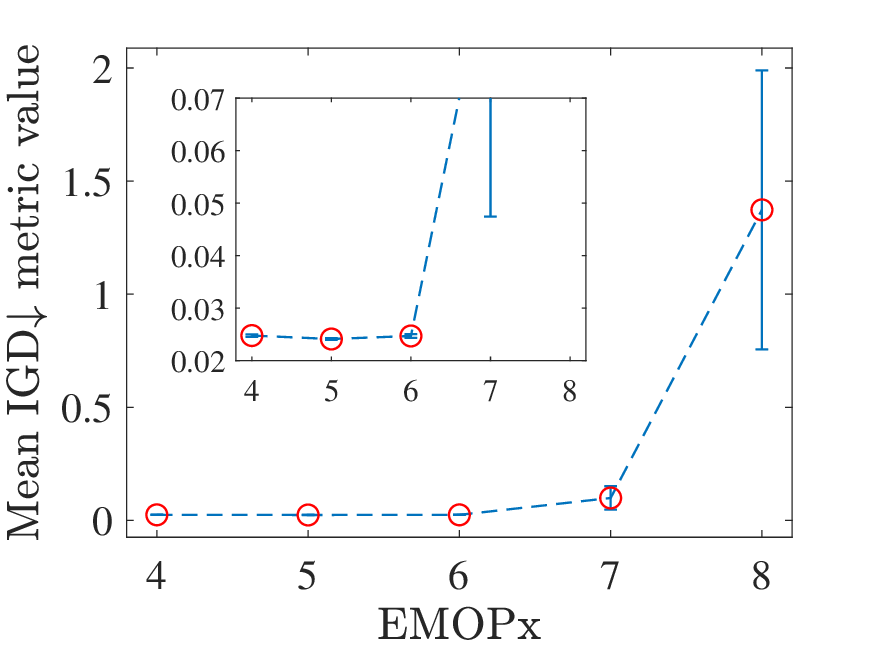}}
\hfil
\subfloat[IMOEA-ARP]{\includegraphics[width=0.32\linewidth]{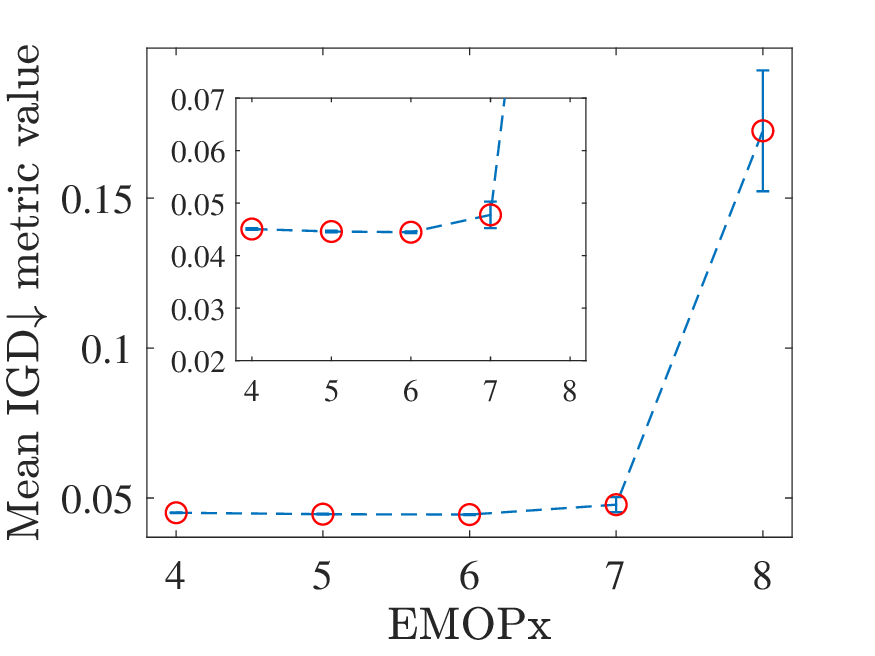}}
\hfil
\caption{Mean IGD$\downarrow$ metric values of MultiGPO, MOEA/D-Gen, and IMOEA-ARP on EMOP4--EMOP8. Error bars represent the standard error of the mean. The best and worst values of the IGD$\downarrow$ metric are excluded.}
\label{fig:errorbar_EMOP4_8}
\end{figure*}

The results on EMOP4--EMOP8 are shown in Table~\ref{tab:n_drs_length} and \figurename~\ref{fig:errorbar_EMOP4_8}. As the overall size of the $WPB$ increases, both $\gamma_1$ and $\gamma_2$ increase, leading to a corresponding increase in $\gamma_1 + \gamma_2$. Among the three MOEAs, the values of $\gamma_1 + \gamma_2$ for MultiGPO have the least variation. The value of $\frac{\gamma_2}{\gamma_1}$ indicates the relative impact of the $WPB_{2,i}$ on the MOEA compared to the $WPB_{1,i}$. For MultiGPO, the values of $\frac{\gamma_2}{\gamma_1}$ are similar across EMOP4 to EMOP8. In contrast, for MOEA/D-Gen, the relative impact of the $WPB_{2,i}$ becomes smaller from EMOP4 to EMOP8, while for IMOEA-ARP, the relative impact is getting greater.
According to the statistical results regarding the IGD$\downarrow$ metric, the performance degradation of MultiGPO is relatively minor, irrespective of changes in the overall size of the $WPB$. However, MOEA/D-Gen exhibits a sharp decline in performance on EMOP7 and EMOP8, and IMOEA-ARP experiences a substantial performance decline on EMOP8. It indicates that the cone-dominance-based method, MultiGPO, is not significantly affected by the overall size of the $WPB$, as discussed in~\cite{wang2024multi}.

\begin{figure*}[ht]
\centering
\subfloat[MultiGPO]{\includegraphics[width=0.32\linewidth]{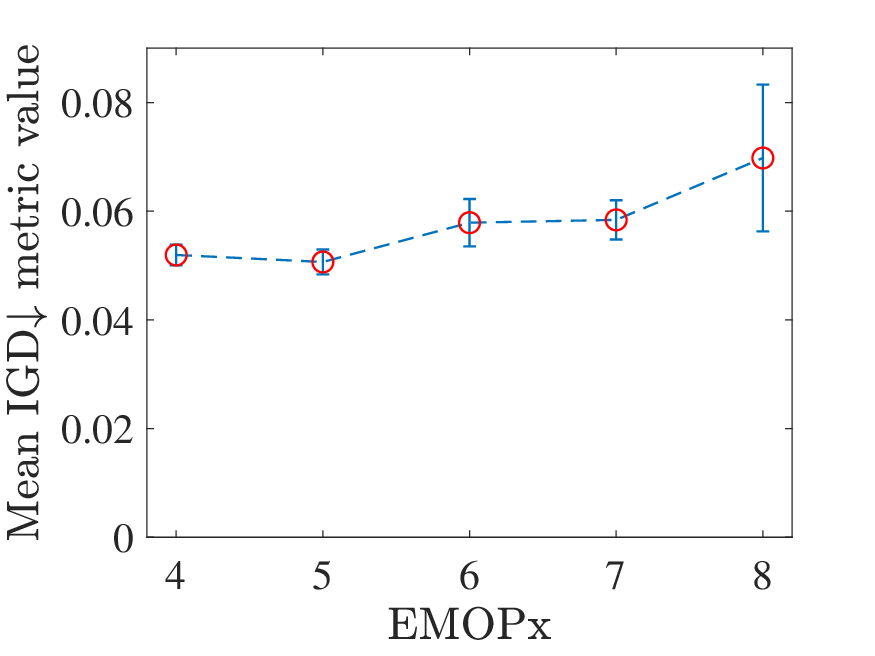}}
\hfil
\subfloat[MOEA/D-Gen]{\includegraphics[width=0.32\linewidth]{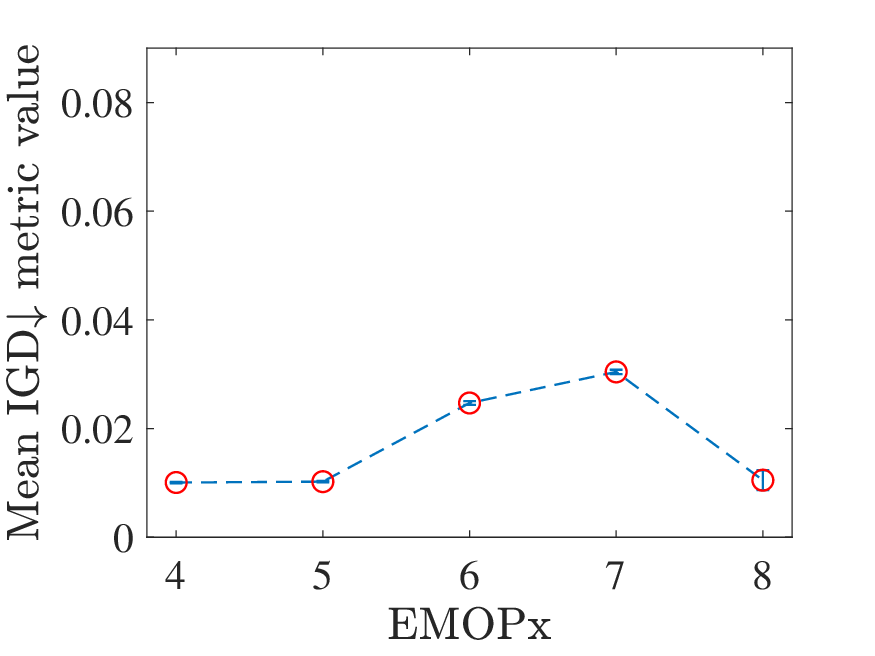}}
\hfil
\subfloat[IMOEA-ARP]{\includegraphics[width=0.32\linewidth]{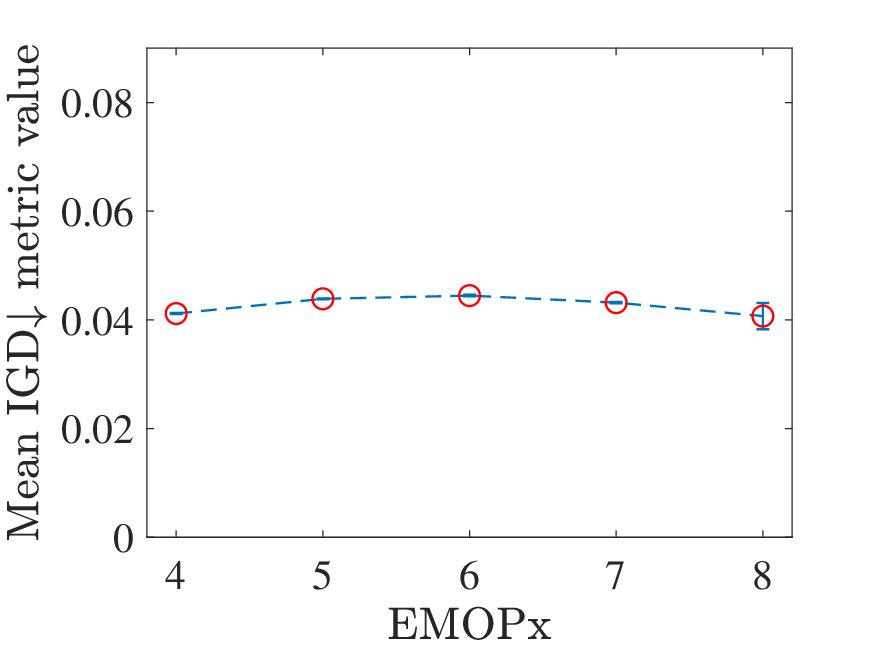}}
\hfil
\caption{Mean IGD$\downarrow$ metric values of MultiGPO, MOEA/D-Gen, and IMOEA-ARP on EMOP9--EMOP13. Error bars represent the standard error of the mean. The best and worst values of the IGD$\downarrow$ metric are excluded.}
\label{fig:errorbar_EMOP9_13}
\end{figure*}

The results on EMOP9--EMOP13 are reported in Table~\ref{tab:n_drs_width} and \figurename~\ref{fig:errorbar_EMOP9_13}. As the relative size of the $WPB_{2,i}$ increases, the resulting values of $\gamma_2$ consistently increase for any candidate MOEA. The values of $\frac{\gamma_2}{\gamma_1}$ also show a consistent increase for each candidate MOEA. $\gamma_1$ consistently decreases for both MultiGPO and IMOEA-ARP, while the changes in $\gamma_1$ exhibit no clear pattern for MOEA/D-Gen.
On these instances, MultiGPO demonstrates the largest performance degradation. Specifically, the mean IGD$\downarrow$ metric values obtained by MultiGPO are the largest among the three algorithms, and show a gradually rising trend. In addition, MultiGPO exhibits considerable performance variability on EMOP13. Compared to MultiGPO, MOEA/D-Gen exhibits the least performance deterioration on these instances. IMOEA-ARP demonstrates the most stable performance.

\begin{table}[t]
  \centering
  \caption{
  $\gamma_\nu$ of MultiGPO, MOEA/D-Gen, and IMOEA-ARP on EMOP4--EMOP8 ($m=3$).
  % Total number of DRSs near the $WPB_{\nu,i}$ across all iterations.
  }
    \setlength{\tabcolsep}{1.6mm}{
    \begin{threeparttable}
    \begin{tabular}{cccccc}
    \toprule
    Algorithm    & Instance     & $\gamma_1$  & $\gamma_2$  & $\gamma_1+\gamma_2$  & $~~\frac{\gamma_2}{\gamma_1}~~$ \\
    \midrule
    \multirow{5}[2]{*}{MultiGPO} & EMOP4 & 1.26e+05 & 1.12e+05 & 2.38e+05 & 0.9  \\
          & EMOP5 & 1.08e+05 & 1.10e+05 & 2.19e+05 & 1.0  \\
          & EMOP6 & 1.56e+05 & 1.40e+05 & 2.96e+05 & 0.9  \\
          & EMOP7 & \cellcolor[rgb]{ .749,  .749,  .749}1.96e+05 & \cellcolor[rgb]{ .749,  .749,  .749}1.59e+05 & \cellcolor[rgb]{ .749,  .749,  .749}3.55e+05 & 0.8  \\
          & EMOP8 & \cellcolor[rgb]{ .502,  .502,  .502}2.02e+05 & \cellcolor[rgb]{ .502,  .502,  .502}1.70e+05 & \cellcolor[rgb]{ .502,  .502,  .502}3.71e+05 & 0.8  \\
    \midrule
    \multirow{5}[2]{*}{MOEA/D-Gen} & EMOP4 & 5.82e+04 & 4.60e+04 & 1.04e+05 & 0.8  \\
          & EMOP5 & 9.52e+04 & 6.10e+04 & 1.56e+05 & 0.6  \\
          & EMOP6 & 1.74e+05 & 8.49e+04 & 2.58e+05 & 0.5  \\
          & EMOP7 & \cellcolor[rgb]{ .749,  .749,  .749}3.68e+05 & \cellcolor[rgb]{ .749,  .749,  .749}9.21e+04 & \cellcolor[rgb]{ .749,  .749,  .749}4.60e+05 & 0.3  \\
          & EMOP8 & \cellcolor[rgb]{ .502,  .502,  .502}6.51e+05 & \cellcolor[rgb]{ .502,  .502,  .502}1.03e+05 & \cellcolor[rgb]{ .502,  .502,  .502}7.54e+05 & 0.2  \\
    \midrule
    \multirow{5}[2]{*}{IMOEA-ARP} & EMOP4 & 3.40e+04 & 3.89e+04 & 7.29e+04 & 1.1  \\
          & EMOP5 & 5.96e+04 & 6.62e+04 & 1.26e+05 & 1.1  \\
          & EMOP6 & 7.51e+04 & 9.83e+04 & 1.73e+05 & 1.3  \\
          & EMOP7 & \cellcolor[rgb]{ .749,  .749,  .749}9.07e+04 & \cellcolor[rgb]{ .749,  .749,  .749}3.28e+05 & \cellcolor[rgb]{ .749,  .749,  .749}4.19e+05 & 3.6  \\
          & EMOP8 & \cellcolor[rgb]{ .502,  .502,  .502}1.14e+05 & \cellcolor[rgb]{ .502,  .502,  .502}7.11e+05 & \cellcolor[rgb]{ .502,  .502,  .502}8.26e+05 & 6.2  \\
    \bottomrule
    \end{tabular}%
    \begin{tablenotes}
        \item[] The largest and second-largest values across EMOP4--EMOP8 are highlighted using gray and light gray shading, respectively.
    \end{tablenotes}
    \end{threeparttable}
    }
  \label{tab:n_drs_length}%
\end{table}%

\begin{table}[t]
  \centering
  \caption{
  $\gamma_\nu$ of MultiGPO, MOEA/D-Gen, and IMOEA-ARP on EMOP9--EMOP13 ($m=3$).
  }
    \setlength{\tabcolsep}{1.6mm}{
    \begin{threeparttable}
    \begin{tabular}{cccccc}
    \toprule
    Algorithm    & Instance     & $\gamma_1$  & $\gamma_2$  & $\gamma_1+\gamma_2$  & $~~\frac{\gamma_2}{\gamma_1}~~$ \\
    \midrule
    \multirow{5}[2]{*}{MultiGPO} & EMOP9 & \cellcolor[rgb]{ .502,  .502,  .502}2.17e+05 & 4.96e+04 & 2.67e+05 & 0.2  \\
          & EMOP10 & \cellcolor[rgb]{ .749,  .749,  .749}1.82e+05 & 8.50e+04 & 2.67e+05 & 0.5  \\
          & EMOP11 & 1.56e+05 & 1.40e+05 & 2.96e+05 & 0.9  \\
          & EMOP12 & 9.80e+04 & \cellcolor[rgb]{ .749,  .749,  .749}2.30e+05 & \cellcolor[rgb]{ .749,  .749,  .749}3.28e+05 & 2.3  \\
          & EMOP13 & 2.40e+04 & \cellcolor[rgb]{ .502,  .502,  .502}3.73e+05 & \cellcolor[rgb]{ .502,  .502,  .502}3.97e+05 & 15.5  \\
    \midrule
    \multirow{5}[2]{*}{MOEA/D-Gen} & EMOP9 & \cellcolor[rgb]{ .749,  .749,  .749}2.04e+05 & 1.92e+04 & 2.23e+05 & 0.1  \\
          & EMOP10 & 1.66e+05 & 4.40e+04 & 2.10e+05 & 0.3  \\
          & EMOP11 & 1.74e+05 & 8.49e+04 & 2.58e+05 & 0.5  \\
          & EMOP12 & 1.78e+05 & \cellcolor[rgb]{ .749,  .749,  .749}1.68e+05 & \cellcolor[rgb]{ .749,  .749,  .749}3.45e+05 & 0.9  \\
          & EMOP13 & \cellcolor[rgb]{ .502,  .502,  .502}4.30e+05 & \cellcolor[rgb]{ .502,  .502,  .502}4.99e+05 & \cellcolor[rgb]{ .502,  .502,  .502}9.29e+05 & 1.2  \\
    \midrule
    \multirow{5}[2]{*}{IMOEA-ARP} & EMOP9 & \cellcolor[rgb]{ .502,  .502,  .502}1.25e+05 & 3.81e+04 & 1.63e+05 & 0.3  \\
          & EMOP10 & \cellcolor[rgb]{ .749,  .749,  .749}9.65e+04 & 6.24e+04 & 1.59e+05 & 0.6  \\
          & EMOP11 & 7.51e+04 & 9.83e+04 & 1.73e+05 & 1.3  \\
          & EMOP12 & 4.83e+04 & \cellcolor[rgb]{ .749,  .749,  .749}1.50e+05 & \cellcolor[rgb]{ .749,  .749,  .749}1.98e+05 & 3.1  \\
          & EMOP13 & 1.57e+04 & \cellcolor[rgb]{ .502,  .502,  .502}3.50e+05 & \cellcolor[rgb]{ .502,  .502,  .502}3.66e+05 & 22.3  \\
    \bottomrule
    \end{tabular}%
    \begin{tablenotes}
        \item[] See notes for Table~\ref{tab:n_drs_length}.
    \end{tablenotes}
    \end{threeparttable}
    }
  \label{tab:n_drs_width}%
\end{table}%

Overall, an increase in the size of the $WPB$ leads to a greater number of DRSs retained in the population, as well as a corresponding decline in the performance and stability of the MOEA. MultiGPO is more sensitive to variations in the relative size of the $WPB_{2,i}$, whereas MOEA/D-Gen and IMOEA-ARP are more sensitive to changes in the overall size of the $WPB$.

\section{Impact of the Spatial Relation of the \texorpdfstring{$WPB$}{WPB}}\label{sec:impact_disc_pf}
This section tests MOEAs on MOPs with discontinuous $PF$s. The experimental settings are the same as those in Section~\ref{sec:experiment}. EMOP14--EMOP16 in Table~\ref{tab:ins_analysis_paras} are used, which have linear, convex, and concave $PF$s, respectively. The discontinuous $PF$ comprises four parts in the 3-objective case: one middle part and three boundary parts.

\begin{table}[ht]
  \centering
  \caption{IGD$\downarrow$ metric results on EMOP1--EMOP3 obtained by MultiGPO, MOEA/D-Gen, and IMOEA-ARP.}
    \setlength{\tabcolsep}{2.2mm}{
    \begin{threeparttable}
    \begin{tabular}{cccccc}
    \toprule
    Instance & IGD$\downarrow$   & MultiGPO & MOEA/D-Gen & IMOEA-ARP \\
    \midrule
    \multirow{2}[1]{*}{EMOP14} & mean  & \cellcolor[rgb]{ .749,  .749,  .749}0.05241 & \cellcolor[rgb]{ .749,  .749,  .749}0.0758 & \cellcolor[rgb]{ .749,  .749,  .749}0.06434 \\
          & std.  & \cellcolor[rgb]{ .749,  .749,  .749}0.02694 & \cellcolor[rgb]{ .749,  .749,  .749}0.01972 & \cellcolor[rgb]{ .749,  .749,  .749}0.1148 \\
    \multirow{2}[1]{*}{EMOP15} & mean  & 0.01464 & 0.001145 & 0.01201 \\
          & std.  & 0.004373 & 0.001169 & 0.0002844 \\
    \multirow{2}[1]{*}{EMOP16} & mean  & 0.03331 & 0.002926 & 0.01484 \\
          & std.  & 0.001338 & 0.0007314 & 0.0005518 \\
    \bottomrule
    \end{tabular}%
    \begin{tablenotes}
        \item[] The largest value across EMOP14--EMOP16 is highlighted using light gray shading.
    \end{tablenotes}
    \end{threeparttable}
    }
  \label{tab:igd_disc}%
\end{table}%

\begin{figure*}[ht]
\centering
\subfloat[MultiGPO]{\includegraphics[width=0.32\linewidth]{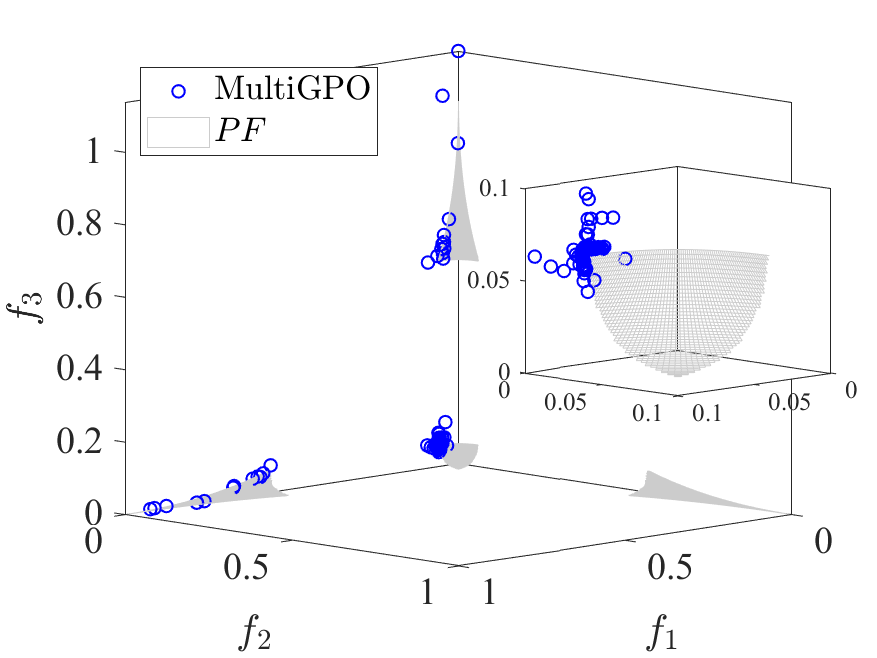}}
\hfil
\subfloat[MOEA/D-Gen]{\includegraphics[width=0.32\linewidth]{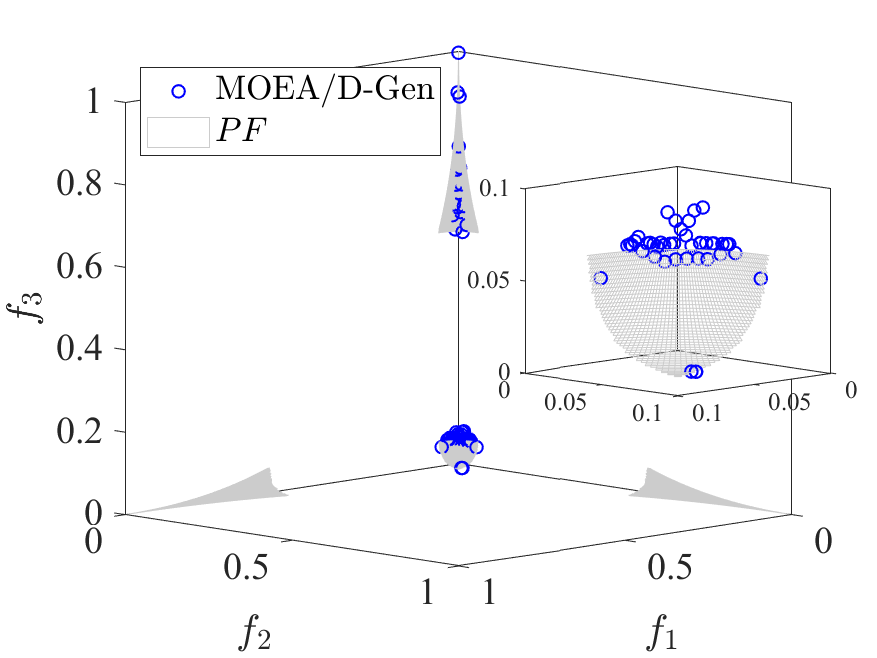}}
\hfil
\subfloat[IMOEA-ARP]{\includegraphics[width=0.32\linewidth]{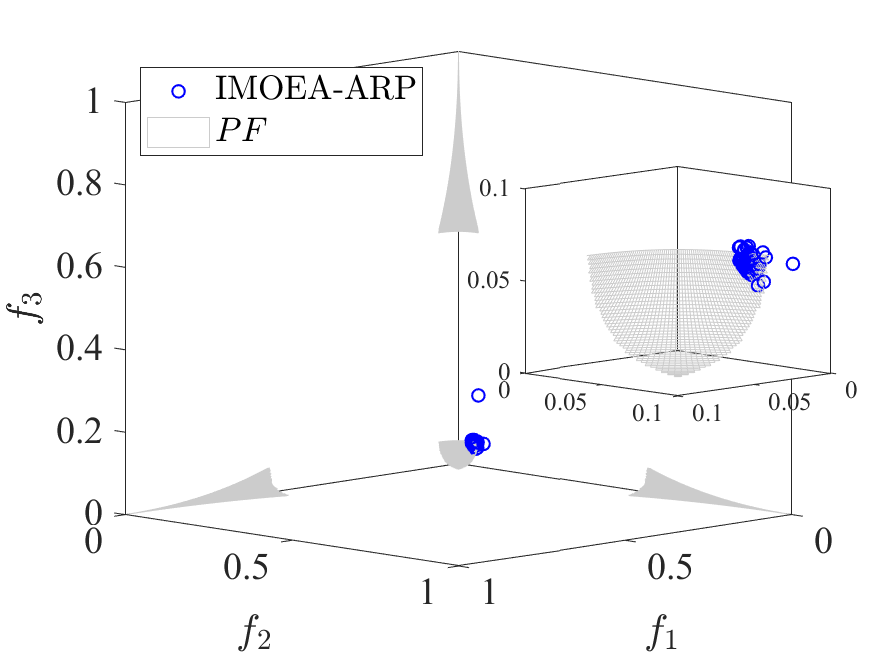}}
\hfil
\caption{Plots of the final populations in the objective space with the second-worst IGD$\downarrow$ metric values across 30 runs on EMOP12.}
\label{fig:objs_disc}
\end{figure*}

Table~\ref{tab:igd_disc} showcases the statistical IGD$\downarrow$ metric values. The three MOEAs demonstrate considerable performance degradation and instability on EMOP14. Among the three algorithms, MOEA/D-Gen exhibits the highest mean IGD$\downarrow$ metric value on EMOP14, whereas IMOEA-ARP shows the largest standard deviation. In comparison, the three MOEAs exhibit much less performance degradation and greater stability on EMOP15 and EMOP16.
The final populations of the MOEAs on EMOP11 are depicted in \figurename~\ref{fig:objs_disc}. MultiGPO fails to find any Pareto-optimal objective vector on the right boundary part. Many DRSs are preserved within the final population of MultiGPO. MOEA/D-Gen's population misses two boundary parts of the $PF$ and also maintains many DRSs. IMOEA-ARP cannot find any boundary part of the $PF$, and its population has poor distribution on the middle part of the $PF$. These observations arise because the boundary parts in EMOP14 resemble the boundary of the extremely convex $PF$, which is often mistakenly identified as a set of DRSs~\cite{yang2021hard}. The techniques used by MultiGPO and MOEA/D-Gen to identify DRSs depend on specific parameter settings (\ie, $\delta_i$ in MultiGPO and $\rho$ in MOEA/D-Gen). In this experiment, both algorithms may incorrectly classify the boundary parts as DRSs under default parameter settings. IMOEA-ARP is parameter-free. The reason for its poor performance on EMOP14 may be the requirement for greater computational resources than usual. The lack of computational resources prevents IMOEA-ARP from effectively distinguishing between Pareto-optimal solutions and DRSs, leading to highly unstable performance.

Overall, the three algorithms exhibit unsatisfactory performance on the MOP with the convex and discontinuous $PF$, as their population distributions fail to achieve uniform coverage across the whole $PF$.

\section{Benchmarking Advanced Algorithms}\label{sec:benchmarking}

\subsection{Experimental Setup}
A set of 16 new test instances for benchmarking is generated and presented in Table~\ref{tab:ins_benchmark_paras}. Their corresponding features are also available in the table. We arrange these 16 instances in ascending order according to both the overall size and the relative size.

Except for the algorithms used in Section~\ref{sec:experiment}, three other advanced algorithms are selected for benchmarking in this section:
PREA~\cite{yuan2021investigating},
RVEA-iGNG~\cite{liu2022adaptive},
and LMPFE~\cite{tian2023local}.
Their parameters are set based on the corresponding references. 

The IGD metric is employed to assess the algorithm's performance for each instance. The baseline set is set to the $PF$. Since the $PF$ is not a discrete set, more than $C_{50+m-1}^{m-1}$ points are uniformly sampled from it. A lower value of the IGD metric signifies better performance.

\begin{table*}[ht]
  \centering
  \caption{Benchmark instances. $m=3$, $\mathbf{z}^{ide}=(1,2,3)^\intercal$, and $\mathbf{s}=(100,10,1)^\intercal$.}
    \setlength{\tabcolsep}{2.2mm}{
    \begin{threeparttable}
    \begin{tabular}{cccccccc}
    \toprule
    \multirow{2}[4]{*}{Instance} & \multirow{2}[4]{*}{$\mathbf{p}$} & \multirow{2}[4]{*}{$\boldsymbol{\ell}$} & \multirow{2}[4]{*}{$\mathbf{d}$} & \multirow{2}[4]{*}{$\mathbf{r}$} & \multicolumn{3}{c}{Characteristics} \\
\cmidrule{6-8}          &       &       &       &       & \makecell{Overall size of \\ the $WPB$} & \makecell{Relative size of \\ the $WPB_{2,i}$} & $PF$ \\
    \midrule
    MOPW1 & 2     & 4     & 0.5   & NA & small & large & convex \\
    MOPW2 & $(0.5,0.5,2)^\intercal$ & 4     & 0.5   & NA & small & large & mixed \\
    MOPW3 & 0.5   & 4     & 0.5   & NA & small & large & concave \\
    MOPW4 & 0.5   & 400   & 0.9   & NA & moderate & small & concave \\
    MOPW5 & 2     & 4     & 0.5   & 0.2   & small & large & convex, discontinuous \\
    MOPW6 & 0.5   & 4     & 0.9   & 2     & large & small & concave, discontinuous \\
    % \midrule
    MOPW7 & 2     & 400   & $(0.5,0.7,0.7)^\intercal$ & NA & moderate & \makecell[l]{large ($i=1$); \\ moderate ($i=2$); \\ moderate ($i=3$)} & convex \\
    MOPW8 & $(0.5,0.5,2)^\intercal$ & 400   & $(0.7,0.5,0.5)^\intercal$ & NA & moderate & \makecell[l]{moderate ($i=1$); \\ large ($i=2$); \\ large ($i=3$)} & mixed \\
    MOPW9 & 0.5   & 40000 & $(1,1,0.5)^\intercal$ & NA & large & \makecell[l]{small ($i=1$); \\ small ($i=2$); \\ large ($i=3$)} & concave \\
    MOPW10 & 0.5   & $(4,400,40000)^\intercal$ & 0.7   & NA & moderate & moderate & concave \\
    MOPW11 & $(0.5,0.5,2)^\intercal$ & 4     & 0.5   & $(2,0,0)^\intercal$ & moderate & large & mixed, discontinuous \\
    MOPW12 & 0.5   & 4     & $(1,0.5,0.5)^\intercal$ & 1     & moderate & \makecell[l]{small ($i=1$); \\ large ($i=2$); \\ large ($i=3$)} & concave, discontinuous \\
    % \midrule
    MOPW13 & 2     & 40000 & 0.5   & NA & large & large & convex \\
    MOPW14 & $(0.5,0.5,2)^\intercal$ & 40000 & 0.5   & NA & large & large & mixed \\
    MOPW15 & 0.5   & 40000 & 0.5   & NA & large & large & concave \\
    MOPW16 & 0.5   & 4     & 0.5   & 2     & large & large & concave, discontinuous \\
    \bottomrule
    \end{tabular}%
    \begin{tablenotes}
        \item[] ``NA'' in the cell with respect to $\mathbf{r}$ indicates that their corresponding instances are obtained by Generator 1 in Section~\ref{sec:problem}. The remaining instances are obtained by Generator 2.
        \item[] A scalar in the cell indicates that all elements of the corresponding parameter equal this scalar. For example, ``2'' for $\mathbf{p}$ means $\mathbf{p} = (2,2,2)^\intercal$.
    \end{tablenotes}
    \end{threeparttable}
    }
  \label{tab:ins_benchmark_paras}%
\end{table*}%

\begin{table*}[ht]
  \centering
  \caption{IGD metric results on MOPW1--MOPW16 obtained by 6 MOEAs.}
    \setlength{\tabcolsep}{3.8mm}{
    \begin{threeparttable}
    \begin{tabular}{cccccccc}
    \toprule
    Instance & IGD   & PREA  & RVEA-iGNG & LMPFE & MultiGPO & MOEA/D-Gen & IMOEA-ARP \\
    \midrule
    \multirow{2}[1]{*}{MOPW1} & mean  & 0.0745543(3)- & 0.0805912(4)- & 0.195671(6)- & \cellcolor[rgb]{ .749,  .749,  .749}0.0704968(2)- & 0.0917155(5)- & \cellcolor[rgb]{ .502,  .502,  .502}0.0552279(1) \\
          & std.  & 0.00508711 & 0.00638841 & 0.0188741 & 0.00322591 & 0.0103945 & 0.000281211 \\
    \multirow{2}[0]{*}{MOPW2} & mean  & 0.0720965(3)- & 0.0876977(4)- & 0.170268(6)- & \cellcolor[rgb]{ .502,  .502,  .502}0.0510354(1)+ & 0.117575(5)- & \cellcolor[rgb]{ .749,  .749,  .749}0.0561297(2) \\
          & std.  & 0.00776305 & 0.0133224 & 0.0216312 & 0.00536214 & 0.0139484 & 0.103324 \\
    \multirow{2}[0]{*}{MOPW3} & mean  & 0.0714596(3)- & 0.0860504(4)- & 0.16067(6)- & \cellcolor[rgb]{ .749,  .749,  .749}0.0384091(2)- & 0.110218(5)- & \cellcolor[rgb]{ .502,  .502,  .502}0.0280996(1) \\
          & std.  & 0.007467 & 0.0298631 & 0.0215119 & 0.00426935 & 0.0128707 & 0.000367253 \\
    \multirow{2}[0]{*}{MOPW4} & mean  & 0.672091(5)- & 0.467115(4)- & 0.975869(6)- & 0.0847353(3)- & \cellcolor[rgb]{ .749,  .749,  .749}0.0742992(2)- & \cellcolor[rgb]{ .502,  .502,  .502}0.0645138(1) \\
          & std.  & 0.135296 & 0.140306 & 0.479495 & 0.0116001 & 0.00105537 & 0.000952424 \\
    \multirow{2}[0]{*}{MOPW5} & mean  & \cellcolor[rgb]{ .749,  .749,  .749}0.0272187(2)+ & \cellcolor[rgb]{ .502,  .502,  .502}0.0244964(1)+ & 0.0289698(3)+ & 0.0994991(6)- & 0.096796(5)- & 0.0303376(4) \\
          & std.  & 0.00091234 & 0.000474931 & 0.00205367 & 0.0328696 & 0.000863114 & 0.000587597 \\
    \multirow{2}[1]{*}{MOPW6} & mean  & \cellcolor[rgb]{ .749,  .749,  .749}0.037832(2)+ & \cellcolor[rgb]{ .502,  .502,  .502}0.0361309(1)+ & 0.044101(6)= & 0.0407728(4)+ & 0.0400754(3)+ & 0.0431418(5) \\
          & std.  & 0.00109727 & 0.00128789 & 0.00326175 & 0.00206828 & 0.000197938 & 0.000840831 \\
    % \midrule
    \multirow{2}[1]{*}{MOPW7} & mean  & 0.57827(5)- & 0.531682(4)- & 2.32825(6)- & \cellcolor[rgb]{ .749,  .749,  .749}0.0886561(2)- & 0.0909368(3)- & \cellcolor[rgb]{ .502,  .502,  .502}0.0523646(1) \\
          & std.  & 0.118551 & 0.19771 & 1.64449 & 0.0156748 & 0.000573039 & 0.000301672 \\
    \multirow{2}[0]{*}{MOPW8} & mean  & 0.576179(5)- & 0.308019(4)- & 1.2373(6)- & \cellcolor[rgb]{ .749,  .749,  .749}0.0663105(2)- & 0.0826477(3)- & \cellcolor[rgb]{ .502,  .502,  .502}0.0486113(1) \\
          & std.  & 0.203274 & 0.0988448 & 0.613523 & 0.0166436 & 0.00178718 & 0.0117313 \\
    \multirow{2}[0]{*}{MOPW9} & mean  & 0.118418(3)- & 0.554508(5)- & 0.18049(4)- & \cellcolor[rgb]{ .502,  .502,  .502}0.105952(1)+ & 9.14241(6)- & \cellcolor[rgb]{ .749,  .749,  .749}0.110608(2) \\
          & std.  & 0.0107259 & 0.190845 & 0.0728996 & 0.0221261 & 19.2332 & 0.0944564 \\
    \multirow{2}[0]{*}{MOPW10} & mean  & 0.321016(3)- & 0.587971(5)- & 0.517642(4)- & \cellcolor[rgb]{ .749,  .749,  .749}0.0837639(2)- & 10.1519(6)- & \cellcolor[rgb]{ .502,  .502,  .502}0.0536997(1) \\
          & std.  & 0.0477572 & 0.257621 & 0.146186 & 0.0252723 & 47.1816 & 0.000844333 \\
    \multirow{2}[0]{*}{MOPW11} & mean  & \cellcolor[rgb]{ .502,  .502,  .502}0.0453229(1)+ & \cellcolor[rgb]{ .749,  .749,  .749}0.0463542(2)+ & 0.0483176(3)+ & 0.0949665(6)- & 0.0810912(5)- & 0.0627861(4) \\
          & std.  & 0.00151415 & 0.00230156 & 0.00201648 & 0.022435 & 0.000287571 & 0.00162099 \\
    \multirow{2}[1]{*}{MOPW12} & mean  & \cellcolor[rgb]{ .749,  .749,  .749}0.0525026(2)+ & \cellcolor[rgb]{ .502,  .502,  .502}0.0497525(1)+ & 0.0571428(3)+ & 0.0730452(6)- & 0.0639918(4)+ & 0.0682657(5) \\
          & std.  & 0.00162342 & 0.00073207 & 0.00258147 & 0.00588929 & 0.00042409 & 0.00228719 \\
    % \midrule
    \multirow{2}[1]{*}{MOPW13} & mean  & 10.9589(5)- & 0.832254(4)- & 236.606(6)- & \cellcolor[rgb]{ .502,  .502,  .502}0.110383(1)+ & 0.298037(3)= & \cellcolor[rgb]{ .749,  .749,  .749}0.247355(2) \\
          & std.  & 33.9883 & 0.496 & 203.089 & 0.0827951 & 0.355841 & 0.0933775 \\
    \multirow{2}[0]{*}{MOPW14} & mean  & 10.0258(5)- & 9.32558(4)- & 67.1759(6)- & \cellcolor[rgb]{ .502,  .502,  .502}0.0645062(1)+ & 0.342366(3)= & \cellcolor[rgb]{ .749,  .749,  .749}0.204409(2) \\
          & std.  & 21.3125 & 47.48 & 99.8717 & 0.0230577 & 0.306823 & 0.0890296 \\
    \multirow{2}[0]{*}{MOPW15} & mean  & 6.71954(5)- & 0.522105(4)- & 126.907(6)- & \cellcolor[rgb]{ .502,  .502,  .502}0.0506183(1)+ & 0.340082(3)= & \cellcolor[rgb]{ .749,  .749,  .749}0.183966(2) \\
          & std.  & 13.5024 & 0.276269 & 180.016 & 0.0180945 & 0.430414 & 0.118939 \\
    \multirow{2}[1]{*}{MOPW16} & mean  & \cellcolor[rgb]{ .749,  .749,  .749}0.041826(2)+ & \cellcolor[rgb]{ .502,  .502,  .502}0.0403934(1)+ & 0.0476254(3)+ & 0.0945093(6)- & 0.053768(4)+ & 0.0677451(5) \\
          & std.  & 0.00146483 & 0.0019857 & 0.00242714 & 0.00627306 & 0.000751424 & 0.00166284 \\
    \midrule
    \multicolumn{2}{c}{Total +/=/-} & 5/0/11 & 5/0/11 & 4/1/11 & 6/0/10 & 3/3/10 & \textbackslash{} \\
    \multicolumn{2}{c}{Average rank} & 3.375(4) & 3.25(3) & 5(6)  & 2.875(2) & 4.0625(5) & 2.4375(1) \\
    \bottomrule
    \end{tabular}%
    \begin{tablenotes}
        \item[] The rank of each algorithm on each instance is provided after the mean of the metric value.
        \item[] The best and second-best mean metric values are highlighted using gray and light gray shading, respectively.
        \item[] The Wilcoxon rank-sum test with a 0.05 significance level is employed to statistically analyze the algorithms' performance on each instance. The symbols ``+'', ``='', and ``-'' indicate that the performance of the corresponding algorithm is statistically better than, comparable to, or worse than that of the rightmost algorithm.
    \end{tablenotes}
    \end{threeparttable}
    }
  \label{tab:res_existing}%
\end{table*}%

\subsection{Results}
Table~\ref{tab:res_existing} shows the statistical results on 16 benchmark instances. No algorithm consistently performs best across the majority of instances. IMOEA-ARP, MultiGPO, RVEA-iGNG, and PREA achieve the best performance on 6, 5, 4, and 1 instances, respectively. IMOEA-ARP demonstrates the best overall performance in terms of the average rank, followed by MultiGPO. 
Moreover, IMOEA-ARP achieves the best IGD metric values on MOPW1, MOPW3, MOPW4, MOPW7, MOPW8, and MOPW10, while MultiGPO delivers the best results on MOPW2, MOPW9, and MOPW13--MOPW15. This observation implies that MultiGPO exhibits robustness to variations in the $WPB$'s overall size but is sensitive to changes in the $WPB_{2,i}$'s relative size, which is similar to the finding in Appendix~\ref{sec:impact_wpb_size}.

\begin{figure*}[ht]
\centering
\subfloat[PREA]{\includegraphics[width=0.32\linewidth]{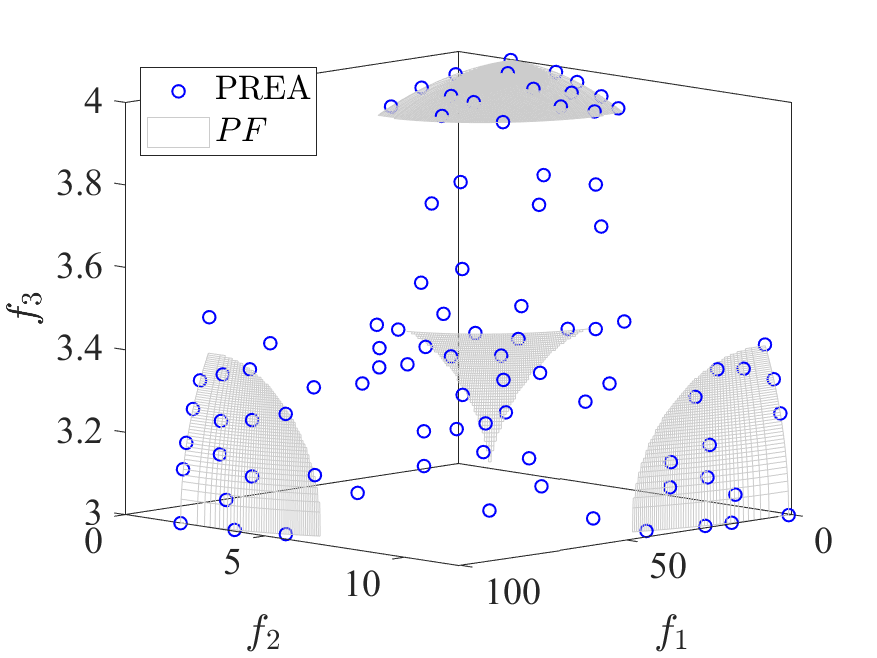}}
\hfil
\subfloat[RVEA-iGNG]{\includegraphics[width=0.32\linewidth]{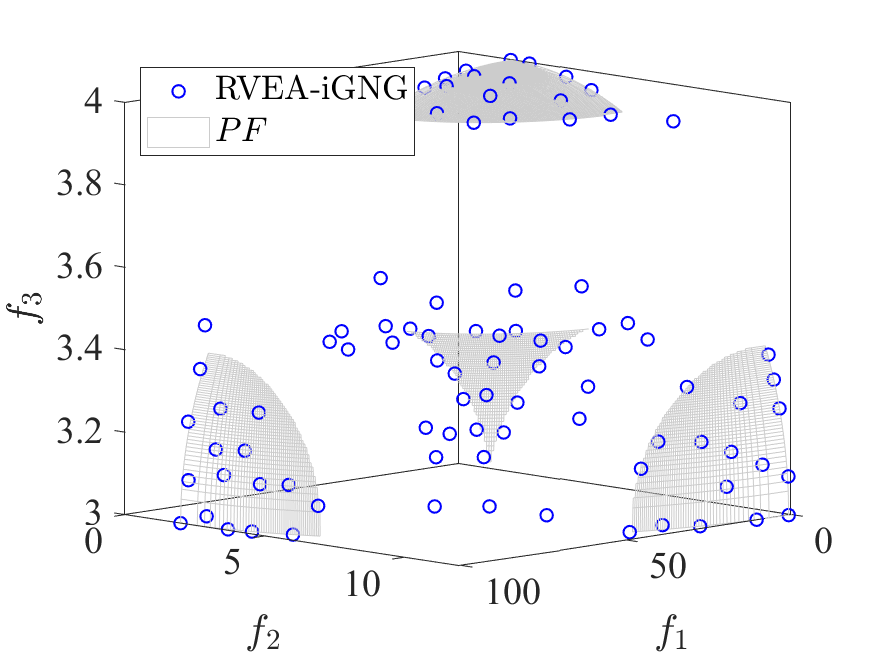}}
\hfil
\subfloat[LMPFE]{\includegraphics[width=0.32\linewidth]{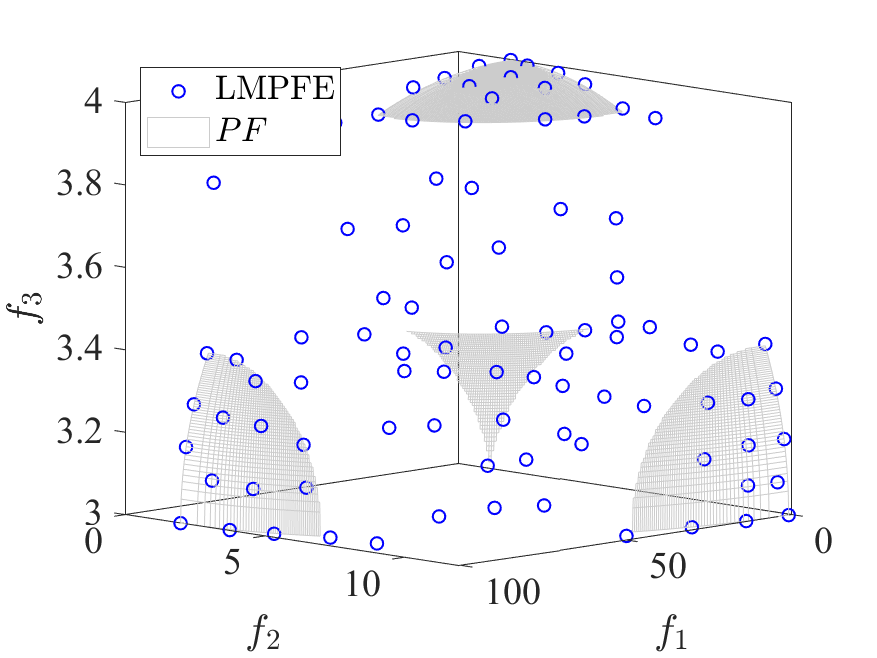}}
\hfil
\subfloat[MultiGPO]{\includegraphics[width=0.32\linewidth]{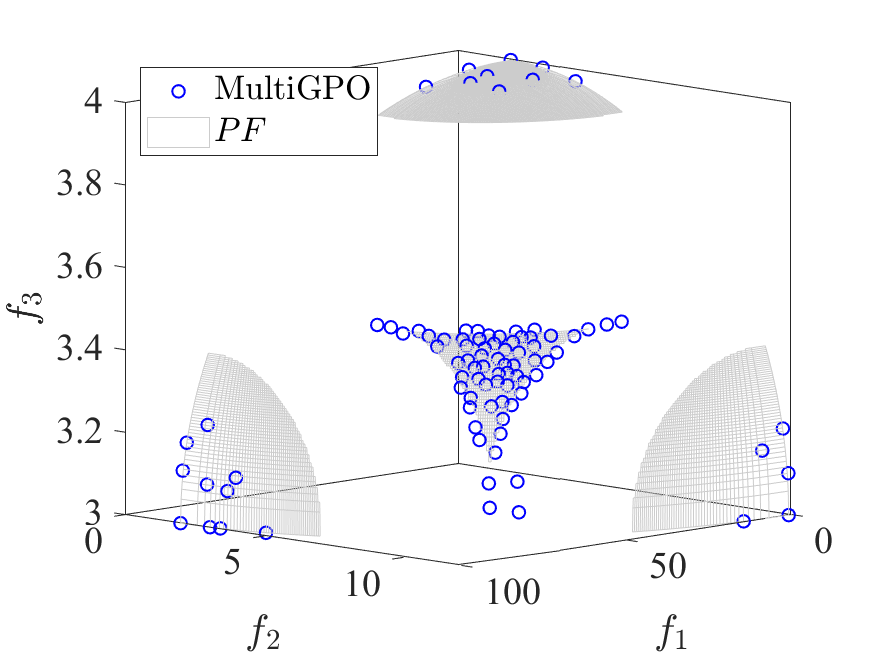}}
\hfil
\subfloat[MOEA/D-Gen]{\includegraphics[width=0.32\linewidth]{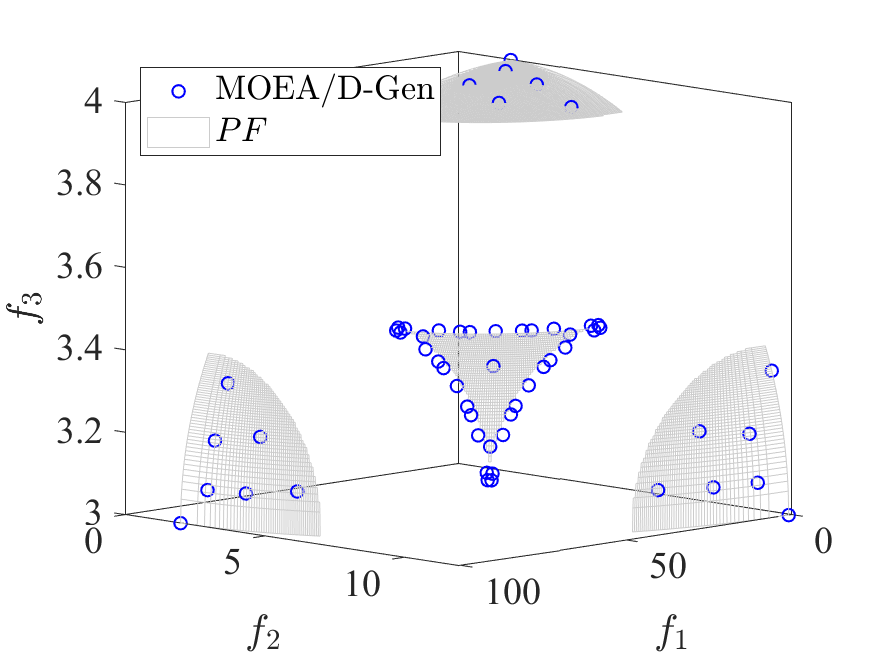}}
\hfil
\subfloat[IMOEA-ARP]{\includegraphics[width=0.32\linewidth]{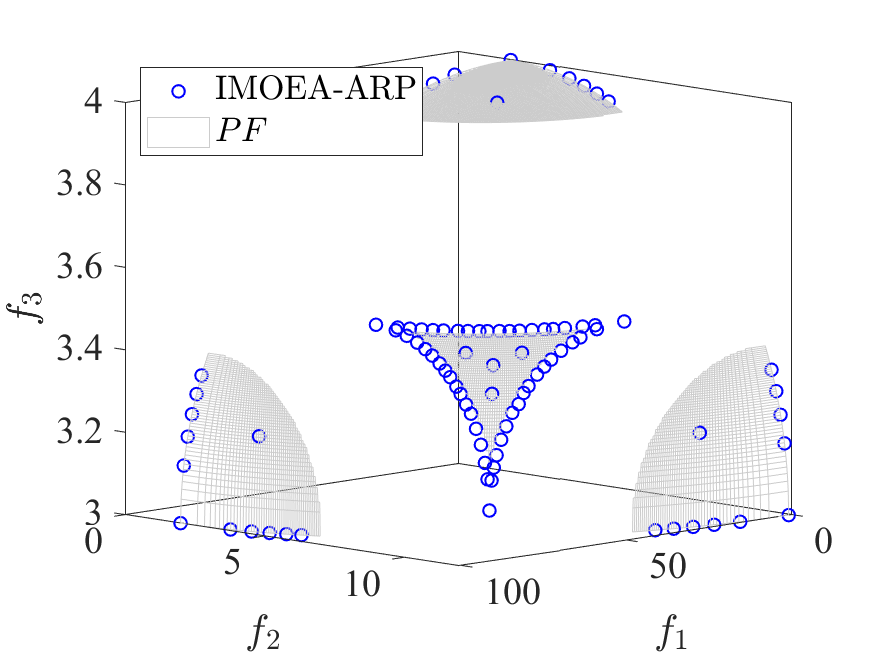}}
\hfil
\caption{Plots of the final populations in the objective space with the second-worst IGD metric values across 30 runs on MOPW16.}
\label{fig:objs_disc_bench}
\end{figure*}

Additionally, MultiGPO and IMOEA-ARP have superior performance on the MOPs with continuous $PF$s (\ie, MOPW1--MOPW3, MOPW7--MOPW9, and MOPW13--MOPW15). However, their performance degrades on the MOPs with discontinuous $PF$s (\ie, MOPW5, MOPW6, MOPW11, MOPW12, and MOPW16). PREA and RVEA-iGNG outperform other algorithms on the MOPs with discontinuous $PF$s. In \figurename~\ref{fig:objs_disc_bench}, we can find that MultiGPO and IMOEA-ARP identify only a limited number of objective vectors on the three boundary parts of the $PF$. And the objective vectors fail to spread widely over the three boundary parts. In contrast, PREA and RVEA-iGNG achieve favorable distribution across all parts of the $PF$, although they maintain some DRSs.

% \newpage

 

% \end{IEEEbiography}



\vfill

% \end{CJK*}
\end{document}